\pdfoutput=1

\documentclass[5p,times]{elsarticle}

\usepackage{lineno}
\usepackage{hyperref}
\usepackage{mathptmx} %
\usepackage{amsmath,amssymb,mathtools}
\usepackage{graphicx,caption,subcaption} \graphicspath{ {figure/} }
\usepackage{enumitem}
\usepackage{acronym}
\usepackage[ruled,linesnumbered]{algorithm2e}
\usepackage{xspace}
\usepackage{setspace}
\usepackage[capitalise]{cleveref}
\hypersetup{colorlinks,
citecolor=[rgb]{0.0,0.3,0.0},
filecolor=[rgb]{0.6,0.0,0.0},
linkcolor=[rgb]{0.0,0.0,0.4},
urlcolor=[rgb]{0.0,0.0,0.4}}

\makeatletter
\def\ps@pprintTitle{%
 \let\@oddhead\@empty
 \let\@evenhead\@empty
 \def\@oddfoot{}%
 \let\@evenfoot\@oddfoot}
\makeatother

\acrodef{ai}[AI]{artificial intelligence}
\acrodef{dnn}[DNNs]{deep neural networks}
\acrodef{rl}[RL]{reinforcement learning}
\acrodef{cc}[CC]{Common Cause}
\acrodef{ce}[CE]{Common Effect}
\acrodef{cc3}[CC3]{Common Cause 3}
\acrodef{ce3}[CE3]{Common Effect 3}
\acrodef{cc4}[CC4]{Common Cause 4}
\acrodef{ce4}[CE4]{Common Effect 4}
\acrodef{mcmc}[MCMC]{markov chain monte carlo}
\acrodef{a2c}[A2C]{Advantage Actor-Critic}
\acrodef{trpo}[TRPO]{Trust Region Policy Optimization}
\acrodef{ppo}[PPO]{Proximal Policy Optimization}
\acrodef{hsg}[HSG]{holistic scene grammar}
\acrodef{lfd}[LfD]{learning from demonstration} 
\acrodef{fem}[FEM]{finite element method}
\acrodef{mpm}[MPM]{material point method}
\acrodef{rpm}[RPM]{raven's prograssive matrices test}
\acrodef{vqa}[VQA]{visual question answering}
\acrodef{asig}[A-SIG]{attributed stochastic image grammar}
\acrodef{nce}[NCE]{noise-contrastive estimation}
\acrodef{mas}[MAS]{multiple-agent systems}

\journal{Engineering}

\hfuzz=\maxdimen
\vfuzz=\maxdimen

\makeatletter
\DeclareRobustCommand\onedot{\futurelet\@let@token\@onedot}
\def\@onedot{\ifx\@let@token.\else.\null\fi\xspace}
\def\eg{\emph{e.g}\onedot} 
\def\ie{\emph{i.e}\onedot} 
 
\def\etc{\emph{etc}\onedot} 
 
\def\etal{\emph{et al}\onedot}
\makeatother

\begin{document}

\begin{frontmatter}

\title{Dark, Beyond Deep: A Paradigm Shift to Cognitive AI with Humanlike Common Sense}

\author[ucla]{Yixin Zhu\corref{cor1}}
\ead{yixin.zhu@ucla.edu}
\author[ucla]{Tao Gao}
\author[ucla]{Lifeng Fan}
\author[ucla]{Siyuan Huang}
\author[ucla]{Mark Edmonds}
\author[ucla]{Hangxin Liu}
\author[ucla]{Feng Gao}
\author[ucla]{Chi Zhang}
\author[ucla]{Siyuan Qi}
\author[ucla]{\quad{}\quad{}Ying Nian Wu}
\author[mit]{Joshua B. Tenenbaum}
\author[ucla]{Song-Chun Zhu}

\cortext[cor1]{Corresponding author}

\address[ucla]{Center for Vision, Cognition, Learning, and Autonomy (VCLA), UCLA}
\address[mit]{Center for Brains, Minds, and Machines (CBMM), MIT}

\begin{abstract}
Recent progress in deep learning is essentially based on a ``big data for small tasks'' paradigm, under which massive amounts of data are used to train a classifier for a single narrow task. In this paper, we call for a shift that flips this paradigm upside down. Specifically, we propose a ``small data for big tasks'' paradigm, wherein a single \ac{ai} system is challenged to develop ``common sense,'' enabling it to solve a wide range of tasks with little training data. We illustrate the potential power of this new paradigm by reviewing models of common sense that synthesize recent breakthroughs in both machine and human vision. We identify functionality, physics, intent, causality, and utility (FPICU) as the five core domains of cognitive \ac{ai} with humanlike common sense. When taken as a unified concept, FPICU is concerned with the questions of ``why'' and ``how,'' beyond the dominant ``what'' and ``where'' framework for understanding vision. They are invisible in terms of pixels but nevertheless drive the creation, maintenance, and development of visual scenes. We therefore coin them the ``dark matter'' of vision. Just as our universe cannot be understood by merely studying observable matter, we argue that vision cannot be understood without studying FPICU. We demonstrate the power of this perspective to develop cognitive \ac{ai} systems with humanlike common sense by showing how to observe and apply FPICU with little training data to solve a wide range of challenging tasks, including tool use, planning, utility inference, and social learning. In summary, we argue that the next generation of \ac{ai} must embrace ``dark'' humanlike common sense for solving novel tasks.
\end{abstract}

\end{frontmatter}

\section{A Call for a Paradigm Shift in Vision and \ac{ai}}

\begin{figure*}[t!]
	\centering
	\includegraphics[width=\linewidth]{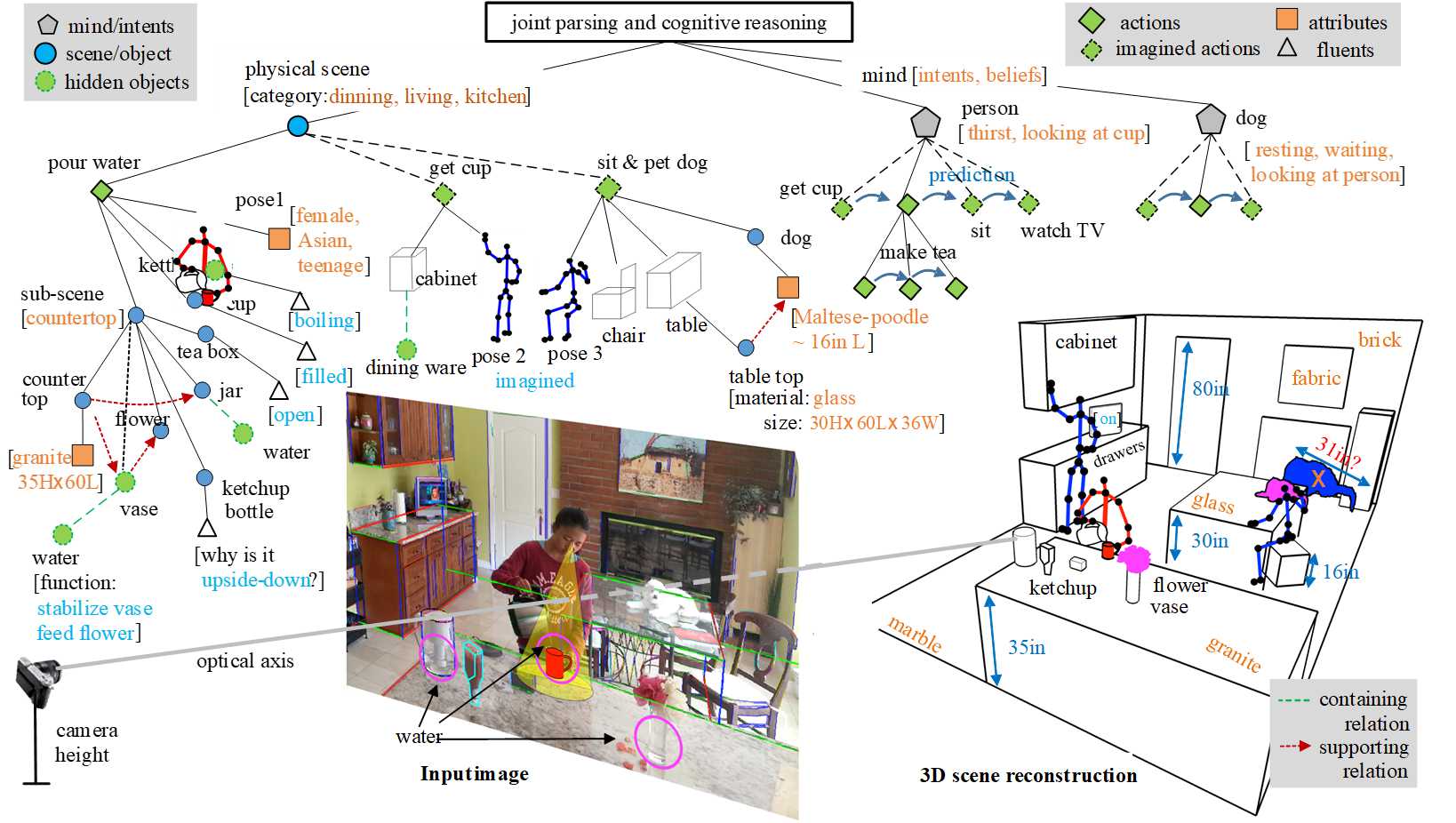}
	\caption{An example of in-depth understanding of a scene or event through joint parsing and cognitive reasoning. From a single image, a computer vision system should be able to jointly (i) reconstruct the 3D scene; (ii) estimate camera parameters, materials, and illumination; (iii) parse the scene hierarchically with attributes, fluents, and relationships; (iv) reason about the intentions and beliefs of agents (\eg, the human and dog in this example); (v) predict their actions in time; and (vi) recover invisible elements such as water, latent object states, and so forth. We, as humans, can effortlessly (i) predict that water is about to come out of the kettle; (ii) reason that the intent behind putting the ketchup bottle upside down is to utilize gravity for easy use; and (iii) see that there is a glass table, which is difficult to detect with existing computer vision methods, under the dog; without seeing the glass table, parsing results would violate the laws of physics, as the dog would appear to be floating in midair. These perceptions can only be achieved by reasoning about unobservable factors in the scene not represented by pixels, requiring us to build an \ac{ai} system with humanlike core knowledge and common sense, which are largely missing from current computer vision research. H: height; L: length; W: width. 1 in = 2.54 cm.}
	\label{fig:motivation}
\end{figure*}

Computer vision is the front gate to \acf{ai} and a major component of modern intelligent systems. The classic definition of computer vision proposed by the pioneer David Marr~\citep{marr1982vision} is to look at ``what'' is ``where.'' Here, ``what'' refers to object recognition (object vision), and ``where'' denotes three-dimensional (3D) reconstruction and object localization (spatial vision)~\citep{mishkin1983object}. Such a definition corresponds to two pathways in the human brain: (i) the ventral pathway for categorical recognition of objects and scenes, and (ii) the dorsal pathway for the reconstruction of depth and shapes, scene layout, visually guided actions, and so forth. This paradigm guided the geometry-based approaches to computer vision of the 1980s-1990s, and the appearance-based methods of the past 20 years.

Over the past several years, progress has been made in object detection and localization with the rapid advancement of \ac{dnn}, fueled by hardware accelerations and the availability of massive sets of labeled data. However, we are still far from solving computer vision or real machine intelligence; the inference and reasoning abilities of current computer vision systems are narrow and highly specialized, require large sets of labeled training data designed for special tasks, and lack a general \emph{understanding} of common facts---that is, facts that are obvious to the average human adult---that describe how our physical and social worlds work. To fill in the gap between modern computer vision and human vision, we must find a broader perspective from which to model and reason about the missing dimension, which is humanlike common sense.

This state of our understanding of vision is analogous to what has been observed in the fields of cosmology and astrophysicists. In the 1980s, physicists proposed what is now the standard cosmology model, in which the mass-energy observed by the electromagnetic spectrum accounts for less than $5\%$ of the universe; the rest of the universe is dark matter ($23\%$) and dark energy ($72\%$)\footnote{\url{https://map.gsfc.nasa.gov/universe/}}. The properties and characteristics of dark matter and dark energy cannot be observed and must be reasoned from the visible mass-energy using a sophisticated model. Despite their invisibility, however, dark matter and energy help to explain the formation, evolution, and motion of the visible universe.

We intend to borrow this physics concept to raise awareness, in the vision community and beyond, of the missing dimensions and the potential benefits of joint representation and joint inference. We argue that humans can make rich inferences from sparse and high-dimensional data, and achieve deep understanding from a single picture, because we have common yet visually imperceptible knowledge that can never be understood just by asking ``what'' and ``where.'' Specifically, human-made objects and scenes are designed with latent functionality, determined by the unobservable laws of physics and their down-stream causal relationships; consider how our understanding of water's flow from of a kettle, or our knowledge that a transparent substance such as glass can serve as a solid table surface, tells us what is happening in \cref{fig:motivation}. Meanwhile, human activities, especially social activities, are governed by causality, physics, functionality, social intent, and individual preferences and utility. In images and videos, many entities (\eg, functional objects, fluids, object fluents, and intent) and relationships (\eg, causal effects and physical supports) are impossible to detect by most of the existing approaches considering appearance alone; these latent factors are not represented in pixels. Yet they are pervasive and govern the placement and motion of the visible entities that are relatively easy for current methods to detect.

These invisible factors are largely missing from recent computer vision literature, in which most tasks have been converted into classification problems, empowered by large-scale annotated data and end-to-end training using neural networks. This is what we call the ``big data for small tasks'' paradigm of computer vision and \ac{ai}.

\begin{figure*}[t!]
	\centering
	\begin{subfigure}[b]{0.297\linewidth}
		\includegraphics[width=\linewidth]{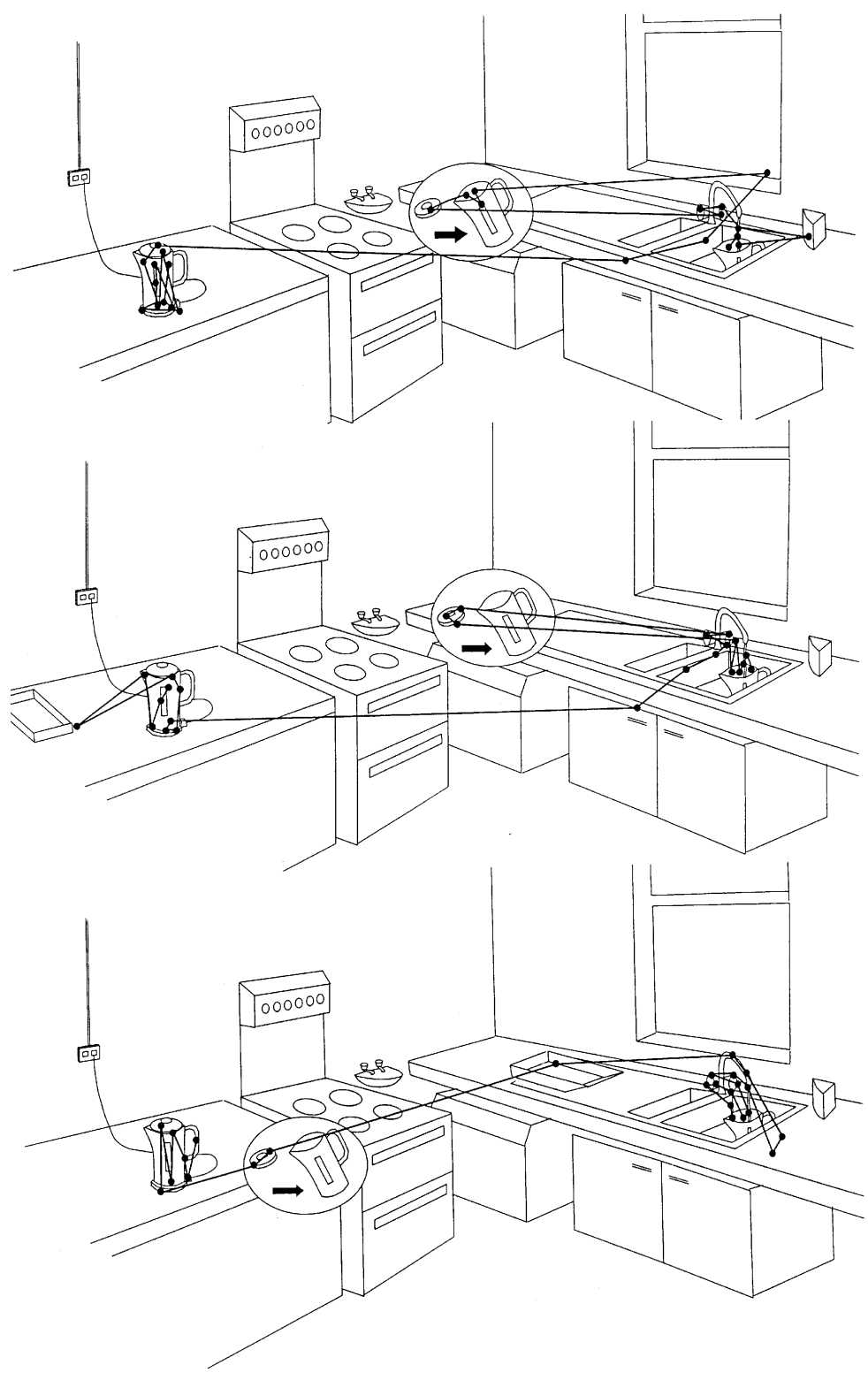}
		\caption{}
	\end{subfigure}%
	\begin{subfigure}[b]{0.703\linewidth}
		\includegraphics[width=\linewidth]{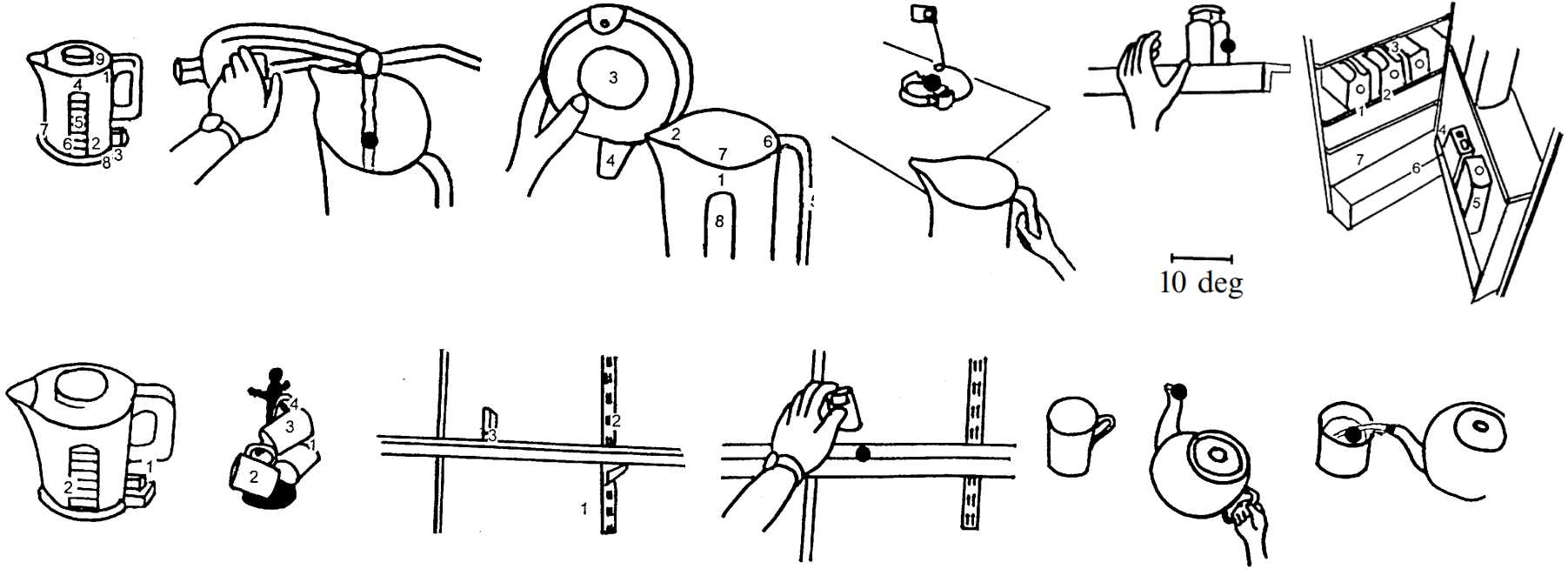}
		\caption{}%
		\includegraphics[width=\linewidth]{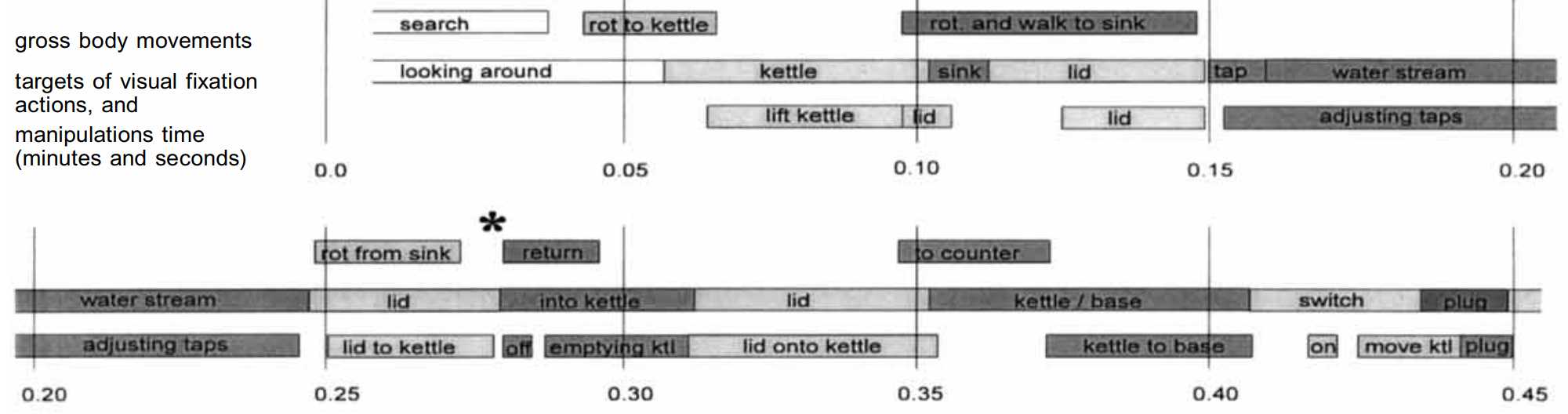}
		\caption{}
	\end{subfigure}%
	\caption{Even for as ``simple'' a task as making a cup of tea, a person can make use of his or her single vision system to perform a variety of subtasks in order to achieve the ultimate goal. (a) Record of the visual fixations of three different subjects performing the same task of making a cup of tea in a small rectangular kitchen; (b) examples of fixation patterns drawn from an eye-movement videotape; (c) a sequence of visual and motor events during a tea-making session. Rot: rotate; ktl: kettle. Reproduced from Ref.~\citep{land1999roles} with permission of SAGE Publication, \textcopyright~1999.}
	\label{fig:thousands_tasks}
\end{figure*}

In this paper, we aim to draw attention to a promising new direction, where consideration of ``dark'' entities and relationships is incorporated into vision and \ac{ai} research. By reasoning about the unobservable factors beyond visible pixels, we could approximate humanlike common sense, using limited data to achieve generalizations across a variety of tasks. Such tasks would include a mixture of both classic ``what and where'' problems (\ie, classification, localization, and reconstruction), and ``why, how, and what if'' problems, including but not limited to causal reasoning, intuitive physics, learning functionality and affordance, intent prediction, and utility learning. We coin this new paradigm ``small data for big tasks.''

Of course, it is well-known that vision is an ill-posed inverse problem~\citep{marr1982vision} where only pixels are seen directly, and anything else is hidden/latent. The concept of ``darkness'' is perpendicular to and richer than the meanings of ``latent'' or ``hidden'' used in vision and probabilistic modeling; ``darkness'' is a measure of the relative difficulty of classifying an entity or inferring about a relationship based on how much invisible common sense needed beyond the visible appearance or geometry. Entities can fall on a continuous spectrum of ``darkness''---from objects such as a generic human face, which is relatively easy to recognize based on its appearance, and is thus considered ``visible,'' to functional objects such as chairs, which are challenging to recognize due to their large intraclass variation, and all the way to entities or relationships that are impossible to recognize through pixels. In contrast, the functionality of the kettle is ``dark;'' through common sense, a human can easily infer that there is liquid inside it. The position of the ketchup bottle could also be considered ``dark,'' as the understanding of typical human intent lets us understand that it has been placed upside down to harness gravity for easy dispensing.

The remainder of this paper starts by revisiting a classic view of computer vision in terms of ``what'' and ``where'' in \cref{sec:vision}, in which we show that the human vision system is essentially task-driven, with its representation and computational mechanisms rooted in various tasks. In order to use ``small data'' to solve ``big tasks,'' we then identify and review five crucial axes of visual common sense: \textbf{F}unctionality, \textbf{P}hysics, perceived \textbf{I}ntent, \textbf{C}ausality, and \textbf{U}tility (FPICU). Causality (\cref{sec:causal}) is the basis for intelligent understanding. The application of causality (\ie, intuitive physics; \cref{sec:physics}) affords humans the ability to understand the physical world we live in. Functionality (\cref{sec:function}) is a further understanding of the physical environment humans use when they interact with it, performing appropriate actions to change the world in service of activities. When considering social interactions beyond the physical world, humans need to further infer intent (\cref{sec:intention}) in order to understand other humans' behavior. Ultimately, with the accumulated knowledge of the physical and social world, the decisions of a rational agent are utility-driven (\cref{sec:utility}). In a series of studies, we demonstrate that these five critical aspects of ``dark entities'' and ``dark relationships'' indeed support various visual tasks beyond just classification. We summarize and discuss our perspectives in \cref{sec:discussion}, arguing that it is crucial for the future of \ac{ai} to master these essential unseen ingredients, rather than only increasing the performance and complexity of data-driven approaches.

\section{Vision: From Data-driven to Task-driven}\label{sec:vision}

What should a vision system afford the agent it serves? From a biological perspective, the majority of living creatures use a \emph{single} (with multiple components) vision system to perform \emph{thousands} of tasks. This contrasts with the dominant contemporary stream of thought in computer vision research, where a single model is designed specifically for a single task. In the literature, this organic paradigm of generalization, adaptation, and transfer among various tasks is referred to as task-centered vision~\citep{ikeuchi1996task}. In the kitchen shown in \cref{fig:thousands_tasks}~\citep{land1999roles}, even a task as simple as making a cup of coffee consists of multiple subtasks, including finding objects (object recognition), grasping objects (object manipulation), finding milk in the refrigerator, and adding sugar (task planning). Prior research has shown that a person can finish making a cup of coffee within 1 min by utilizing a single vision system to facilitate the performance of a variety of subtasks~\citep{land1999roles}.

\begin{figure}[t!]
	\centering
	\begin{subfigure}[b]{0.85\linewidth}
		\includegraphics[width=\linewidth]{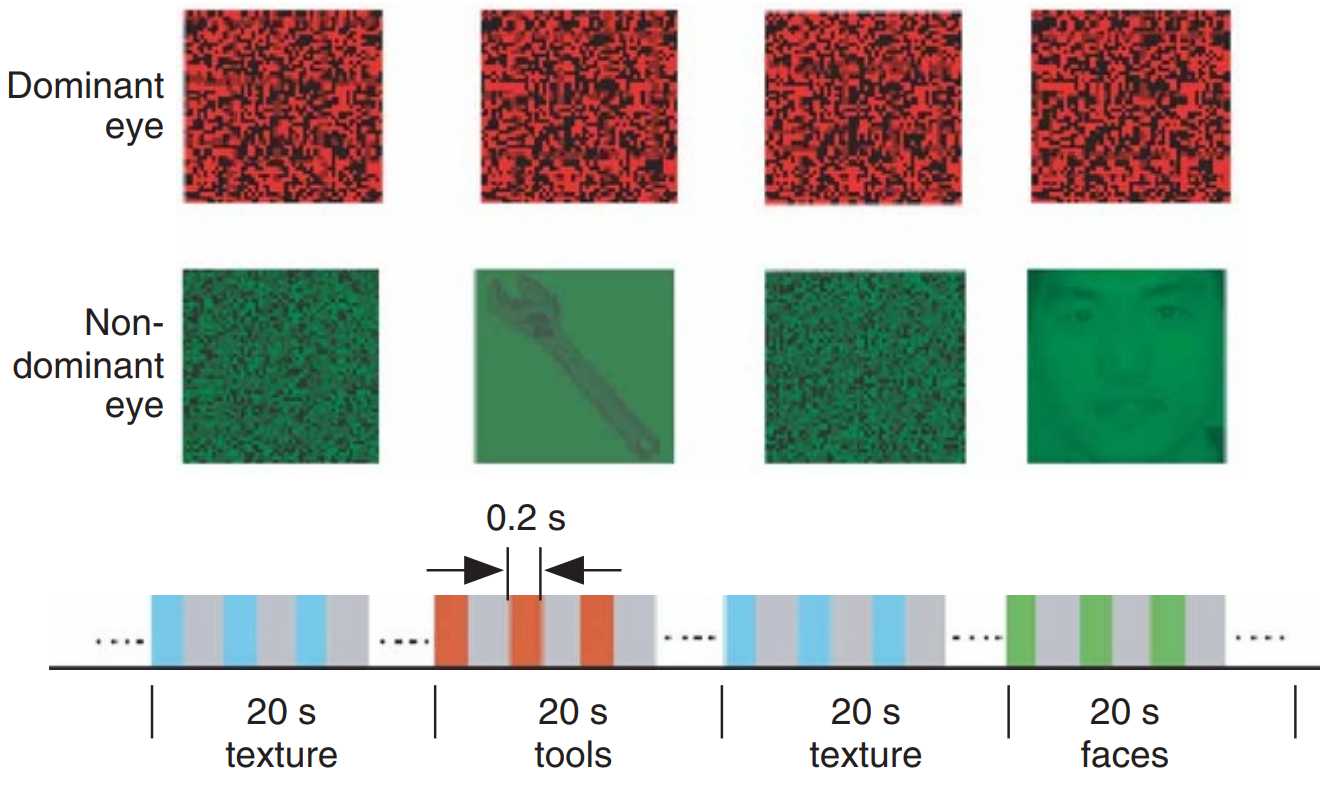}
		\caption{}
	\end{subfigure}%
	\\
	\begin{subfigure}[b]{0.85\linewidth}
		\includegraphics[width=\linewidth]{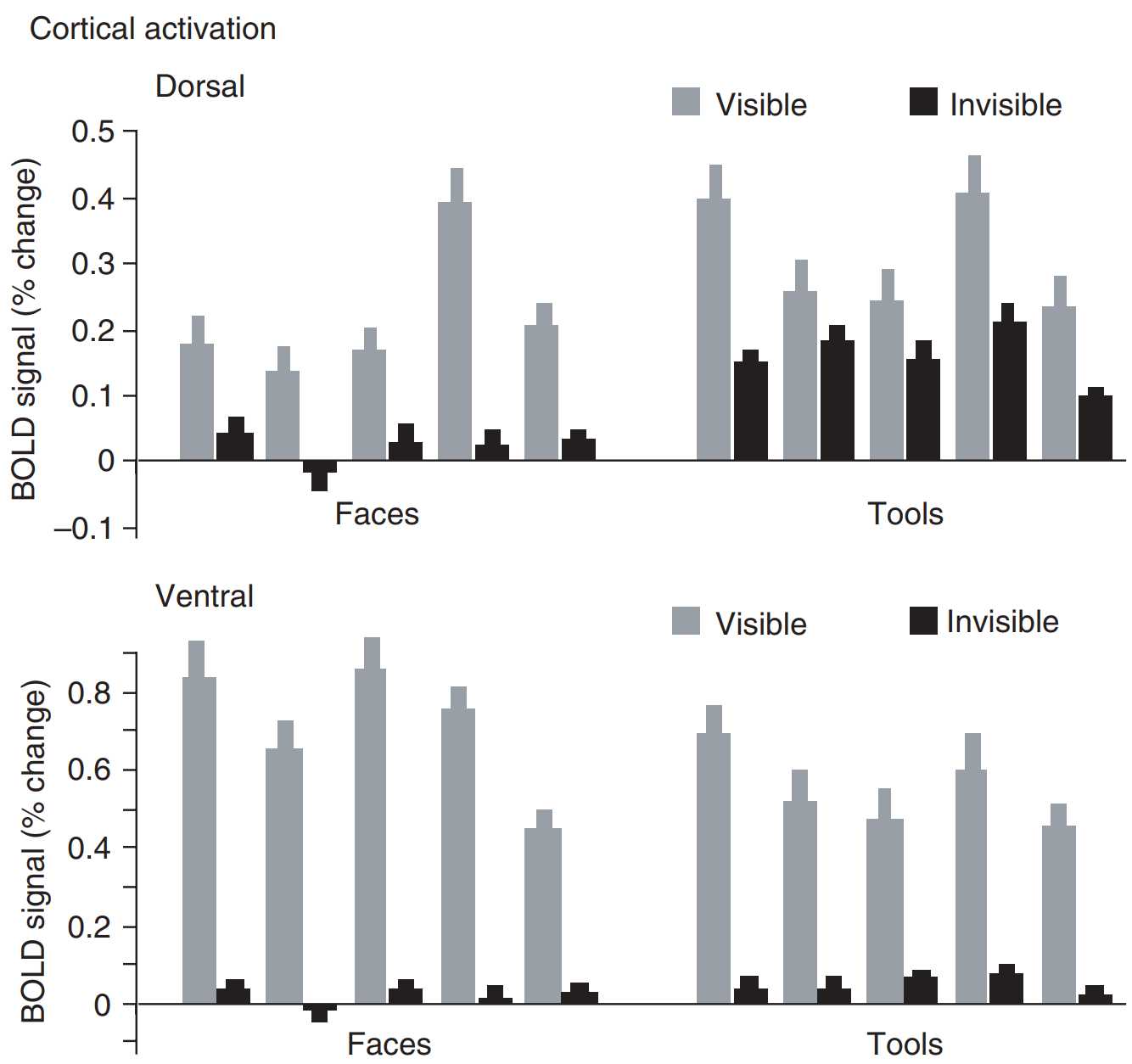}
		\caption{}
	\end{subfigure}%
	\caption{Cortical responses to invisible objects in the human dorsal and ventral pathways. (a) Stimuli (tools and faces) and experimental procedures; (b) both the dorsal and ventral areas responded to tools and faces. When stimuli were suppressed by high-contrast dynamic textures, the dorsal response remained responsive to tools, but not to faces, while neither tools or faces evoked much activation in the ventral area. BOLD: blood oxygen level-dependent. Reproduced from Ref.~\citep{fang2005cortical} with permission of Nature Publishing Group, \textcopyright~2005.}
	\label{fig:dorsal_ventral}
\end{figure}

Neuroscience studies suggest similar results, indicating that the human vision system is far more capable than any existing computer vision system, and goes beyond merely memorizing patterns of pixels. For example, Fang and He~\citep{fang2005cortical} showed that recognizing a face inside an image utilizes a different mechanism from recognizing an object that can be manipulated as a tool, as shown in \cref{fig:dorsal_ventral}; indeed, their results show that humans may be even more visually responsive to the appearance of tools than to faces, driving home how much reasoning about how an object can help perform tasks is ingrained in visual intelligence. Other studies~\citep{creem2005neural} also support the similar conclusion that images of tools ``potentiate'' actions, even when overt actions are not required. Taken together, these results indicate that our biological vision system possesses a mechanism for perceiving object functionality (\ie, how an object can be manipulated as a tool) that is independent of the mechanism governing face recognition (and recognition of other objects). All these findings call for a quest to discover the mechanisms of the human vision system and natural intelligence.

\begin{figure}[t!]
	\centering
	\includegraphics[width=\linewidth]{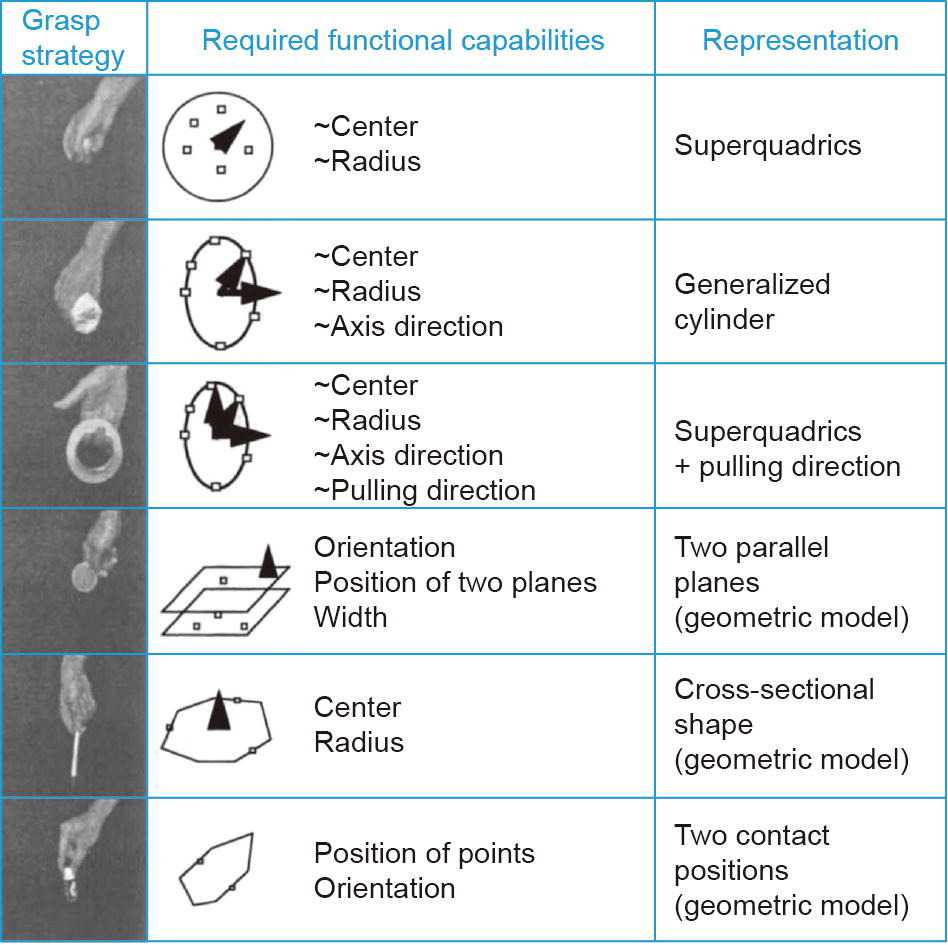}
	\caption{Different grasping strategies require various functional capabilities. Reproduced from Ref.~\citep{ikeuchi1992task} with permission of IEEE, \textcopyright~1992.}
	\label{fig:task_centered_vision}
\end{figure}

\subsection{``What'': Task-centered Visual Recognition}

The human brain can grasp the ``gist'' of a scene in an image within 200 ms, as observed by Potter in the 1970s~\citep{potter1975meaning,potter1976short}, and by Schyns and Oliva~\citep{schyns1994blobs} and Thorpe \etal~\citep{thorpe1996speed} in the 1990s. This line of work often leads researchers to treat categorization as a data-driven process~\citep{greene2009briefest,greene2009recognition,fei2007we,rousselet2005long,oliva2001modeling}, mostly in a feed-forward network architecture~\citep{delorme2000ultra,serre2007feedforward}. Such thinking has driven image classification research in computer vision and machine learning in the past decade and has achieved remarkable progress, including the recent success of \ac{dnn}~\citep{krizhevsky2012imagenet,kavukcuoglu2010learning,deng2009imagenet}.
	
\begin{figure*}[t!]
	\centering
	\begin{subfigure}[b]{0.333\linewidth}
		\includegraphics[width=\linewidth]{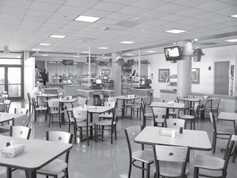}
		\caption{}
	\end{subfigure}%
	\begin{subfigure}[b]{0.333\linewidth}
		\includegraphics[width=\linewidth]{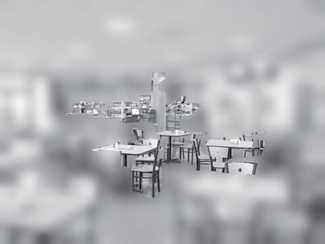}
		\caption{}
	\end{subfigure}%
	\begin{subfigure}[b]{0.333\linewidth}
		\includegraphics[width=\linewidth]{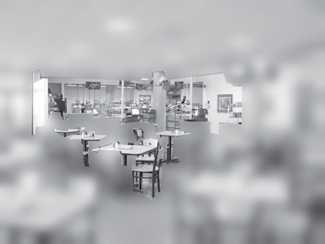}
		\caption{}
	\end{subfigure}%
	\caption{The experiment presented in Ref.~\citep{malcolm2014beyond}, demonstrating the diagnostically driven, bidirectional interplay between top-down and bottom-up information for the categorization of scenes at specific hierarchical levels. (a) Given the same input image of a scene, subjects will show different gaze patterns if they are asked to categorize the scene at (b) a basic level (\eg, restaurant) or (c) a subordinate level (\eg, cafeteria), indicating a task-driven nature of scene categorization. Reproduced from Ref.~\citep{malcolm2014beyond} with permission of the authors, \textcopyright~2014.}
	\label{fig:scene_functionality_categorization}
\end{figure*}

Despite the fact that these approaches achieved good performances on scene categorization in terms of recognition accuracy in publicly available datasets, a recent large-scale neuroscience study~\citep{rajalingham2018large} has shown that current \ac{dnn} cannot account for the image-level behavior patterns of primates (both humans and monkeys), calling attention to the need for more precise accounting for the neural mechanisms underlying primate object vision. Furthermore, data-driven approaches have led the focus of scene categorization research away from an important determinant of visual information---the categorization task itself~\citep{oliva1997coarse,schyns1998diagnostic}. Simultaneously, these approaches have left unclear how classification interacts with scene semantics and enables cognitive reasoning. Psychological studies suggest that human vision organizes representations during the inference process even for ``simple'' categorical recognition tasks. Depending on a viewer's needs (and tasks), a kitchen can be categorized as an indoor scene, a place to cook, a place to socialize, or specifically as one's own kitchen (\cref{fig:scene_functionality_categorization})~\citep{malcolm2014beyond}. As shown in Ref.~\citep{malcolm2014beyond}, scene categorization and the information-gathering process are constrained by these categorization tasks~\citep{qi2017predicting,pei2011parsing}, suggesting a bidirectional interplay between the visual input and the viewer's needs/tasks~\citep{schyns1998diagnostic}. Beyond scene categorization, similar phenomena were also observed in facial recognition~\citep{gosselin2001bubbles}.

In an early work, Ikeuchi and Hebert~\citep{ikeuchi1992task} proposed a task-centered representation inspired by robotic grasping literature. Specifically, without recovering the detailed 3D models, their analysis suggested that various grasp strategies require the object to afford different functional capabilities; thus, the representation of the same object can vary according to the planned task (\cref{fig:task_centered_vision})~\citep{ikeuchi1992task}. For example, grasping a mug could result in two different grasps---the cylindrical grasp of the mug body and the hook grasp of the mug handle. Such findings also suggest that vision (in this case, identifying graspable parts) is largely driven by tasks; different tasks result in diverse visual representations.

\subsection{``Where'': Constructing 3D Scenes as a Series of Tasks}

\begin{figure}[t!]
	\centering
	\includegraphics[width=\linewidth]{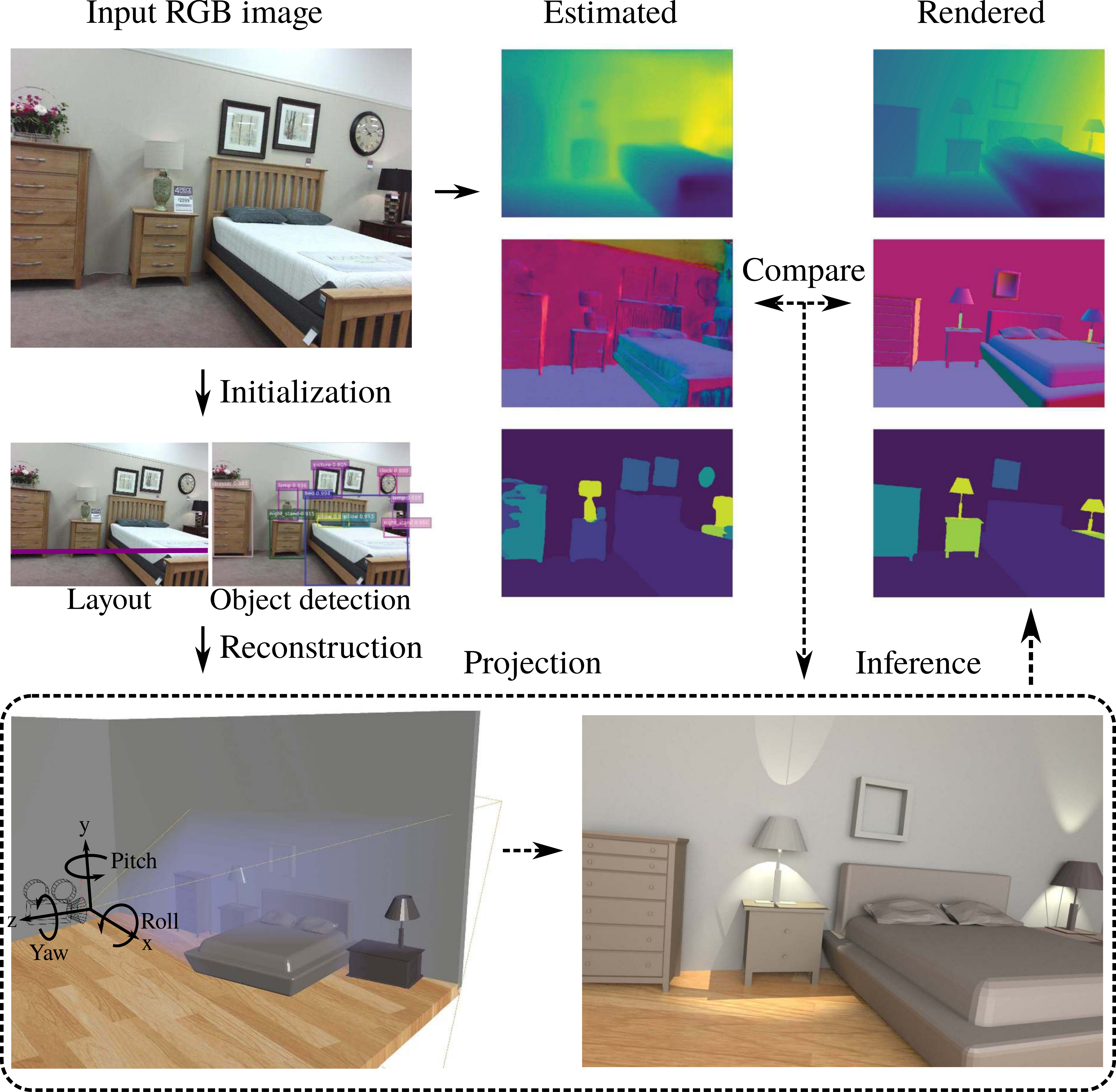}
	\caption{Illustration of 3D indoor scene parsing and reconstruction in an analysis-by-synthesis fashion~\citep{huang2018holistic}. A 3D representation is initialized by individual vision tasks (\eg, object detection, 2D layout estimation). A joint inference algorithm compares the differences between the rendered normal, depth, and segmentation maps and the ones estimated directly from the input RGB image, and adjusts the 3D structure iteratively. Reproduced from Ref.~\citep{huang2018holistic} with permission of Springer,
	\textcopyright~2018.}
	\label{fig:holistic_scene_parsing}
\end{figure}

In the literature, approaches to 3D machine vision have assumed that the goal is to build an accurate 3D model of the scene from the camera/observer's perspective. These structure-from-motion (SfM) and simultaneous localization and mapping (SLAM) methods~\citep{hartley2003multiple} have been the prevailing paradigms in 3D scene reconstruction. In particular, scene reconstruction from a single two-dimensional (2D) image is a well-known ill-posed problem; there may exist an infinite number of possible 3D configurations that match the projected 2D observed images~\citep{ma2012invitation}. However, the goal here is not to precisely match the 3D ground-truth configuration, but to enable agents to perform tasks by generating the best possible configuration in terms of functionality, physics, and object relationships. This line of work has mostly been studied separately from recognition and semantics until recently~\citep{gupta2010estimating,schwing2013box,choi2013understanding,zhao2013scene,liu2018single,huang2018holistic,chen2019holistic,huang20193d}; see \cref{fig:holistic_scene_parsing}~\citep{huang2018holistic} for an example.

The idea of reconstruction as a ``cognitive map'' has a long history~\citep{tolman1948cognitive}. However, our biological vision system does not rely on such precise computations of features and transformations; there is now abundant evidence that humans represent the 3D layout of a scene in a way that fundamentally differs from any current computer vision algorithms~\citep{wang2003comparative,koenderink2002large}. In fact, multiple experimental studies do not countenance global metric representations~\citep{warren2017wormholes,gillner1998navigation,foo2005humans,chrastil2014cognitive,byrne1979memory,tversky1992distortions}; human vision is error-prone and distorted in terms of localization~\citep{ogle1950researches,foley1980binocular,luneburg1947mathematical,indow1991critical,gogel1990theory}. In a case study, Glennerster \etal~\citep{glennerster2006humans} demonstrated an astonishing lack of sensitivity on the part of observers to dramatic changes in the scale of the environment around a moving observer performing various tasks.

Among all the recent evidence, grid cells are perhaps the most well-known discovery to indicate the non-necessity of precise 3D reconstruction for vision tasks~\citep{hafting2005microstructure,killian2012map,o1978hippocampus}. Grid cells encode a cognitive representation of Euclidean space, implying a different mechanism for perceiving and processing locations and directions. This discovery was later awarded the 2014 Nobel Prize in Physiology or Medicine. Surprisingly, this mechanism not only exists in humans~\citep{jacobs2013direct}, but is also found in mice~\citep{fyhn2008grid,doeller2010evidence}, bats~\citep{yartsev2011grid}, and other animals. Gao \etal~\citep{gao2019learning} and Xie \etal~\citep{xie2019representation} proposed a representational model for grid cells, in which the 2D self-position of an agent is represented by a high-dimensional vector, and the 2D self-motion or displacement of the agent is represented by a matrix that transforms the vector. Such a vector-based model is capable of learning hexagon patterns of grid cells with error correction, path integral, and path planning. A recent study also showed that view-based methods actually perform better than 3D reconstruction-based methods in certain human navigation tasks~\citep{gootjes2017comparison}.

Despite these discoveries, how we navigate complex environments while remaining able at all times to return to an original location (\ie, homing) remains a mystery in biology and neuroscience. Perhaps a recent study from Vuong \etal~\citep{vuong2018human} providing evidence for the task-dependent representation of space can shed some light. Specifically, in this experiment, participants made large, consistent pointing errors that were poorly explained by any single 3D representation. Their study suggests that the mechanism for maintaining visual directions for reaching unseen targets is neither based on a stable 3D model of a scene nor a distorted one; instead, participants seemed to form a flat and task-dependent representation.

\subsection{Beyond ``What'' and ``Where'': Towards Scene Understanding with Humanlike Common Sense}

\begin{figure}[t!]
	\centering
	\begin{subfigure}[b]{0.55\linewidth}
	\includegraphics[width=\linewidth]{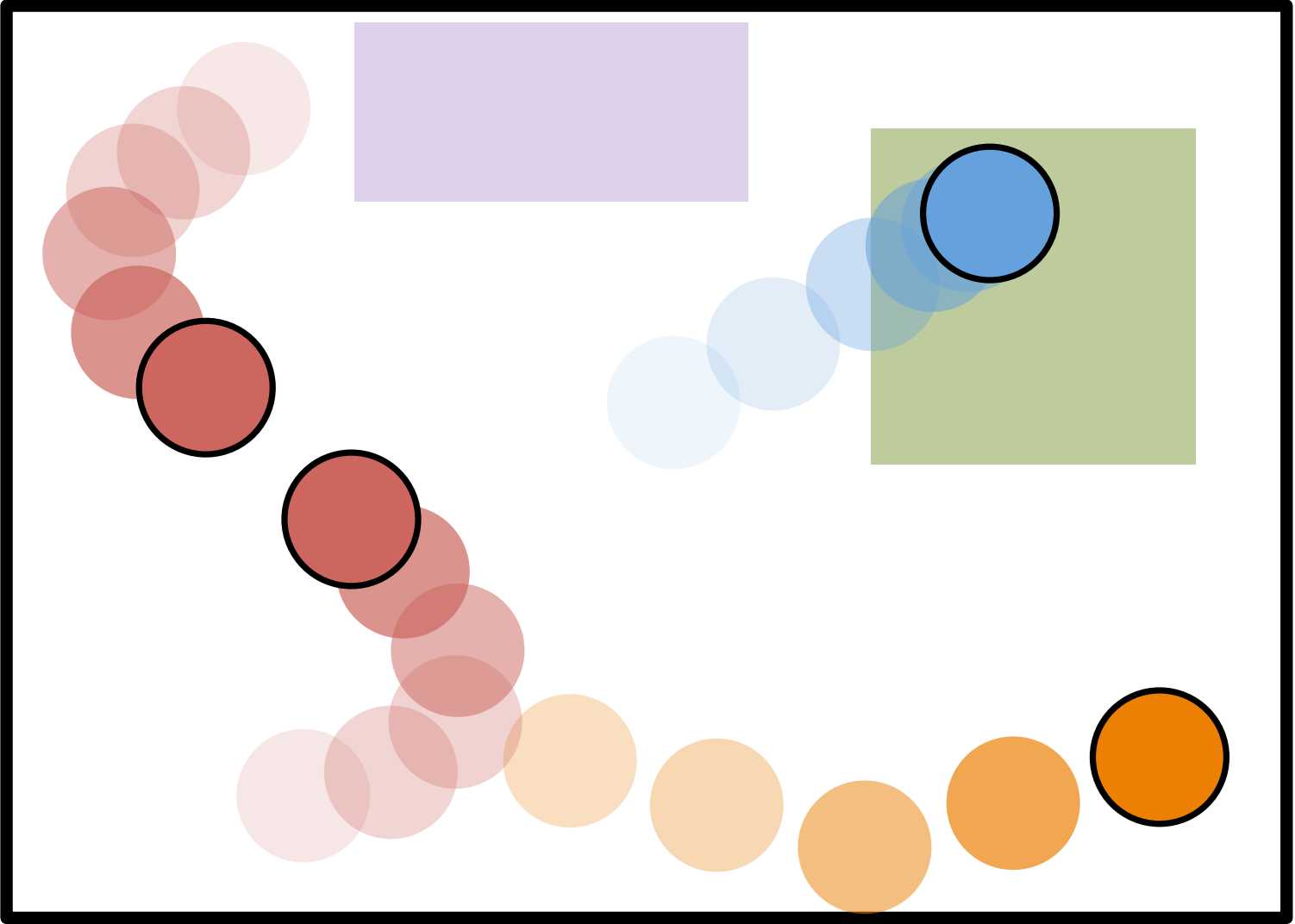}
	\caption{}
	\end{subfigure}%
	\begin{subfigure}[b]{0.45\linewidth}
	\includegraphics[width=\linewidth]{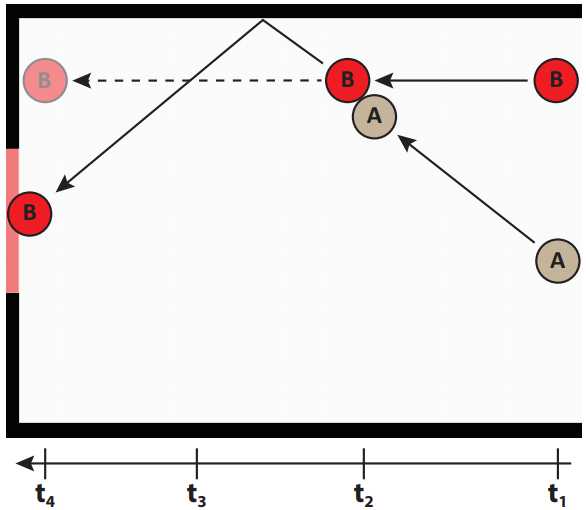}
	\caption{}
	\end{subfigure}%
	\caption{(a) An animation illustrates the intent, mood, and role of the agents~\citep{ullman2014learning}. The motion and interaction of four different pucks moving on a 2D plane are governed by latent physical properties and dynamic laws such as mass, friction, and global and pairwise forces. (b) Intuitive theory and counterfactual reasoning about the dynamics of the scene~\citep{gerstenberg2017intuitive}. Schematic diagram of a collision event between two billiard balls, A and B, where the solid lines indicate the balls' actual movement paths and the dashed line indicates how Ball B would have moved if Ball A had not been present in the scene.}
	\label{fig:physical_social}
\end{figure}

Psychological studies have shown that human visual experience is much richer than ``what'' and ``where.'' As early as infancy, humans quickly and efficiently perceive causal relationships (\eg, perceiving that object A launches object B)~\citep{choi2006perceiving,scholl2004illusory}, agents and intentions (\eg, understanding that one entity is chasing another)~\citep{scholl2013perceiving,scholl2001objects,vul2009explaining}, and the consequences of physical forces (\eg, predicting that a precarious stack of rocks is about to fall in a particular direction)~\citep{battaglia2013simulation,hamrick2011internal}. Such physical and social concepts can be perceived from both media as rich as videos~\citep{xie2018learning} and much sparser visual inputs~\citep{ullman2014learning,gerstenberg2017intuitive}; see examples in \cref{fig:physical_social}.

To enable an artificial agent with similar capabilities, we call for joint reasoning algorithms on a joint representation that integrates (i) the ``visible'' traditional recognition and categorization of objects, scenes, actions, events, and so forth; and (ii) the ``dark'' higher level concepts of fluent, causality, physics, functionality, affordance, intentions/goals, utility, and so forth. These concepts can in turn be divided into five axes: fluent and perceived causality, intuitive physics, functionality, intentions and goals, and utility and preference, described below.

\subsubsection{Fluent and Perceived Causality}

\begin{figure}[t!]
	\centering
	\begin{subfigure}[b]{\linewidth}
		\includegraphics[width=\linewidth]{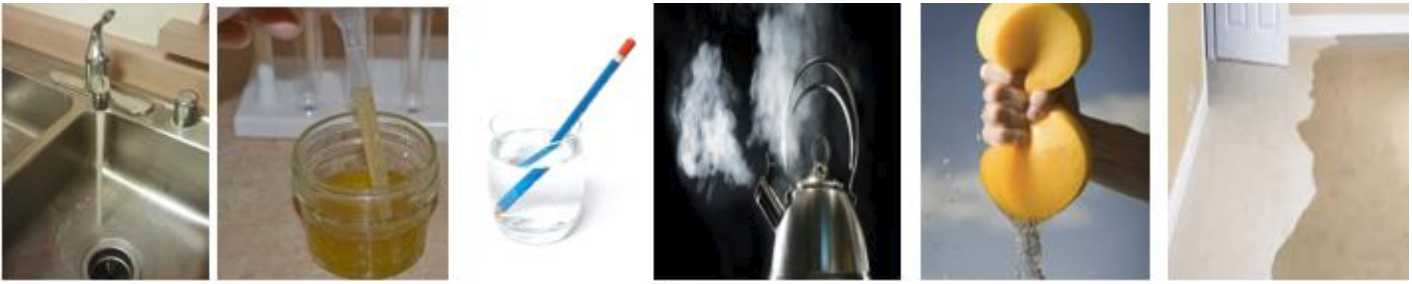}
		\caption{}
	\end{subfigure}%
	\\
	\begin{subfigure}[b]{\linewidth}
		\includegraphics[width=\linewidth]{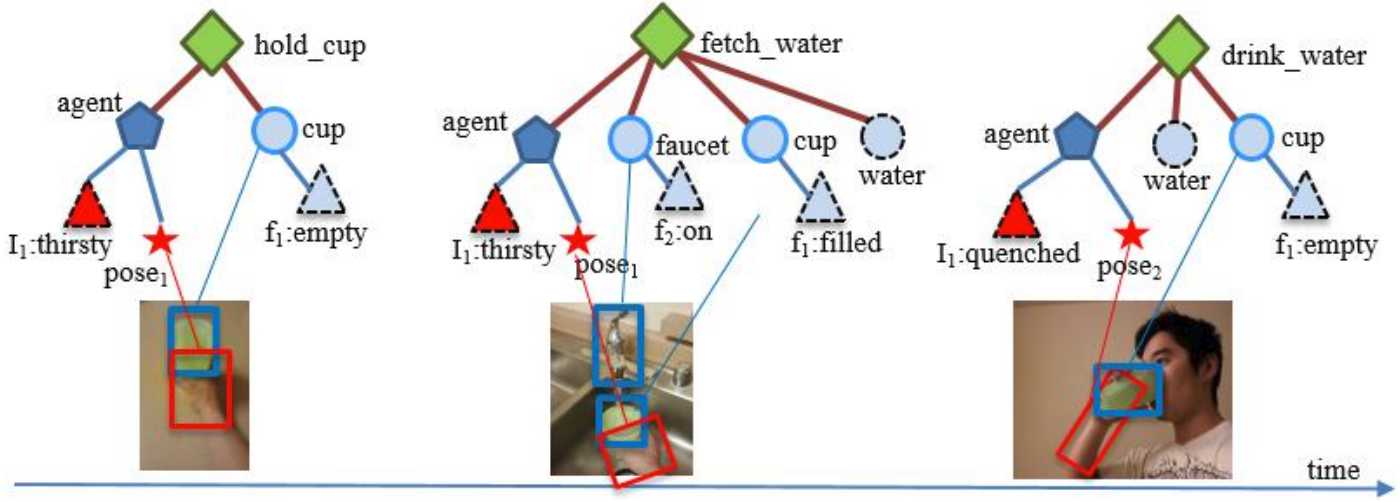}
		\caption{}
	\end{subfigure}%
	\vspace{-6pt}
	\caption{Water and other clear fluids play important roles in a human's daily life, but are barely detectable in images. (a) Water causes only minor changes in appearance;
	(b) the ``dark'' entities of water, fluents (here, a cup and faucet, represented by triangles), and the intention of a human are shown in dashed nodes. The actions (diamonds) involve agents (pentagons) and cups (objects in circles).}
	\label{fig:fluent}
\end{figure}

A \emph{fluent}, which is a concept coined and discussed by Isaac Newton~\citep{newton1736method} and Maclaurin~\cite{maclaurin1742treatise}, respectively, and adopted by \ac{ai} and commonsense reasoning~\citep{mueller2014commonsense,mueller1990daydreaming}, refers to a transient state of an object that is time-variant, such as a cup being empty or filled, a door being locked, a car blinking to signal a left turn, and a telephone ringing; see \cref{fig:fluent} for other examples of ``dark'' fluents in images. Fluents are linked to perceived causality~\citep{michotte1963perception} in the psychology literature. Even infants with limited exposure to visual experiences have the innate ability to learn causal relationships from daily observation, which leads to a sophisticated understanding of the semantics of events~\citep{carey2009origin}.

\begin{figure*}[t!]
	\centering
	\begin{subfigure}[b]{0.144\linewidth}
		\includegraphics[width=\linewidth]{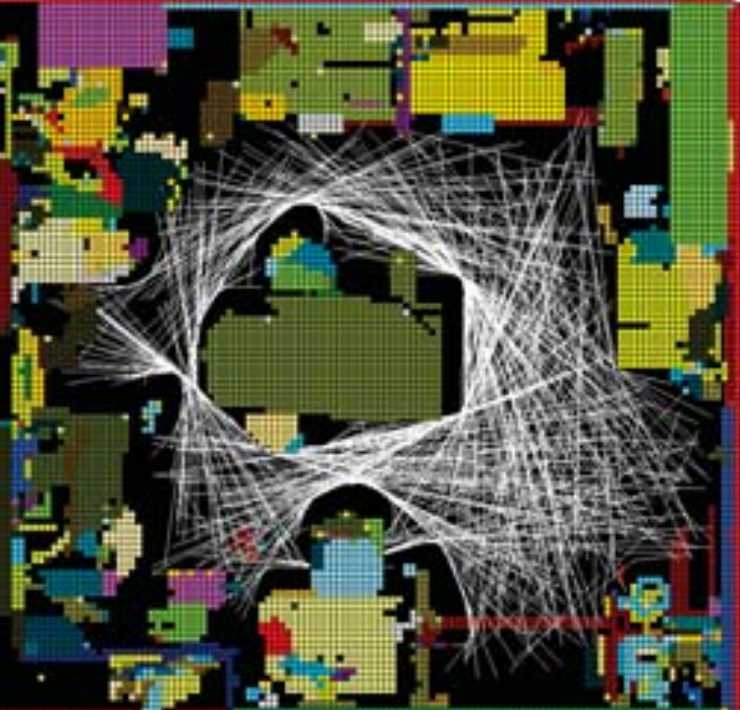}
		\caption{}
	\end{subfigure}%
	\hfill%
	\begin{subfigure}[b]{0.144\linewidth}
		\includegraphics[width=\linewidth]{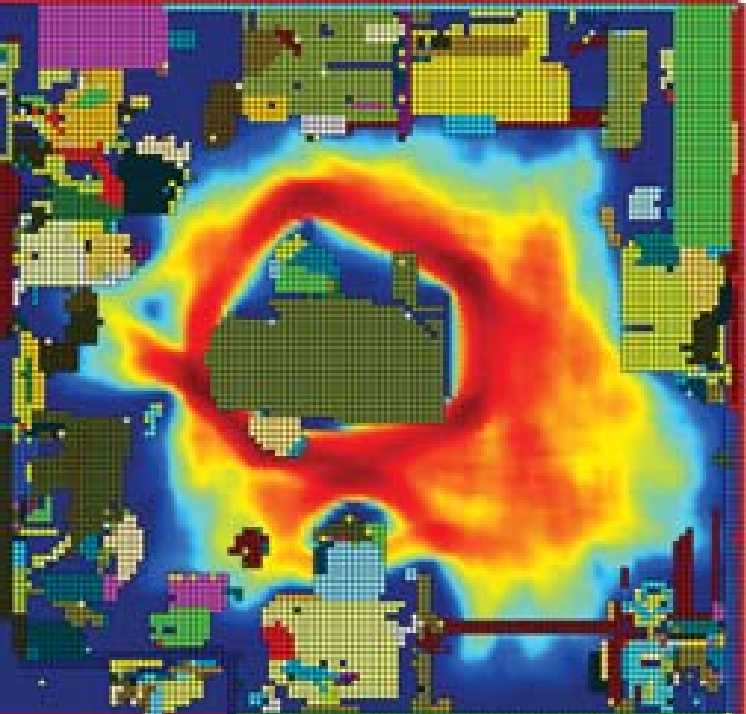}
		\caption{}
	\end{subfigure}%
	\hfill%
	\begin{subfigure}[b]{0.237\linewidth}
		\includegraphics[width=\linewidth]{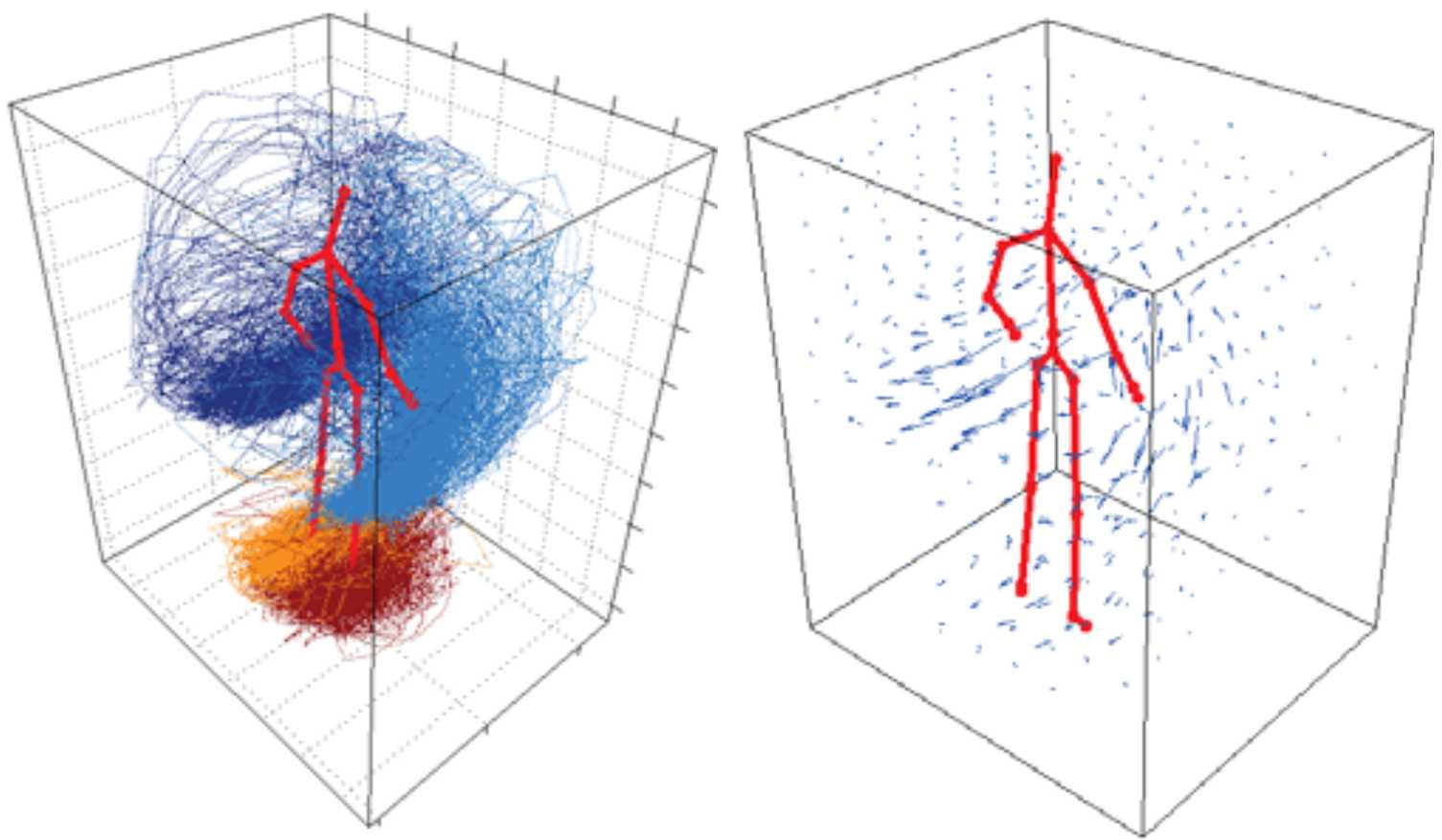}
		\caption{}
	\end{subfigure}%
	\hfill%
	\begin{subfigure}[b]{0.465\linewidth}
		\includegraphics[width=\linewidth]{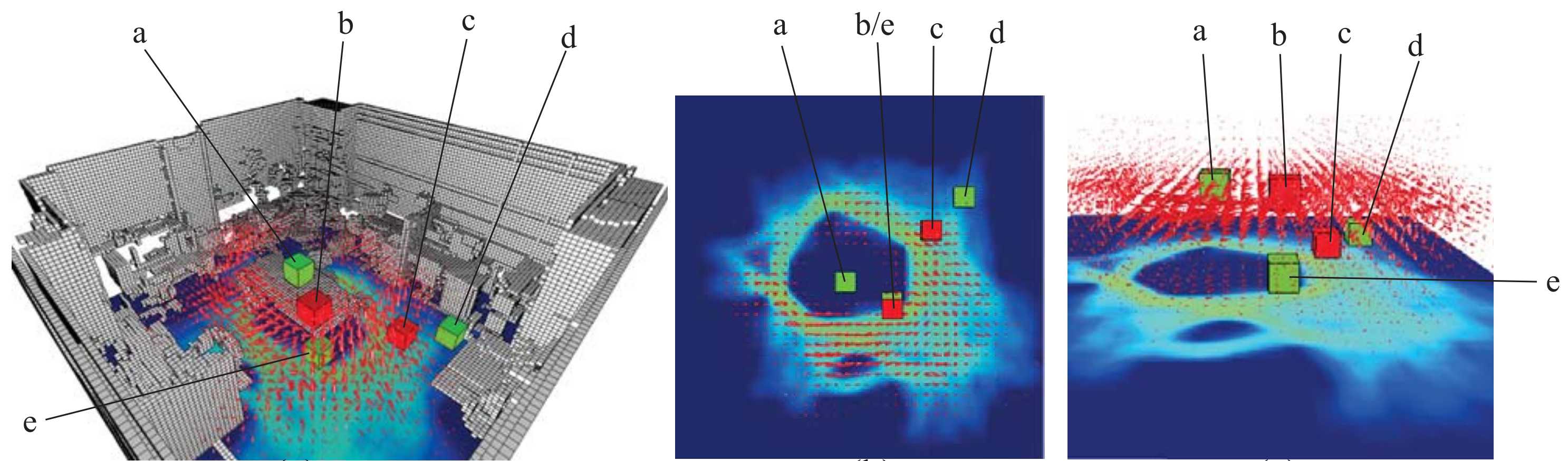}
		\caption{}
	\end{subfigure}%
	\caption{Inferring the potential for objects to fall from human actions and natural disturbances. (a) The imagined human trajectories; (b) the distribution of primary motion space; (c) the secondary motion field; (d) the integrated human action field, built by integrating primary motions with secondary motions. The five objects \textbf{a-e} are typical cases in the disturbance field: The objects \textbf{b} on the edge of a table and \textbf{c} along the pathway exhibit greater disturbance (in the form of accidental collisions) than other objects such as \textbf{a} in the center of the table, \textbf{e} below the table, and \textbf{d} in a concave corner of the room. Reproduced from Ref.~\citep{zheng2014detecting} with permission of IEEE, \textcopyright~2014.}
	\label{fig:physics}
\end{figure*}

Fluents and perceived causality are different from the visual \emph{attributes}~\citep{farhadi2009describing,parikh2011relative} of objects. The latter are permanent over the course of observation; for example, the gender of a person in a short video clip should be an attribute, not a fluent. Some fluents are visible, but many are ``dark.'' Human cognition has the innate capability (observed in infants)~\citep{carey2009origin} and strong inclination to perceive the \emph{causal effects} between \emph{actions} and \emph{changes of fluents}; for example, realizing that flipping a switch causes a light to turn on. To recognize the change in an object caused by an action, one must be able to perceive and evaluate the state of the object's changeable characteristics; thus, perceiving fluents, such as whether the light switch is set to the up or down position, is essential for recognizing actions and understanding events as they unfold. Most vision research on action recognition has paid a great deal of attention to the position, pose, and movement of the human body in the process of activities such as walking, jumping, and clapping, and to human-object interactions such as drinking and smoking~\citep{laptev2008learning,yao2009learning,yao2013animated,wang2012mining}; but most daily actions, such as opening a door, are defined by cause and effect (a door's fluent changes from ``closed'' to ``open,'' regardless of how it is opened), rather than by the human's position, movement, or spatial-temporal features~\citep{dalal2005histograms,sadanand2012action}. Similarly, actions such as putting on clothes or setting up a tent cannot be defined simply by their appearance features; their complexity demands causal reasoning to be understood. Overall, the status of a scene can be viewed as a collection of fluents that \emph{record the history of actions}. Nevertheless, fluents and causal reasoning have not yet been systematically studied in machine vision, despite their ubiquitous presence in images and videos.

\subsubsection{Intuitive Physics}

Psychology studies suggest that approximate Newtonian principles underlie human judgments about dynamics and stability~\citep{fleming2010perceived,zago2005visual}. Hamrick \etal~\citep{hamrick2011internal} and Battaglia \etal~\citep{battaglia2013simulation} showed that the knowledge of Newtonian principles and probabilistic representations is generally applied in human physical reasoning, and that an intuitive physical model is an important aspect of human-level complex scene understanding. Other studies have shown that humans are highly sensitive to whether objects in a scene violate certain understood physical relationships or appear to be physically unstable~\citep{kellman1983perception,baillargeon1985object,johnson1995perception,needham1997factors,baillargeon2004infants}.

Invisible physical fields govern the layout and placement of objects in a human-made scene. By human design, objects should be physically stable and safe with respect to gravity and various other potential disturbances~\citep{zheng2013beyond,zheng2014detecting,zheng2015scene}, such as an earthquake, a gust of wind, or the actions of other humans. Therefore, any 3D scene interpretation or parsing (\eg, object localization and segmentation) must be physically plausible~\citep{zheng2013beyond,zheng2014detecting,zheng2015scene,qi2018human,huang2018holistic,huang2018cooperative}; see \cref{fig:physics}. This observation sets useful constraints to scene understanding and is important for robotics applications~\citep{zheng2014detecting}. For example, in a search-and-rescue mission at a disaster-relief site, a robot must be able to reason about the stability of various objects, as well as about which objects are physically supporting which other objects, and then use this information to move cautiously and avoid creating dangerous new disturbances. 

\subsubsection{Functionality}

Most human-made scenes are designed to serve multiple human functions, such as sitting, eating, socializing, and sleeping, and to satisfy human needs with respect to those functions, such as illumination, temperature control, and ventilation. These functions and needs are invisible in images, but shape the scene's layout~\citep{gupta20113d,zhao2013scene}, its geometric dimensions, the shape of its objects, and the selection of its materials.

Through functional magnetic resonance imaging (fMRI) and neurophysiology experiments, researchers identified mirror neurons in the pre-motor cortical area that seem to encode actions through poses and interactions with objects and scenes~\citep{iacoboni2005grasping}. Concepts in the human mind are not only represented by prototypes---that is, exemplars as in current computer vision and machine learning approaches---but also by functionality~\citep{carey2009origin}.

\subsubsection{Intentions and Goals}

Cognitive studies~\citep{csibra2007obsessed} show that humans have a strong inclination to interpret events as a series of goals driven by the intentions of agents. Such a teleological stance inspired various models in the cognitive literature for intent estimation as an inverse planning problem~\citep{baker2007goal,baker2008theory}.

\begin{figure}[t!]
	\centering
	\includegraphics[width=\linewidth]{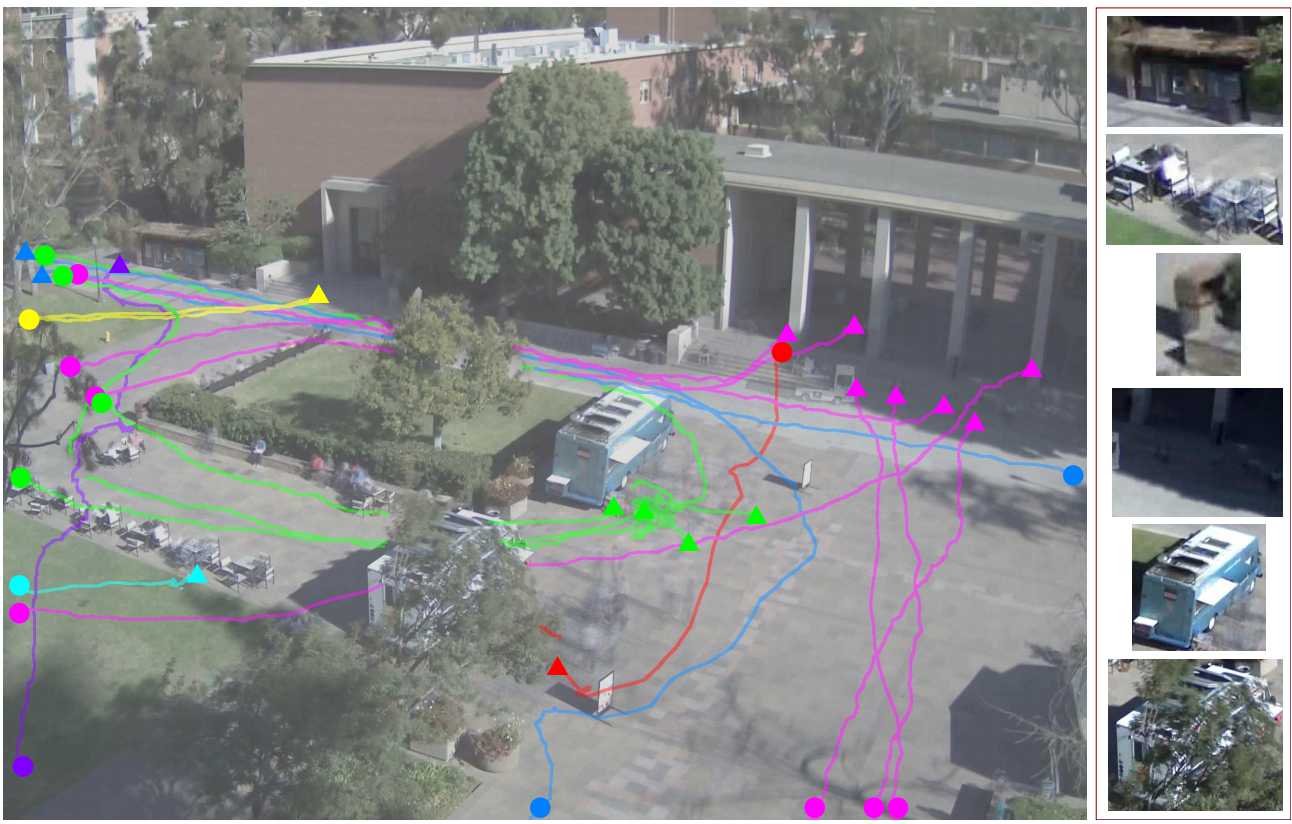}
	\caption{People's trajectories are color-coded to indicate their shared destination. The triangles denote destinations, and the dots denote start positions; \eg, people may be heading toward the food truck to buy food (green), or to the vending machine to quench thirst (blue). Due to low resolution, poor lighting, and occlusions, objects at the destinations are very difficult to detect based only on their appearance and shape. Reproduced from Ref.~\citep{xie2018learning} with permission of IEEE, \textcopyright~2018.}
	\label{fig:physical_social_foodtruck}
\end{figure}

We argue that intent can be treated as the transient status of agents (humans and animals), such as being ``thirsty,'' ``hungry,'' or ``tired.'' They are similar to, but more complex than, the fluents of objects, and come with the following characteristics: (i) They are hierarchically organized in a sequence of goals and are the main factors driving actions and events in a scene. (ii) They are completely ``dark,'' that is, not represented by pixels. (iii) Unlike the instant change of fluents in response to actions, intentions are often formed across long spatiotemporal ranges. For example, in \cref{fig:physical_social_foodtruck}~\citep{xie2018learning}, when a person is hungry and sees a food truck in the courtyard, the person decides (intends) to walk to the truck.

During this process, an attraction relationship is established at a long distance. As will be illustrated later in this paper, each functional object, such as a food truck, trashcan, or vending machine, emits a field of attraction over the scene, not much different from a gravity field or an electric field. Thus, a scene has many layers of attraction or repulsion fields (\eg, foul odor, or grass to avoid stepping on), which are completely ``dark.'' The trajectory of a person with a certain intention moving through these fields follows a least-action principle in Lagrange mechanics that derives all motion equations by minimizing the potential and kinematic energies integrated over time.

Reasoning about intentions and goals will be crucial for the following vision and cognition tasks: (i) early event and trajectory prediction~\citep{hoai2014max}; (ii) discovery of the invisible attractive/repulsive fields of objects and recognizing their functions by analyzing human trajectories~\citep{xie2018learning}; (iii) understanding of scenes by function and activity~\citep{qi2017predicting}, where the attraction fields are longer range in a scene than the functionality maps~\citep{pei2011parsing,turek2010unsupervised} and affordance maps~\citep{grabner2011makes,jia20133d,jiang2013hallucinated} studied in recent literature; (iv) understanding multifaceted relationships among a group of people and their functional roles~\citep{shu2016critical,shu2018perception,shu2019partitioning}; and (v) understanding and inferring the mental states of agents~\citep{baker2011bayesian,zhao2015represent}.

\subsubsection{Utility and Preference}

Given an image or a video in which agents are interacting with a 3D scene, we can mostly assume that the observed agents make near-optimal choices to minimize the cost of certain tasks; that is, we can assume there is no deception or pretense. This is known as the rational choice theory; that is, a rational person's behavior and decision-making are driven by maximizing their utility function. In the field of mechanism design in economics and game theory, this is related to the revelation principle, in which we assume that each agent \emph{truthfully} reports its preferences; see Ref.~\citep{nisan2001algorithmic} for a short introductory survey. Building computational models for human utility can be traced back to the English philosopher Jeremy Bentham, and to his works on ethics known as utilitarianism~\citep{bentham1789introduction}. 

By observing a rational person's behavior and choices, it is possible to reverse-engineer their reasoning and learning process, and estimate their values. Utility, or values, are also used in the field of \ac{ai} in planning schemes such as the Markov decision process (MDP), and are often associated with the states of a task. However, in the literature of the MDP, ``value'' is not a reflection of true human preference and, inconveniently, is tightly dependent on the agent's actions~\citep{shukla2019utility}. We argue that such utility-driven learning could be more invariant than traditional supervised training for computer vision and \ac{ai}.

\subsubsection{Summary}

Despite their apparent differences at first glance, the five FPICU domains interconnect in ways that are theoretically important. These interconnections include the following characteristics: (i) The five FPICU domains usually do not easily project onto explicit visual features; (ii) most of the existing computer vision and \ac{ai} algorithms are neither competent in these domains nor (in most cases) applicable at all; and (iii) human vision is nevertheless highly efficient in these domains, and human-level reasoning often builds upon prior knowledge and capability with FPICU.

We argue that the incorporation of these five key elements would advance a vision or \ac{ai} system in at least three aspects:
\begin{enumerate}[leftmargin=*,noitemsep,nolistsep]
	\item Generalization. As a higher level representation, the FPICU concept tends to be globally invariant across the entire human living space. Therefore, knowledge learned in one scene can be transferred to novel situations.
	\item Small sample learning. FPICU encodes essential prior knowledge for understanding the environment, events, and behavior of agents. As FPICU is more invariant than appearance or geometric features, the learning of FPICU, which is more consistent and noise-free across different domains and data sources, is possible even without ``big data.''
	\item Bidirectional inference. Inference with FPICU requires the combination of top-down inference based on abstract knowledge and bottom-up inference based on visual pattern. This means that systems would both continue to make data-driven inferences from the observation of visible, pixel-represented scene aspects, as they do today, and make inferences based on FPICU understanding. These two processes can feed on each other, boosting overall system performance.
\end{enumerate}

In the following sections, we discuss these five key elements in greater detail.

\section{Causal Perception and Reasoning: The Basis for Understanding}\label{sec:causal}

\begin{figure}[t!]
	\centering
	\includegraphics[width=\linewidth]{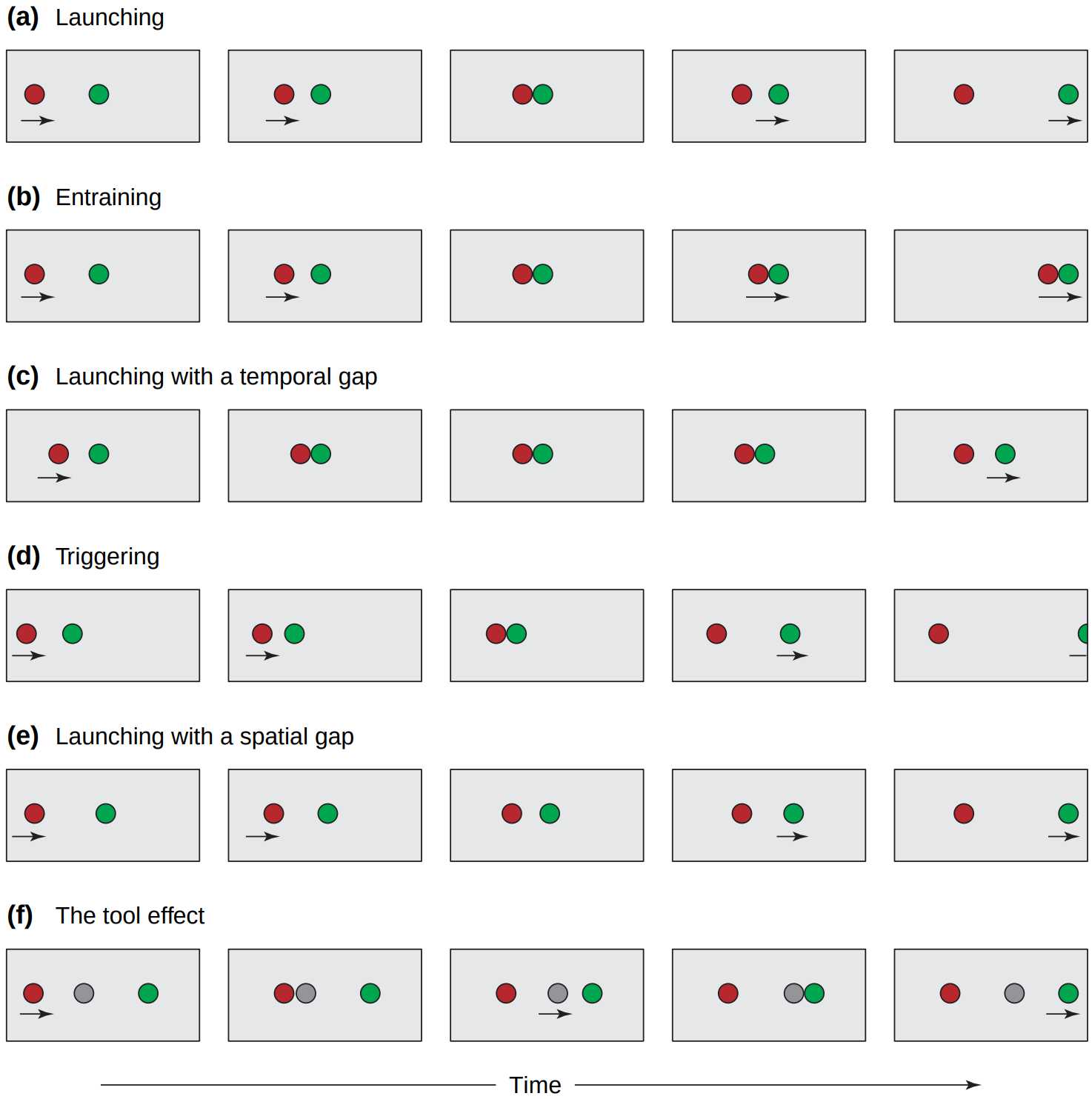}
	\caption{Examples of some of Michotte's basic demonstrations of perceptual causality, regarding the perception of two objects, A and B (here shown as red and green circles, respectively). (a) The launching effect; (b) the entraining effect, wherein A seems to carry B along with it; (c) the launching effect is eliminated by adding a temporal gap between A's and B's motions; (d) the triggering effect, wherein B's motion is seen as autonomous, despite still being caused by A; (e) the launching effect is also eliminated by adding a spatial gap between A's final position and B's initial position; (f) the tool effect, wherein an intermediate item (gray circle) seems merely a tool by which A causes the entire motion sequence. These are some of the many cause-effect relationships between objects that humans understand intuitively, and that \ac{ai} must learn to recognize. Reproduced from Ref.~\citep{scholl2000perceptual} with permission of Elsevier Science Ltd., \textcopyright~2000.}
	\label{fig:perceptual_causality_scholl}
\end{figure}

Causality is the abstract notion of cause and effect derived from our perceived environment, and thus can be used as a prior foundation to construct notions of time and space~\citep{robb1911optical,malament1977class,robb2014geometry}. People have innate assumptions about causes, and causal reasoning can be activated almost automatically and irresistibly~\citep{corrigan1996causal,white1988causal}. In our opinion, causality is the foundation of the other four FPICU elements (functionality, physics, intent, and utility). For example, an agent must be able to reason about the causes of others' behavior in order to understand their intent and understand the likely effects of their own actions to use functional objects appropriately. To a certain degree, much of human understanding depends on the ability to comprehend causality. Without understanding what causes an action, it is very difficult to consider what may happen next and respond effectively.

In this section, we start with a brief review of the causal perception and reasoning literature in psychology, followed by a review of a parallel stream of work in statistical learning. We conclude the section with case studies of causal learning in computer vision and \ac{ai}.

\subsection{Human Causal Perception and Reasoning}

Humans reason about causal relationships through high-level cognitive reasoning. But can we ``see'' causality directly from vision, just as we see color and depth? In a series of behavioral experiments, Chen and Scholl~\citep{chen2016perception} showed that the human visual system can \emph{perceive} causal history through commonsense visual reasoning, and can represent objects in terms of their inferred underlying causal history---essentially representing shapes by wondering about ``how they got to be that way.'' Inherently, causal events cannot be directly interpreted merely from vision; they must be interpreted by an agent that understands the distal world~\citep{holyoak2011causal}.

Early psychological work focused on an associative mechanism as the basis for human causal learning and reasoning~\citep{shanks1988associative}. During this time, the Rescorla-Wagner model was used to explain how humans (and animals) build expectations using the cooccurrence of perceptual stimuli~\citep{rescorla1972theory}. However, more recent studies have shown that human causal learning is a rational Bayesian process~\citep{holyoak2011causal,lu2008bayesian,edmonds2019decomposing} involving the acquisition of \emph{abstract} causal structure~\citep{waldmann1992predictive,edmonds2018human} and strength values for cause-effect relationships~\citep{cheng1997covariation}.

The perception of causality was first systematically studied by the psychologist Michotte~\citep{michotte1963perception} through observation of one billiard ball (A) hitting another (B); see \cref{fig:perceptual_causality_scholl}~\citep{michotte1963perception} for a detailed illustration. In the classic demonstration, Ball A stops the moment it touches B, and B immediately starts to move, at the \emph{same} speed A had been traveling. This visual display describes not only kinematic motions, but a causal interaction in which A ``launches'' B. Perception of this ``launching effect'' has a few notable properties that we enumerate below; see Ref.~\citep{scholl2000perceptual} for a more detailed review.
\begin{enumerate}[leftmargin=*,noitemsep,nolistsep]
	\item Irresistibility: Even if one is told explicitly that A and B are just patches of pixels that are incapable of mechanical interactions, one is still compelled to perceive launching. One cannot stop seeing salient causality, just as one cannot stop seeing color and depth.
	\item Tight control by spatial-temporal patterns of motion: By adding even a small temporal gap between the stop of A and the motion of B, perception of the launching effect will break down; instead, B's motion will be perceived as self-propelled.
	\item Richness: Even the interaction of only two balls can support a variety of causal effects. For example, if B moves with a speed \emph{faster} (vs. the same) than that of A, then the perception would not be that A ``triggers'' B's motion. Perceptual causality also includes ``entraining,'' which is superficially identical to launching, except that A \emph{continues} to move along with B after they make contact. 
\end{enumerate}

Recent cognitive science studies~\citep{rolfs2013visual} provide still more striking evidence of how deeply human vision is rooted in causality, making the comparison between color and causality still more profound. In human vision science, ``adaptation'' is a phenomenon in which an observer adapts to stimuli after a period of sustained viewing, such that their perceptual response to those stimuli becomes weaker. In a particular type of adaptation, the stimuli must appear in the same retinotopic position, defined by the reference frame shared by the retina and visual cortex. This type of retinotopic adaptation has been taken as strong evidence of early visual processing of that stimuli. For example, it is well-known that the perception of color can induce retinotopic adaptation~\citep{mccollough1965color}. Strikingly, recent evidence revealed that retinotopic adaptation also takes place for the perception of causality. After prolonged viewing of the launching effect, subsequently viewed displays were judged more often as non-causal only if the displays were located within the same retinotopic coordinates. This means that physical causality is extracted during early visual processing. By using retinotopic adaptation as a tool, Kominsky and Scholl~\citep{kominksy2018retinotopically} recently explored whether launching is a fundamentally different category from \emph{entraining}, in which Ball A moves together with Ball B after contact. The results showed that retinotopically specific adaptation did not transfer between launching and entraining, indicating that there are indeed fundamentally distinct categories of causal perception in vision.

The importance of causal perception goes beyond placing labels on different causal events. One unique function of causality is the support of counterfactual reasoning. Observers recruit their counterfactual reasoning capacity to interpret visual events. In other words, interpretation is not based only on what is observed, but also on what would have happened but did not. In one study~\citep{gerstenberg2017eye}, participants judged whether one billiard ball caused another to go or prevented it from going through a gate. The participants' viewing patterns and judgments demonstrated that the participants simulated where the target ball would have gone if the candidate cause had been removed from the scene. The more certain participants were that the outcome would have been different, the stronger the causal judgments. These results clearly demonstrated that spontaneous counterfactual simulation plays a critical role in scene understanding.

\subsection{Causal Transfer: Challenges for Machine Intelligence}

Despite all the above evidence demonstrating the important and unique role of causality in human vision, there remains much debate in the literature as to whether causal relationship understanding is necessary for high-level machine intelligence. However, learning causal concepts is of the utmost importance to agents that are expected to operate in observationally varying domains with common latent dynamics. To make this concrete, our environment on Earth adheres to relatively constant environmental dynamics, such as constant gravity. Perhaps more importantly, much of our world is \emph{designed} by other humans and largely adheres to common causal concepts: Switches turn things off and on, knobs turn to open doors, and so forth. Even though objects in different settings appear different, their causal effect is constant because they all fit and cohere to a consistent causal design. Thus, for agents expected to work in varying but human-designed environments, the ability to learn generalizable and transferable causal understanding is crucial.

\begin{figure}[t!]
	\centering
	\begin{subfigure}[b]{0.61\linewidth}
	\includegraphics[width=\linewidth,trim={1cm 1cm .45cm 0.4cm},clip]{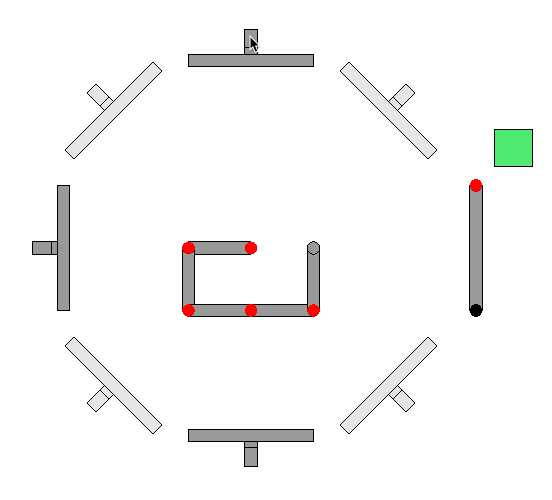}
	\caption{}
	\end{subfigure}%
	\begin{subfigure}[b]{0.39\linewidth}
	\includegraphics[width=\linewidth]{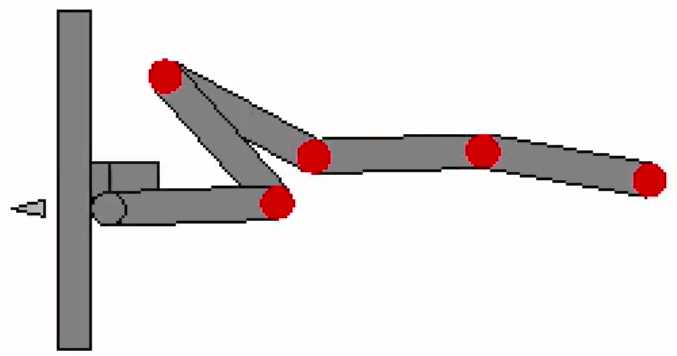}
	\caption{}%
	\includegraphics[width=\linewidth]{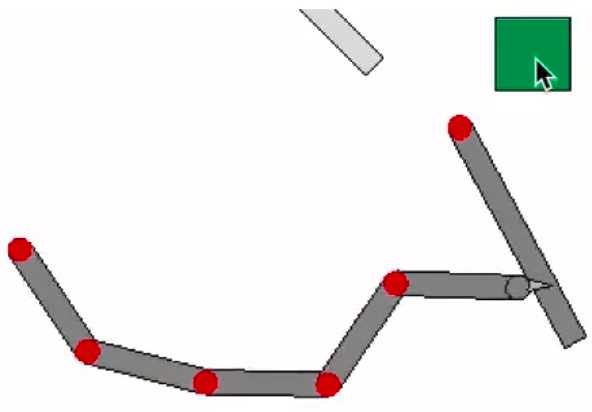}
	\caption{}
	\end{subfigure}%
	\caption{The OpenLock task presented in Ref.~\citep{edmonds2018human}. (a) Starting configuration of a three-lever trial. All levers are being pulled toward the robot arm, whose base is anchored to the center of the display. The arm interacts with levers by either \emph{pushing} outward or \emph{pulling} inward. This is achieved by clicking either the outer or inner regions of the levers' radial tracks, respectively. Only push actions are needed to unlock the door in each lock situation. Light gray levers are always locked, which is unknown to both human subjects and \acf{rl}-trained agents at the beginning of training. Once the door is unlocked, the green button can be clicked to command the arm to push the door open. The black circle located opposite the door's red hinge represents the door lock indicator: present if locked, absent if unlocked. (b) Pushing a lever. (c) Opening the door by clicking the green button}
	\label{fig:openlock_task}
\end{figure}

Recent successes of systems such as deep \acf{rl} showcase a broad range of applications~\citep{mnih2015human,schulman2015trust,silver2016mastering,levine2016end,schulman2017proximal}, the vast majority of which do not learn explicit causal relationships. This results in a significant challenge for transfer learning under today's dominant machine learning paradigm~\citep{zhang2018study,kansky2017schema}. One approach to solving this challenge is to learn a causal encoding of the environment, because causal knowledge inherently encodes a transferable representation of the world. Assuming the dynamics of the world are constant, causal relationships will remain true regardless of observational changes to the environment (\eg, changing an object's color, shape, or position).

In a study, Edmonds \etal~\citep{edmonds2018human} presented a complex hierarchical task that requires humans to reason about abstract causal structure. The work proposed a set of virtual ``escape rooms,'' where agents must manipulate a series of levers to open a door; see an example in \cref{fig:openlock_task}~\citep{edmonds2018human}. Critically, this task is designed to force agents to form a causal structure by requiring agents to find \emph{all} the ways to escape the room, rather than just one. The work used three- and four-lever rooms and two causal structures: \acf{cc} and \acf{ce}. These causal structures encode different combinations into the rooms' locks.

\begin{figure}[t!]
	\centering
	\begin{subfigure}[b]{\linewidth}
	\includegraphics[width=\linewidth]{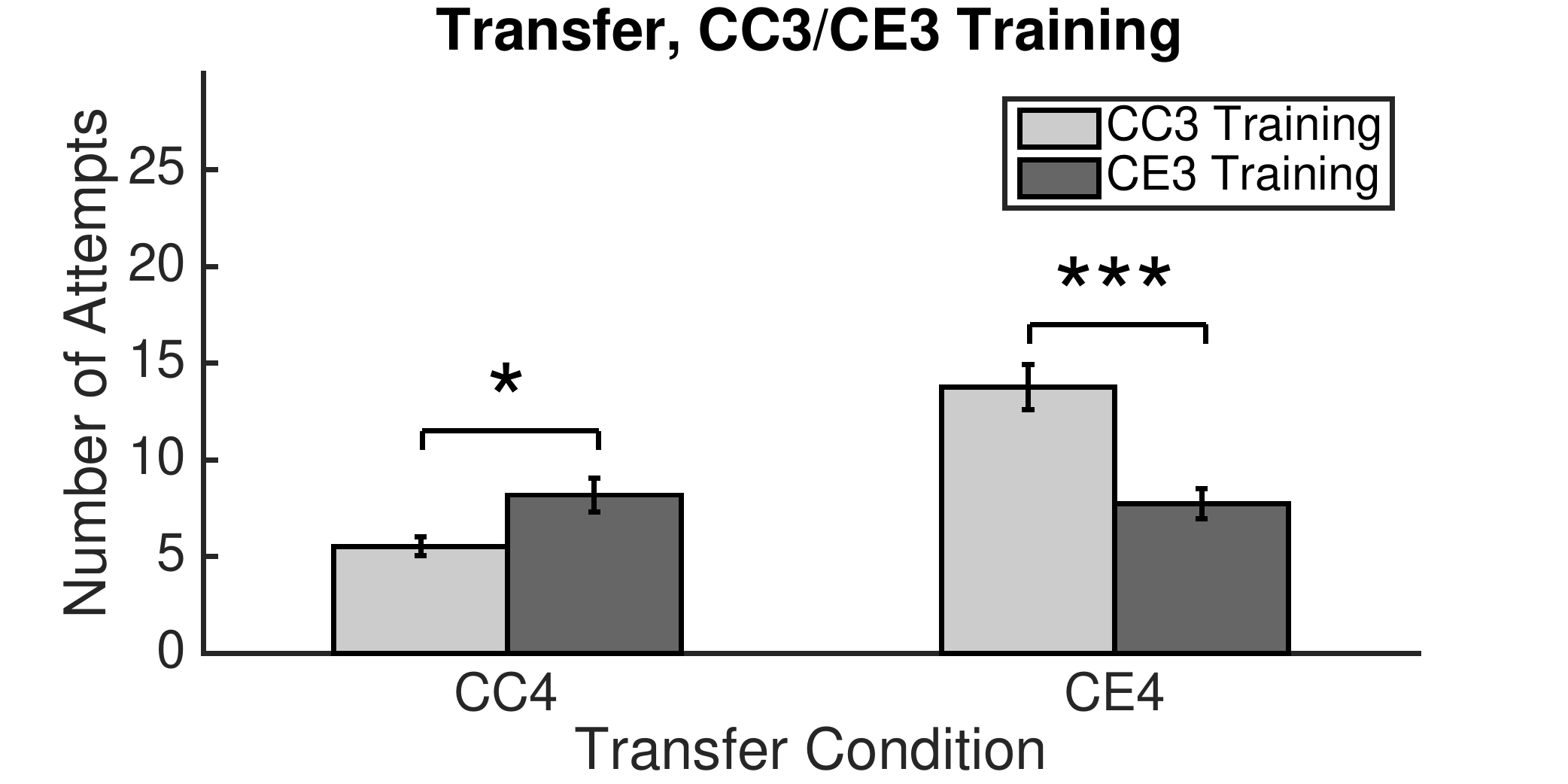}
	\caption{Transfer trial results of human participants.}
	\end{subfigure}%
	\\
	\begin{subfigure}[b]{\linewidth}
	\includegraphics[width=\linewidth]{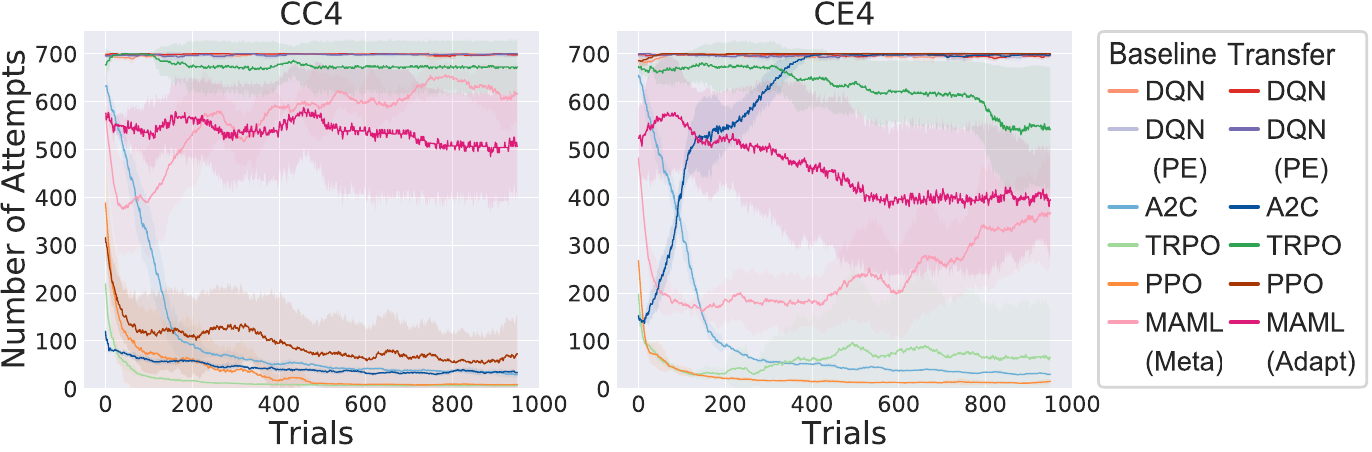}
	\caption{Transfer trial results of \ac{rl} agents.}
	\end{subfigure}%
	\caption{Comparisons between human causal learners and typical \ac{rl} agents~\citep{edmonds2020theory}. \acf{cc4} and \acf{ce4} denote two transfer conditions used by Edmonds \etal~\citep{edmonds2018human}. (a) Average number of attempts human participants needed to find all unique solutions under four-lever Common Cause (\ac{cc4}; left) and Common Effect (\ac{ce4}; right) conditions, showing a positive causal transfer after learning. Light and dark gray bars indicate \acf{cc3} and \acf{ce3} training, respectively. Error bars indicate standard error of the mean. (b) In contrast, \ac{rl} agents have difficulties transferring learned knowledge to solve similar tasks. Baseline (no transfer) results show that the best-performing algorithms (\ac{ppo}, \ac{trpo}) achieve success in 10 and 25 attempts by the end of the baseline training for \ac{cc4} and \ac{ce4}, respectively. \ac{a2c} is the only algorithm to show positive transfer; \ac{a2c} performed better with training for the \ac{cc4} condition. DQN: deep Q-network; DQN (PE): deep Q-network with prioritized experience replay; MAML: model-agnostic meta-learning.}
	\label{fig:openlock_transfer}
\end{figure}

After completing a single room, agents are then placed into a room where the perceived environment has been changed, but the underlying abstract, latent causal structure remains the same. In order to reuse the causal structure information acquired in the previous room, the agent needs to learn the relationship between its perception of the new environment and the same latent causal structure on the fly. Finally, at the end of the experiment, agents are placed in a room with one additional lever; this new room may follow the same (congruent) or different (incongruent) underlying causal structures, to test whether the agent can generalize its acquired knowledge to more complex circumstances.

This task setting is unique and challenging for two major reasons: (i) transferring agents between rooms tests whether or not agents form \emph{abstract} representations of the environment; and (ii) transferring between three- and four-lever rooms examines how well agents are able to adapt causal knowledge to similar but different causal circumstances.

In this environment, human subjects show a remarkable ability to acquire and transfer knowledge under observationally different but structurally equivalent causal circumstances; see comparisons in \cref{fig:openlock_transfer}~\citep{edmonds2019decomposing,edmonds2020theory}. Humans approached optimal performance and showed positive transfer effects in rooms with an additional lever in both congruent and incongruent conditions. In contrast, recent deep \ac{rl} methods failed to account for necessary causal abstraction, and showed a negative transfer effect. These results suggest that systems operating under current machine learning paradigms cannot learn a proper abstract encoding of the environment; that is, they do not learn an abstract causal encoding. Thus, we treat learning causal understanding from perception and interaction as one type of ``dark matter'' facing current \ac{ai} systems, which should be explored further in future work.

\subsection{Causality in Statistical Learning}

Rubin~\citep{rubin1974estimating} laid the foundation for causal analysis in statistical learning in his seminal paper, ``Estimating causal effects of treatments in randomized and nonrandomized studies;'' see also Ref.~\citep{imbens2015causal}. The formulation this work demonstrated is commonly called the Rubin causal model. The key concept in the Rubin causal model is potential outcomes. In the simplest scenario, where there are two treatments for each subject (\eg, smoking or not smoking), the causal effect is defined as the difference between potential outcomes under the two treatments. The difficulty with causal inference is that, for each subject, we only observe the outcome under the one treatment that is actually assigned to the subject; the potential outcome, if the other treatment had been assigned to that subject, is missing. If the assignment of the treatment to each subject depends on the potential outcomes under the two treatments, a naive analysis comparing the observed average outcomes of the treatments that are actually assigned to the subjects will result in misleading conclusions. A common manifestation of this problem is the latent variables that influence both the treatment assignment and the potential outcomes (\eg, a genetic factor influencing both one's tendency to smoke and one's health). A large body of research has been developed to solve this problem. A very prominent example is the propensity score~\citep{rosenbaum1983central}, which is the conditional probability of assigning one treatment to a subject given the background variables of the subject. Valid causal inference is possible by comparing subjects with similar propensity scores.

Causality was further developed in Pearl's probabilistic graphical model (\ie, causal Bayesian networks (CBNs))~\citep{pearl2000causality}. CBNs enabled economists and epidemiologists to make inferences for quantities that cannot be intervened upon in the real world. Under this framework, an expert modeler typically provides the structure of the CBN. The parameters of the model are either provided by the expert or learned from data, given the structure. Inferences are made in the model using the $do$ operator, which allows modelers to answer the question, \emph{if $X$ is intervened and set to a particular value, how is $Y$ affected}? Concurrently, researchers embarked on a quest to recover causal relationships from observational data~\citep{spirtes2000causation}. These efforts tried to determine under what circumstances the structure (presence and direction of an edge between two variables in CBN) could be determined from purely observational data~\citep{spirtes2000causation,chickering2002optimal,peters2014causal}.

This framework is a powerful tool in fields where real-world interventions are difficult (if not impossible)---such as economics and epidemiology---but lacks many properties necessary for humanlike \ac{ai}. First, despite attempts to learn causal structure from observational data, most structure learning approaches cannot typically succeed beyond identifying a Markov equivalence class of possible structures~\citep{peters2014causal}; therefore, structure learning remains an unsolved problem. Recent work has attempted to tackle this limitation by introducing \emph{active intervention} that enables agents to explore possible directions of undirected causal edges~\citep{he2008active,bramley2017formalizing}. However, the space of possible structures and parameters is exponential, which has limited the application of CBNs to cases with only a handful of variables. This difficulty is partially due to the strict formalism imposed by CBNs, where all possible relationships must be considered. Humanlike \ac{ai} should have the ability to constrain the space of possible relationships to what is heuristically ``reasonable'' given the agent's understanding of the world, while acknowledging that such a learning process may not result in the ground-truth causal model. That is, we suggest that for building humanlike \ac{ai}, learners should relax the formalism imposed by CBNs to accommodate significantly more variables without disregarding explicit causal structure (as is currently done by nearly all deep learning models). To make up for this approximation, learners should be in a constant state of active and interventional learning, where their internal causal world model is updated with new confirming or contradictory evidence.

\begin{figure*}[t!]
	\centering
	\includegraphics[width=\linewidth]{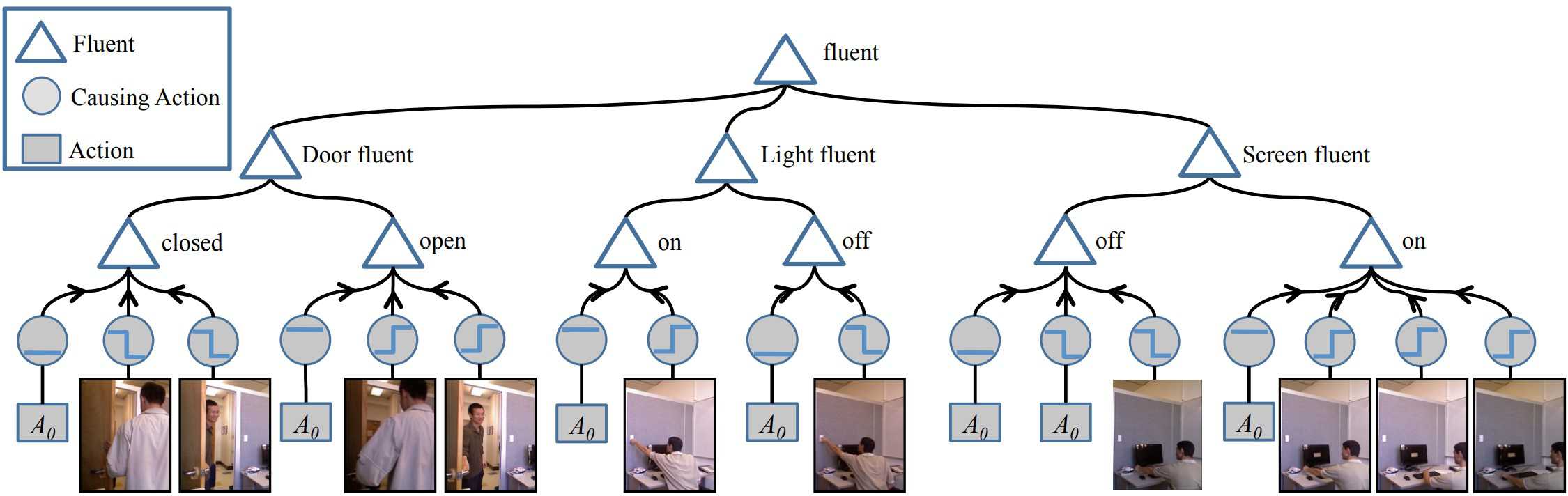}
	\caption{An example of perceptual causality in computer vision~\citep{fire2016learning}, with a causal and-or graph for door status, light status, and screen status. Action $A_0$ represents non-action (a lack of state-changing agent action). Non-action is also used to explain the change of the monitor status to off when the screensaver activates. Arrows point from causes to effects, and undirected lines show deterministic definition.}
	\label{fig:perceptual_causality}
\end{figure*}

\subsection{Causality in Computer Vision}

The classical and scientific clinical setting for learning causality is Fisher's randomized controlled experiments~\citep{fisher1937design}. Under this paradigm, experimenters control as many confounding factors as possible to tightly restrict their assessment of a causal relationship. While useful for formal science, it provides a stark contrast to the human ability to perceive causal relationships from observations alone~\citep{scholl2000perceptual,shanks1988associative,rescorla1972theory}. These works suggest that human causal perception is less rigorous than formal science but still maintains effectiveness in learning and understanding of daily events.

Accordingly, computer vision and \ac{ai} approaches should focus on how humans perceive causal relationships from observational data. Fire and Zhu~\citep{fire2013using,fire2016learning} proposed a method to learn ``dark'' causal relationships from image and video inputs, as illustrated in \cref{fig:perceptual_causality}~\citep{fire2013using}; in this study, systems learn how the status of a door, light, and screen relate to human actions. Their method achieves this iteratively by asking the same question at different intervals: \emph{given the observed videos and the current causal model, what causal relationship should be added to the model to best match the observed statistics describing the causal events?} To answer this question, the method utilizes the information projection framework~\citep{zhu1997minimax}, maximizing the amount of information gain after adding a causal relation, and then minimizing the divergence between the model and observed statistics.

This method was tested on video datasets consisting of scenes from everyday life: opening doors, refilling water, turning on lights, working at a computer, and so forth. Under the information projection framework, the top-scoring causal relationships consistently matched what humans perceived to be a cause of action in the scene, while low-scoring causal relations matched what humans perceived to \emph{not} be a cause of action in the scene. These results indicate that the information projection framework is capable of capturing the same judgments made by human causal learners. While computer vision approaches are ultimately observational methods and therefore are not guaranteed to uncover the complete and true causal structure, perceptual causality provides a mechanism to achieve humanlike learning from observational data.

Causality is crucial for humans' understanding and reasoning about videos, such as tracking humans that are interacting with objects whose visibility might vary over time. Xu \etal~\citep{xu2018causal} used a Causal And-Or Graph (C-AOG) model to tackle this kind of ``visibility fluent reasoning'' problem. They consider the visibility status of an object as a fluent variable, whose change is mostly attributed to its interaction with its surroundings, such as crossing behind another object, entering a building, or getting into a vehicle. The proposed C-AOG can represent the cause-effect relationship between an object's activities and its visibility fluent; based on this, the researchers developed a probabilistic graphical model to jointly reason about the visibility fluent change and track humans. Experimental results demonstrate that with causal reasoning, they can recover and describe complete trajectories of humans interacting frequently in complicated scenarios. Xiong \etal~\citep{xiong2016robot} also defined causality as a fluent change due to relevant action, and used a C-AOG to describe the causal understanding demonstrated by robots that successfully folded clothes after observing humans doing the same.

\begin{figure*}[t!]
	\centering
	\begin{subfigure}[t]{0.43\linewidth}
	\includegraphics[width=\linewidth]{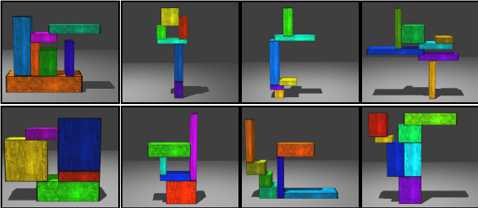}
	\caption{Will it fall?}
	\end{subfigure}%
	\begin{subfigure}[t]{0.246\linewidth}
	\includegraphics[width=\linewidth]{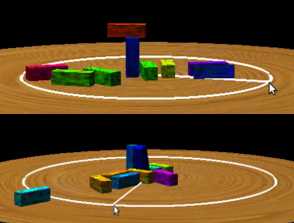}
	\caption{In which direction?}
	\end{subfigure}%
	\begin{subfigure}[t]{0.324\linewidth}
	\includegraphics[width=\linewidth]{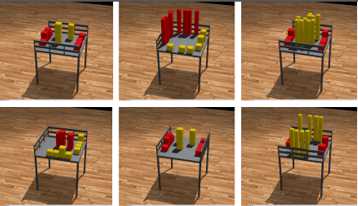}
	\caption{Which is more likely to fall if the table was bumped hard enough, the yellow or the red?}
	\end{subfigure}%
	\caption{Sample tasks of dynamic scene inferences about physics, stability, and support relationships presented in Ref.~\citep{battaglia2013simulation}: Across a variety of tasks, the intuitive physics engine accounted well for diverse physical judgments in \emph{novel} scenes, even in the presence of varying object properties and unknown external forces that could perturb the environment. This finding supports the hypothesis that human judgment of physics can be viewed as a form of probabilistic inference over the principles of Newtonian mechanics.}
	\label{fig:falling_tower}
\end{figure*}

\section{Intuitive Physics: Cues of the Physical World}\label{sec:physics}

Perceiving causality, and using this perception to interact with an environment, requires a commonsense understanding of how the world operates at a physical level. Physical understanding does not necessarily require us to precisely or explicitly invoke Newton's laws of mechanics; instead, we rely on intuition, built up through interactions with the surrounding environment. Humans excel at understanding their physical environment and interacting with objects undergoing dynamic state changes, making approximate predictions from observed events. The knowledge underlying such activities is termed \emph{intuitive physics}~\citep{mccloskey1983intuitive}. The field of intuitive physics has been explored for several decades in cognitive science and was recently reinvigorated by new techniques linked to \ac{ai}.

Surprisingly, humans develop physical intuition at an early age~\citep{carey2009origin}, well before most other types of high-level reasoning, suggesting the importance of intuitive physics in comprehending and interacting with the physical world. The fact that physical understanding is rooted in visual processing makes visual task completion an important goal for future machine vision and \ac{ai} systems. We begin this section with a short review of intuitive physics in human cognition, followed by a review of recent developments in computer vision and \ac{ai} that use physics-based simulation and physical constraints for image and scene understanding.

\subsection{Intuitive Physics in Human Cognition}

Early research in intuitive physics provides several examples of situations in which humans demonstrate common misconceptions about how objects in the environment behave. For example, several studies found that humans exhibit striking deviations from Newtonian physical principles when asked to explicitly reason about the expected continuation of a dynamic event based on a static image representing the situation at a single point in time~\citep{mccloskey1980curvilinear,mccloskey1983intuitive,disessa1982unlearning}. However, humans' intuitive understanding of physics was shown later to be much more accurate, rich, and sophisticated than previously expected once \emph{dynamics} and proper \emph{context} were provided~\citep{kaiser1986intuitive,smith2013consistent,kaiser1992influence,kaiser1985judgments,kim1999perception}.

These later findings are fundamentally different from prior work that systematically investigated the development of infants' physical knowledge~\citep{piaget1952origins,piaget1954construction} in the 1950s. The reason for such a difference in findings is that the earlier research included not only tasks of merely reasoning about physical knowledge, but also other tasks~\citep{hespos2006decalage,hespos2008young}. To address such difficulties, researchers have developed alternative experimental approaches~\citep{bower1974development,baillargeon1985object,leslie1987six,luo2003reasoning} to study the development of infants' physical knowledge. The most widely used approach is the violation-of-expectation method, in which infants see two test events: an expected event, consistent with the expectation shown, and an unexpected event, violating the expectation. A series of these kinds of studies have provided strong evidence that humans---even young infants---possess expectations about a variety of physical events~\citep{baillargeon2008account,baillargeon2002acquisition}.

\setstretch{0.96}

In a single glance, humans can perceive whether a stack of dishes will topple, whether a branch will support a child's weight, whether a tool can be lifted, and whether an object can be caught or dodged. In these complex and dynamic events, the ability to perceive, predict, and therefore appropriately interact with objects in the physical world relies on rapid physical inference about the environment. Hence, intuitive physics is a core component of human commonsense knowledge and enables a wide range of object and scene understanding.

In an early work, Achinstein~\citep{achinstein1983nature} argued that the brain builds mental models to support inference through mental simulations, analogous to how engineers use simulations for the prediction and manipulation of complex physical systems (\eg, analyzing the stability and failure modes of a bridge design before construction). This argument is supported by a recent brain imaging study~\citep{fischer2016functional} suggesting that systematic parietal and frontal regions are engaged when humans perform physical inferences even when simply viewing physically rich scenes. These findings suggest that these brain regions use a generalized mental engine for intuitive physical inference---that is, the brain's ``physics engine.'' These brain regions are much more active when making physical inferences relative to when making inferences about \emph{nonphysical} but otherwise highly similar scenes and tasks. Importantly, these regions are not exclusively engaged in physical inference, but are also overlapped with the brain regions involved in action planning and tool use. This indicates a very intimate relationship between the cognitive and neural mechanisms for understanding intuitive physics, and the mechanisms for preparing appropriate actions. This, in turn, is a critical component linking perception to action.

To construct humanlike commonsense knowledge, a computational model for intuitive physics that can support the performance of \emph{any} task that involves physics, not just one narrow task, must be explicitly represented in an agent's environmental understanding. This requirement stands against the recent ``end-to-end'' paradigm in \ac{ai}, in which neural networks directly map an input image to an output action for a specific task, leaving an implicit internal task representation ``baked'' into the network's weights.

\begin{figure*}[t!]
	\centering
	\begin{subfigure}[b]{0.16666\linewidth}
	\includegraphics[width=\linewidth]{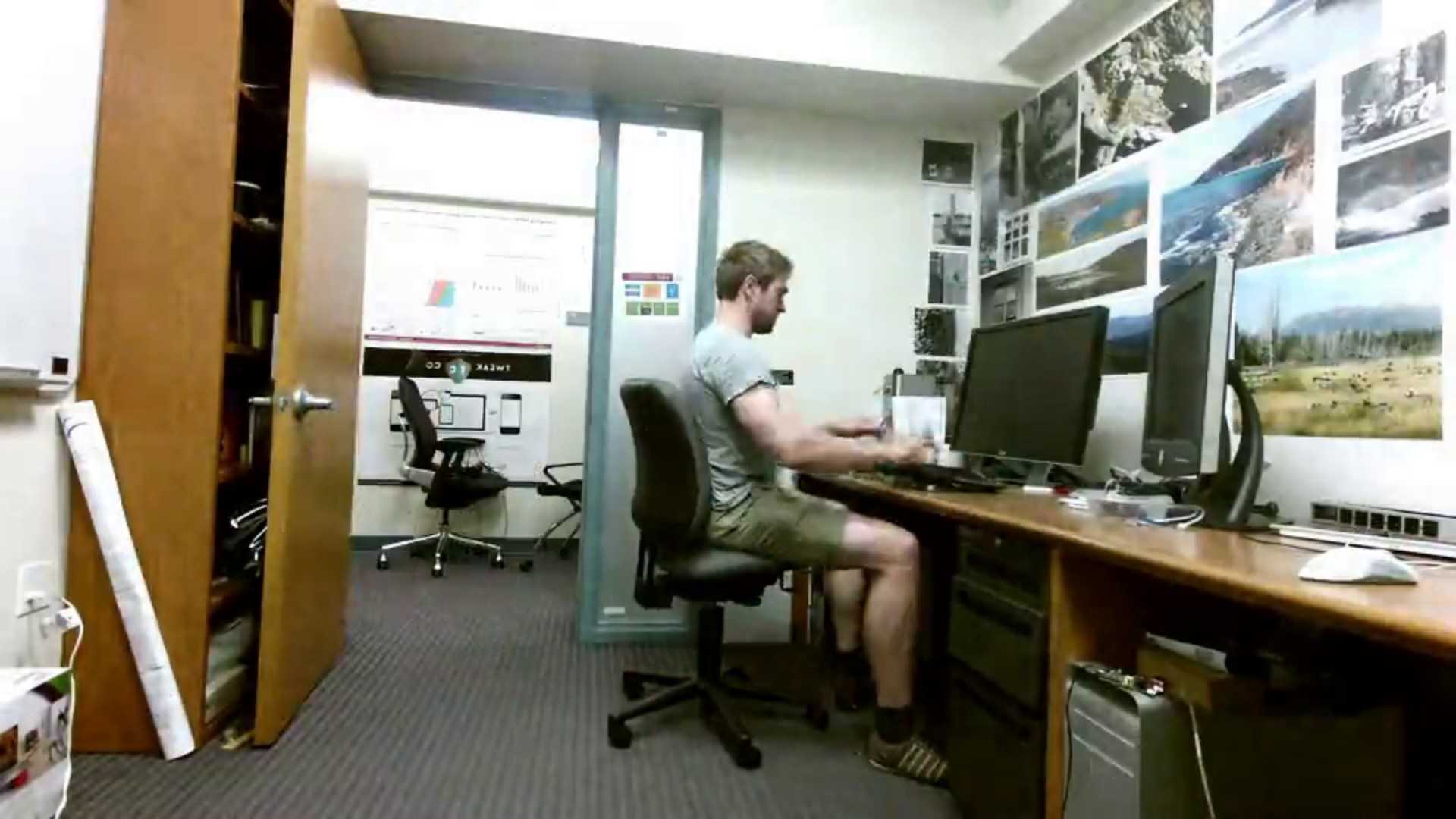}
	\includegraphics[width=\linewidth]{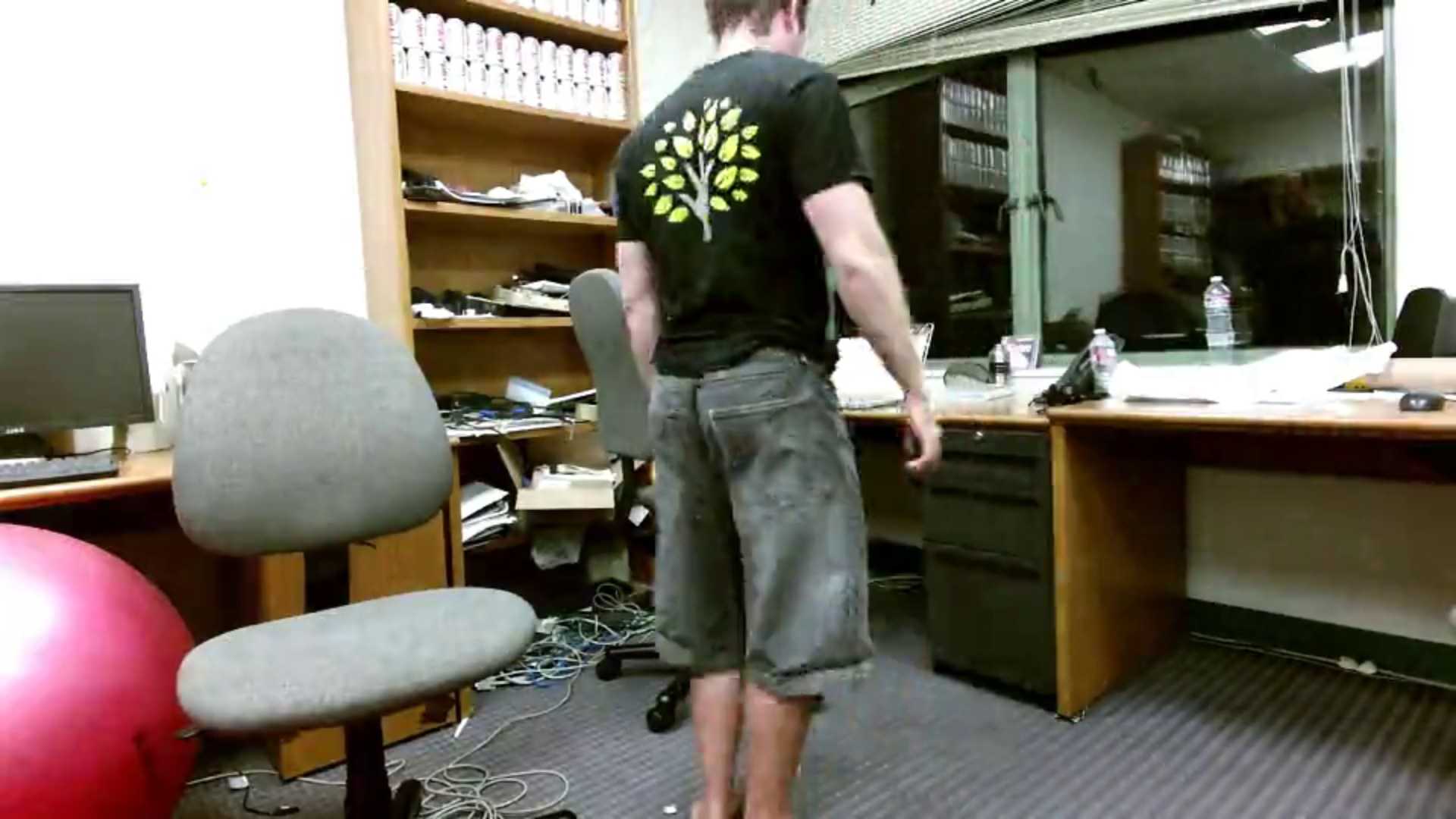}
	\includegraphics[width=\linewidth]{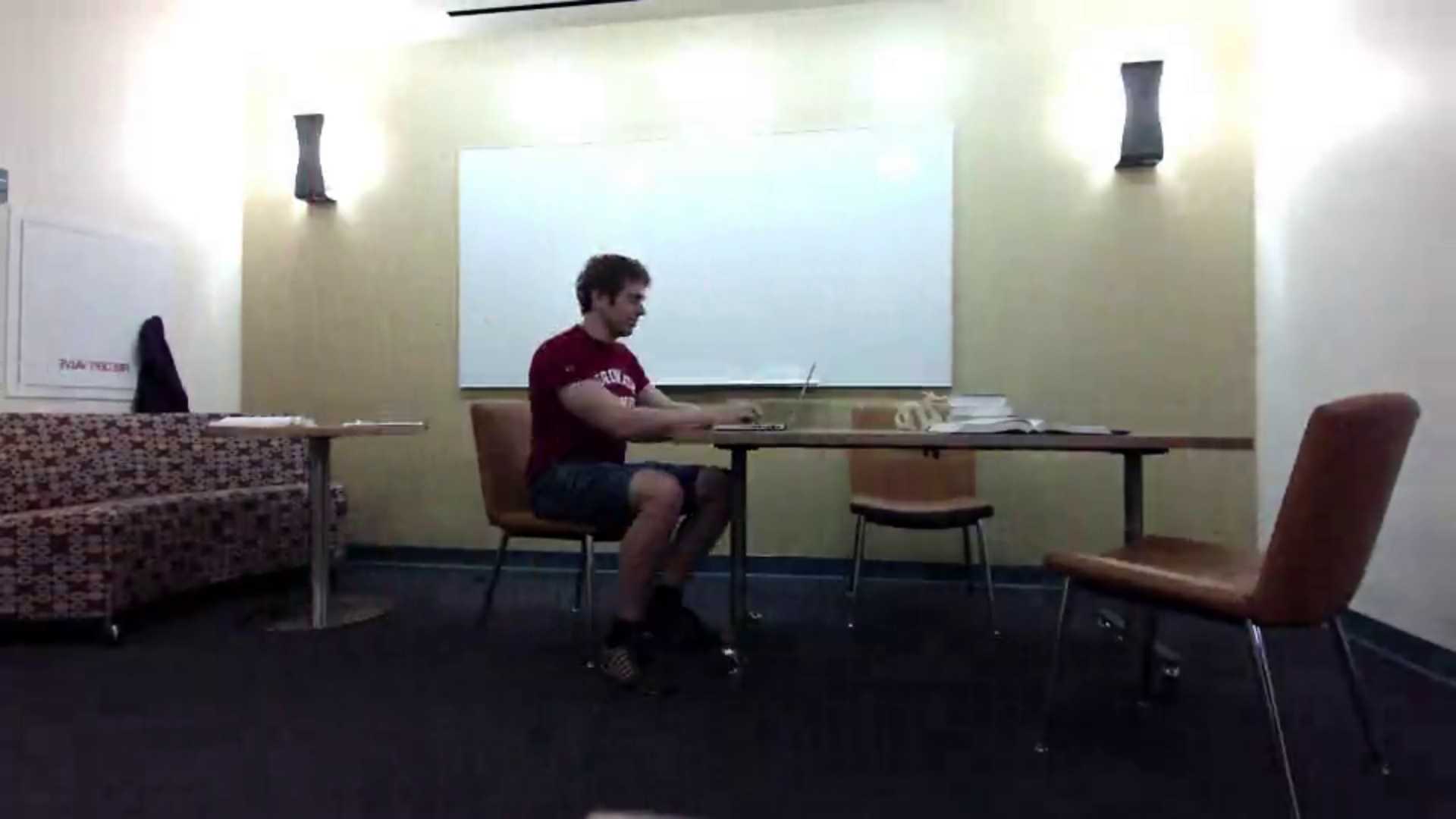}
	\includegraphics[width=\linewidth]{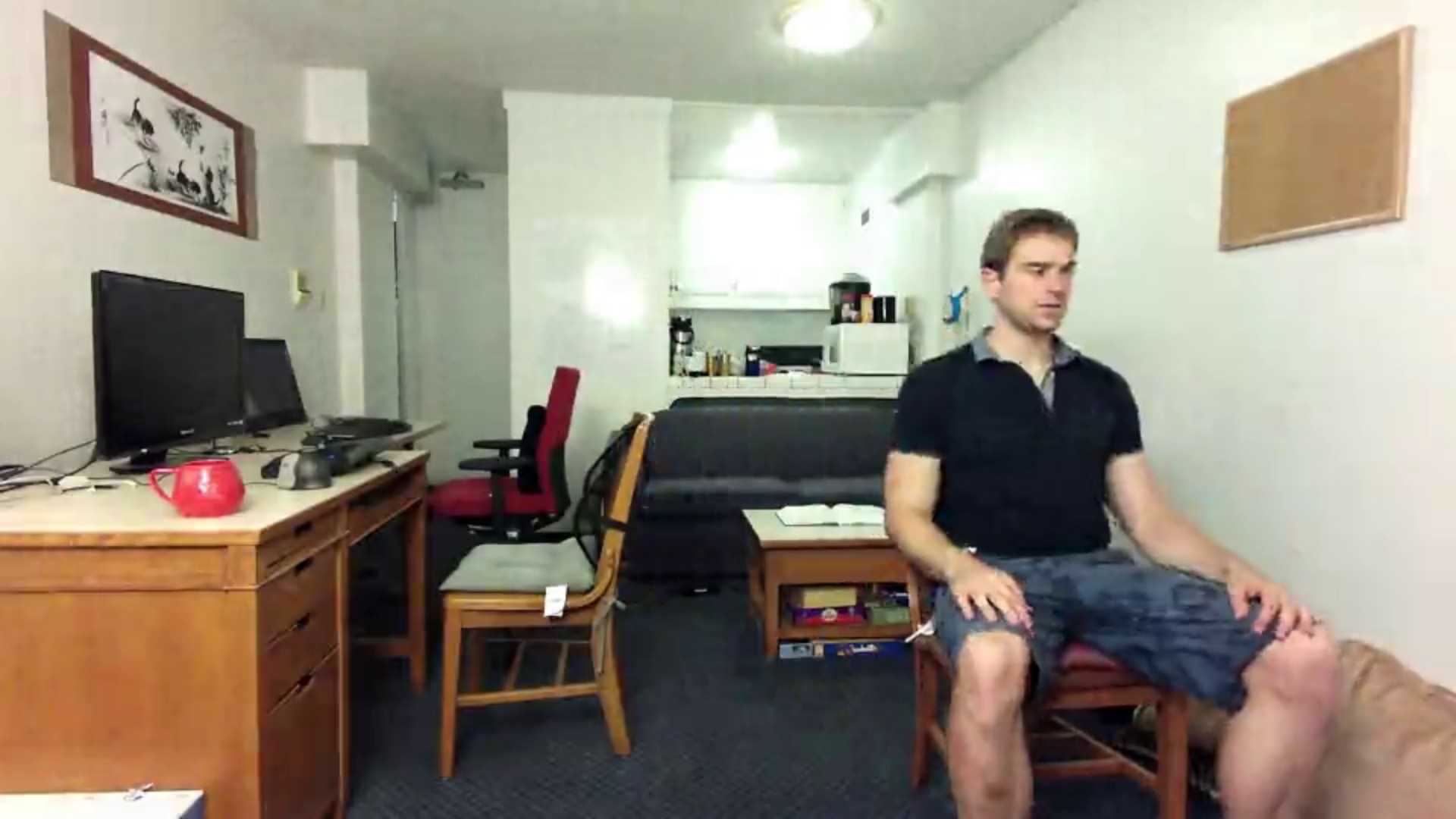}
	\caption{Input}
	\end{subfigure}%
	\begin{subfigure}[b]{0.16666\linewidth}
	\includegraphics[width=\linewidth,trim={0cm 1.7cm 0cm 1.7cm},clip]{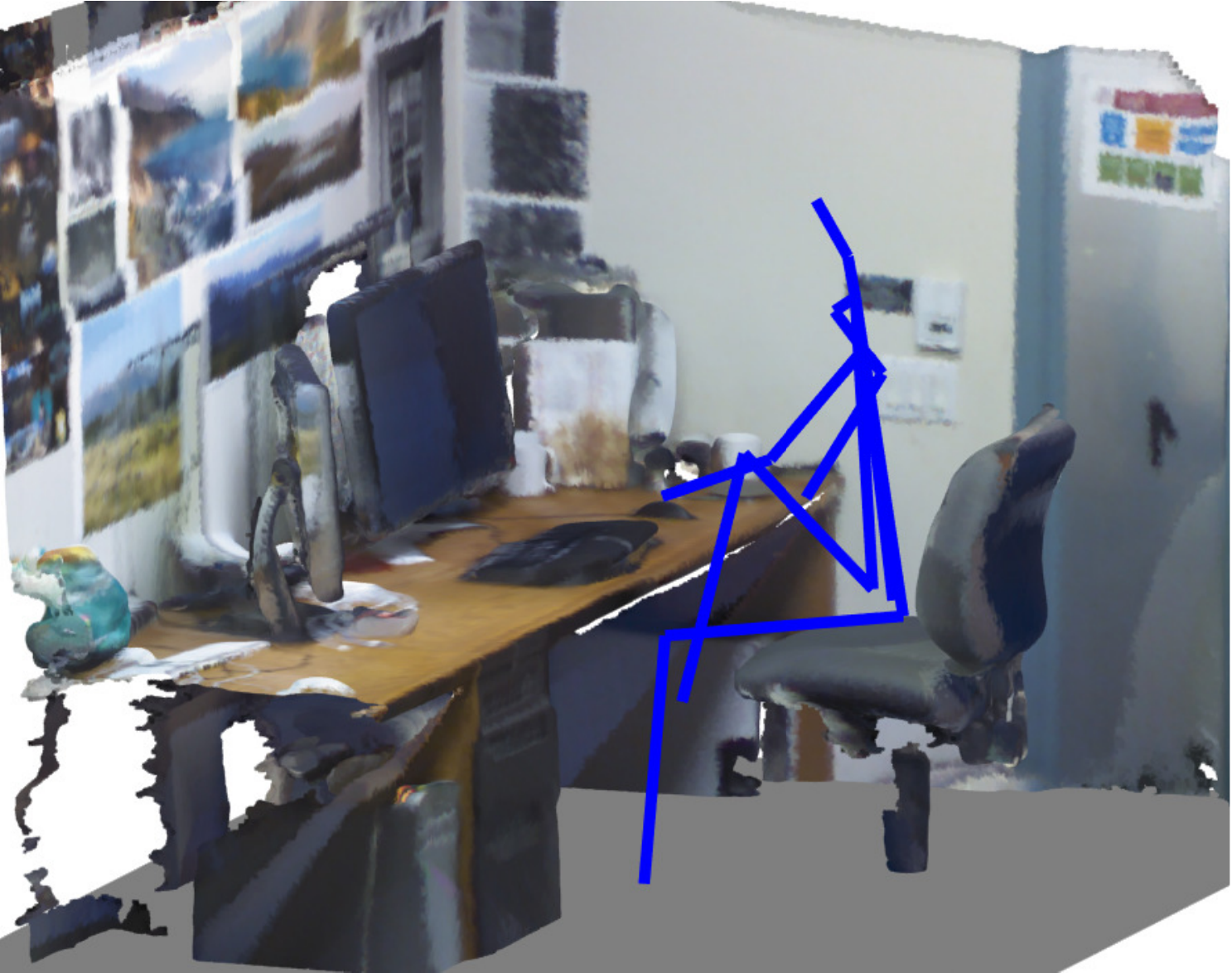}
	\includegraphics[width=\linewidth,trim={0cm 1.7cm 0cm 1.8cm},clip]{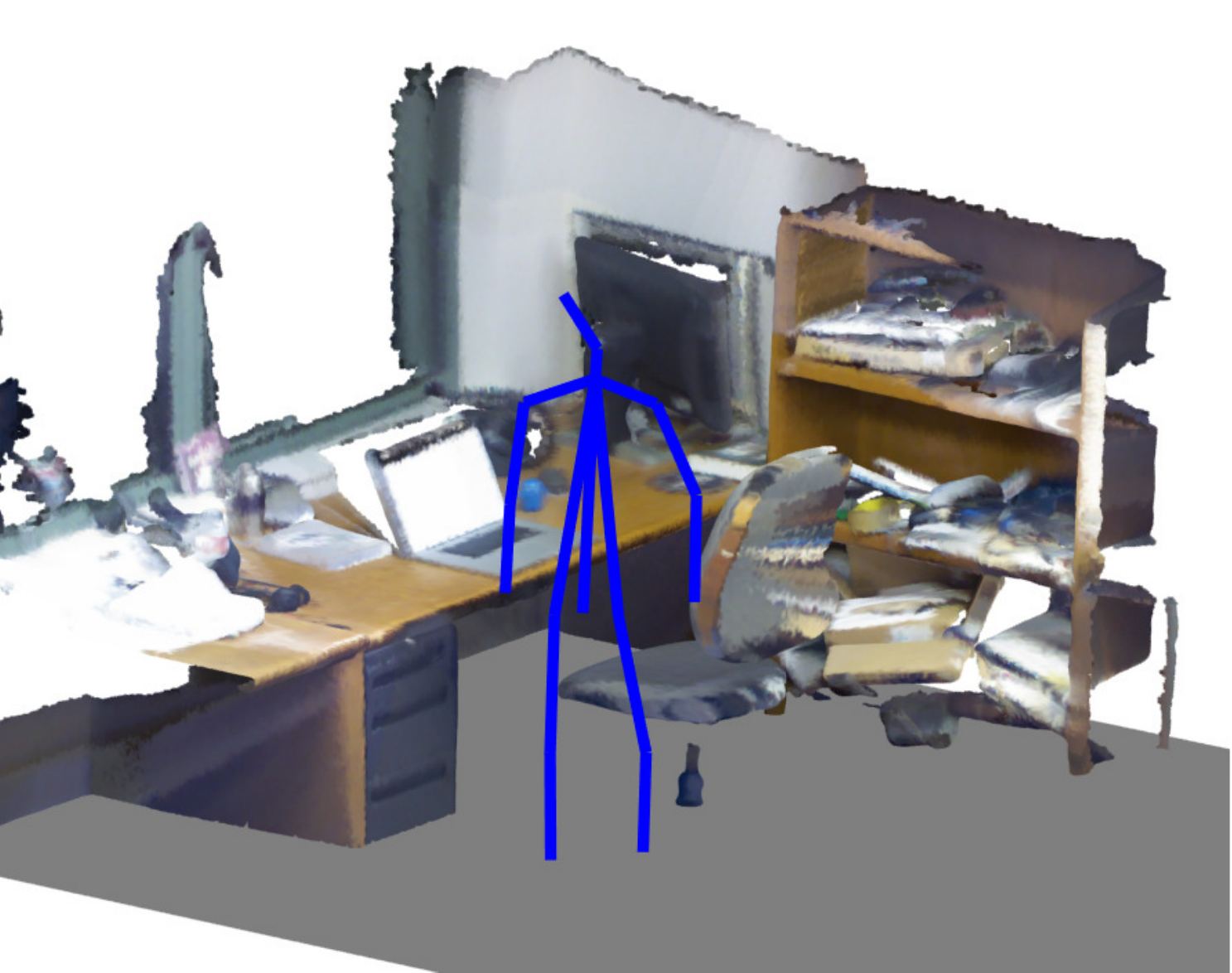}
	\includegraphics[width=\linewidth,trim={0cm 1.7cm 0cm 1.7cm},clip]{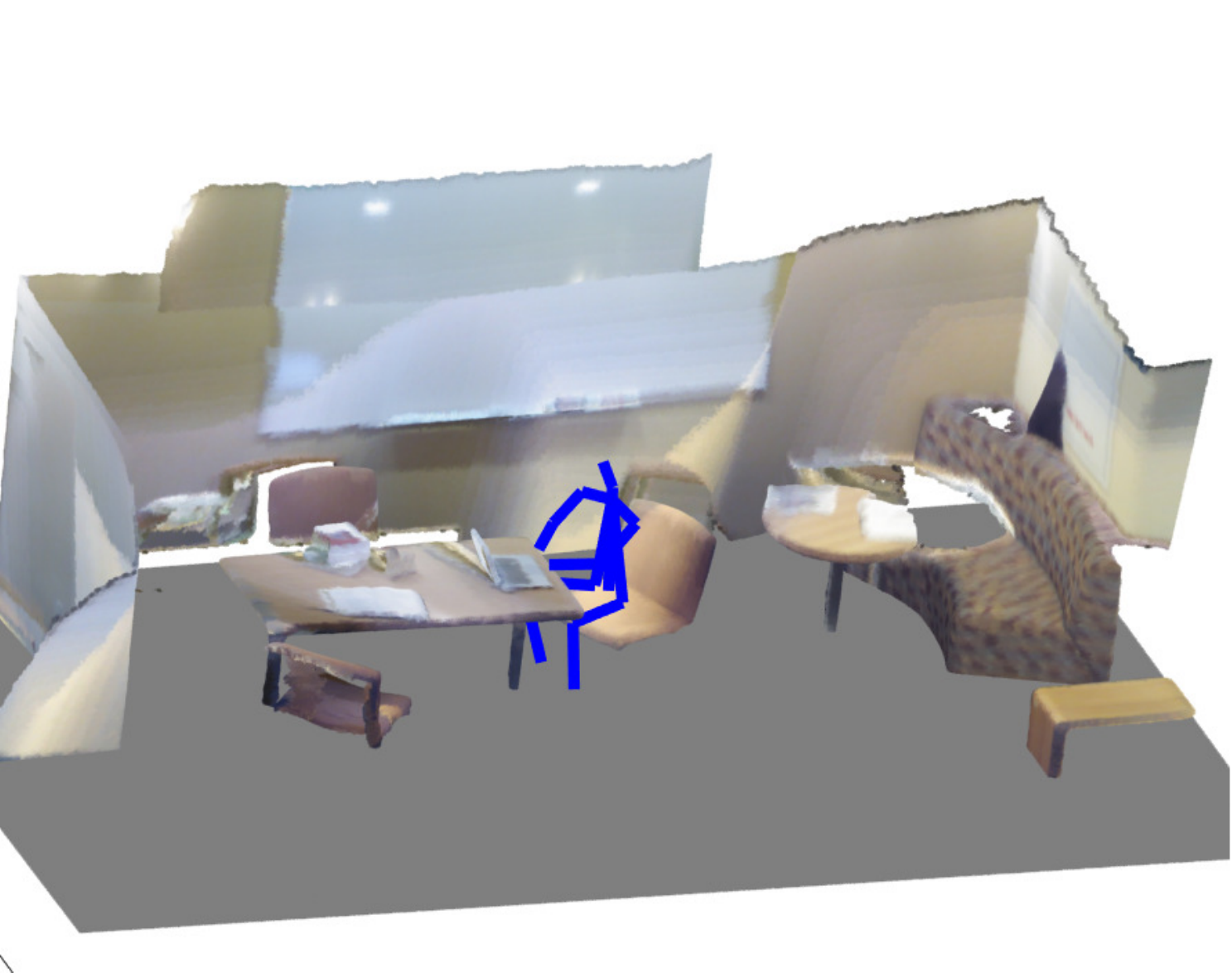}
	\includegraphics[width=\linewidth,trim={0cm 6cm 0cm 9.5cm},clip]{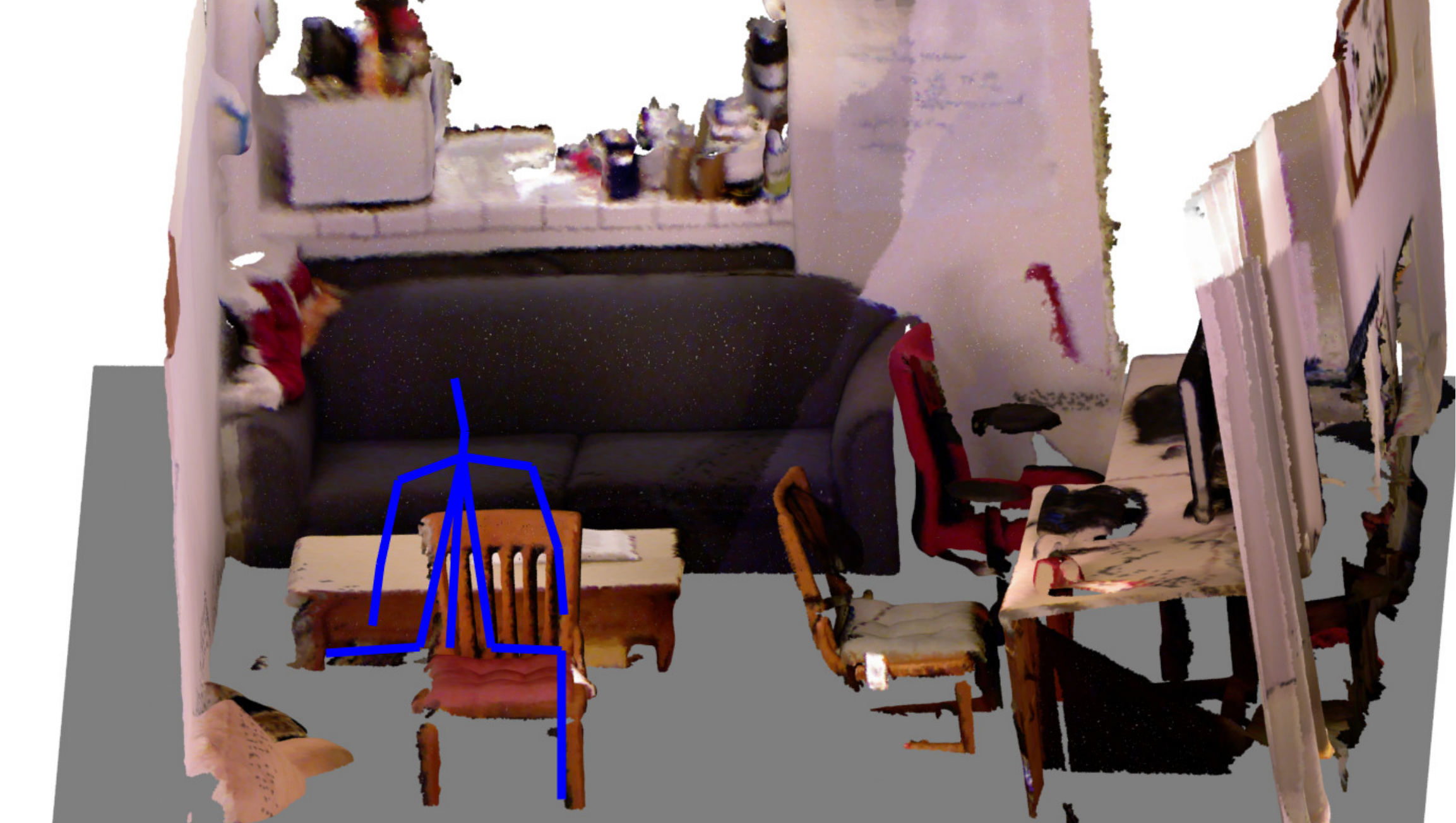}
	\caption{Ground-truth}
	\end{subfigure}%
	\begin{subfigure}[b]{0.16666\linewidth}
	\includegraphics[width=\linewidth,trim={2.5cm 6cm 2.5cm 6cm},clip]{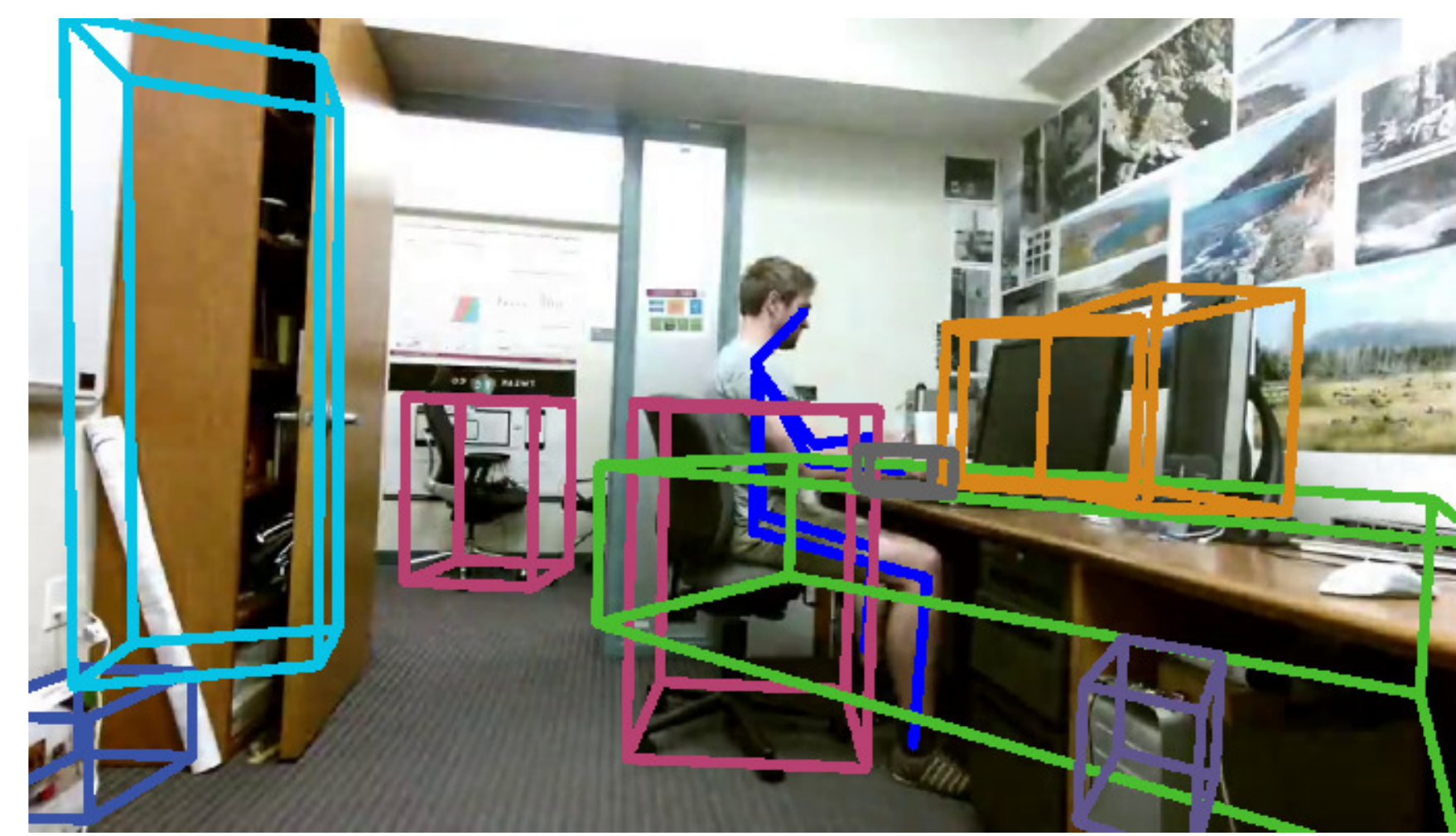}
	\includegraphics[width=\linewidth,trim={2cm 2.6cm 2cm 2.65cm},clip]{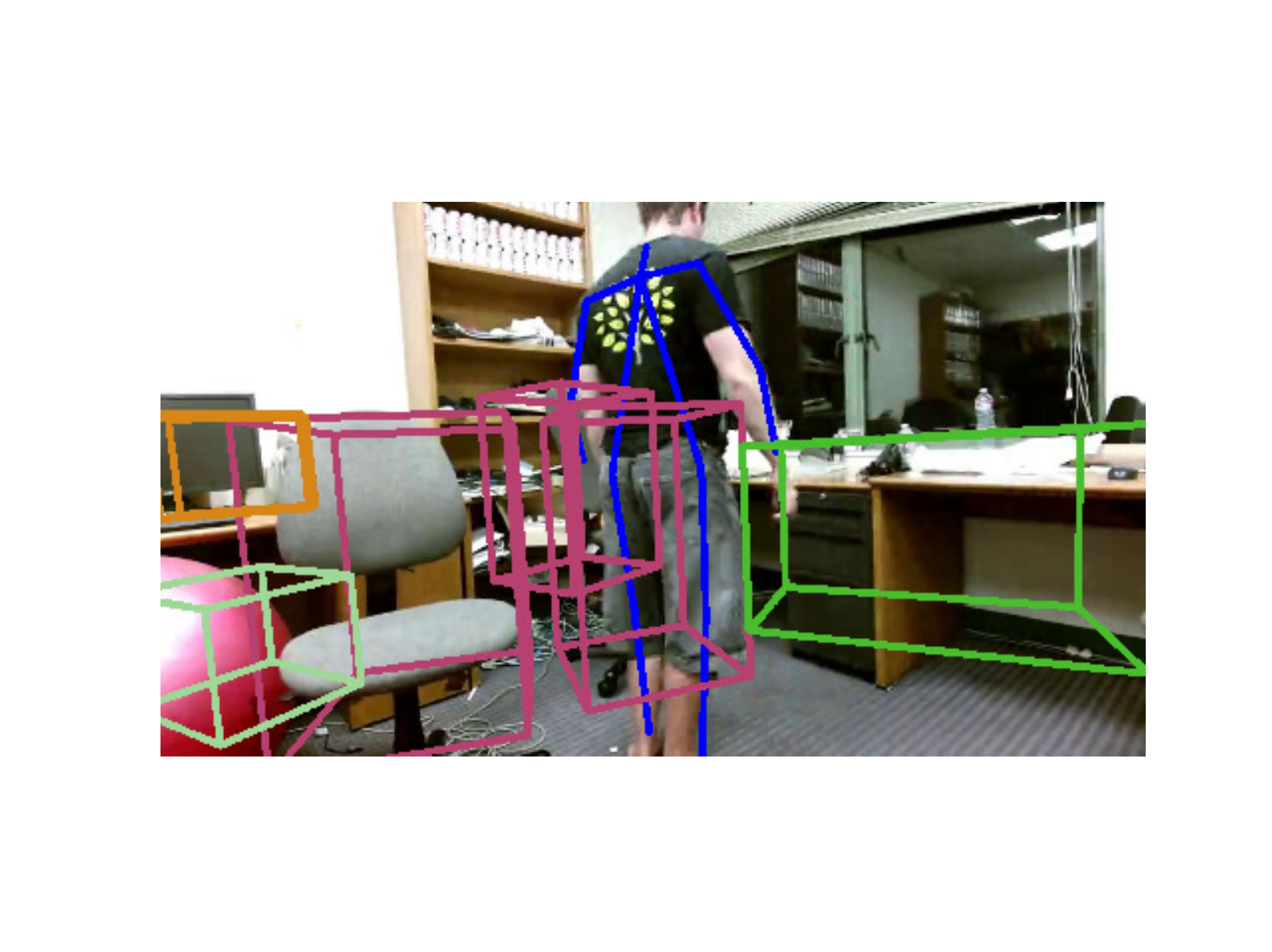}
	\includegraphics[width=\linewidth,trim={2cm 2.6cm 2cm 2.65cm},clip]{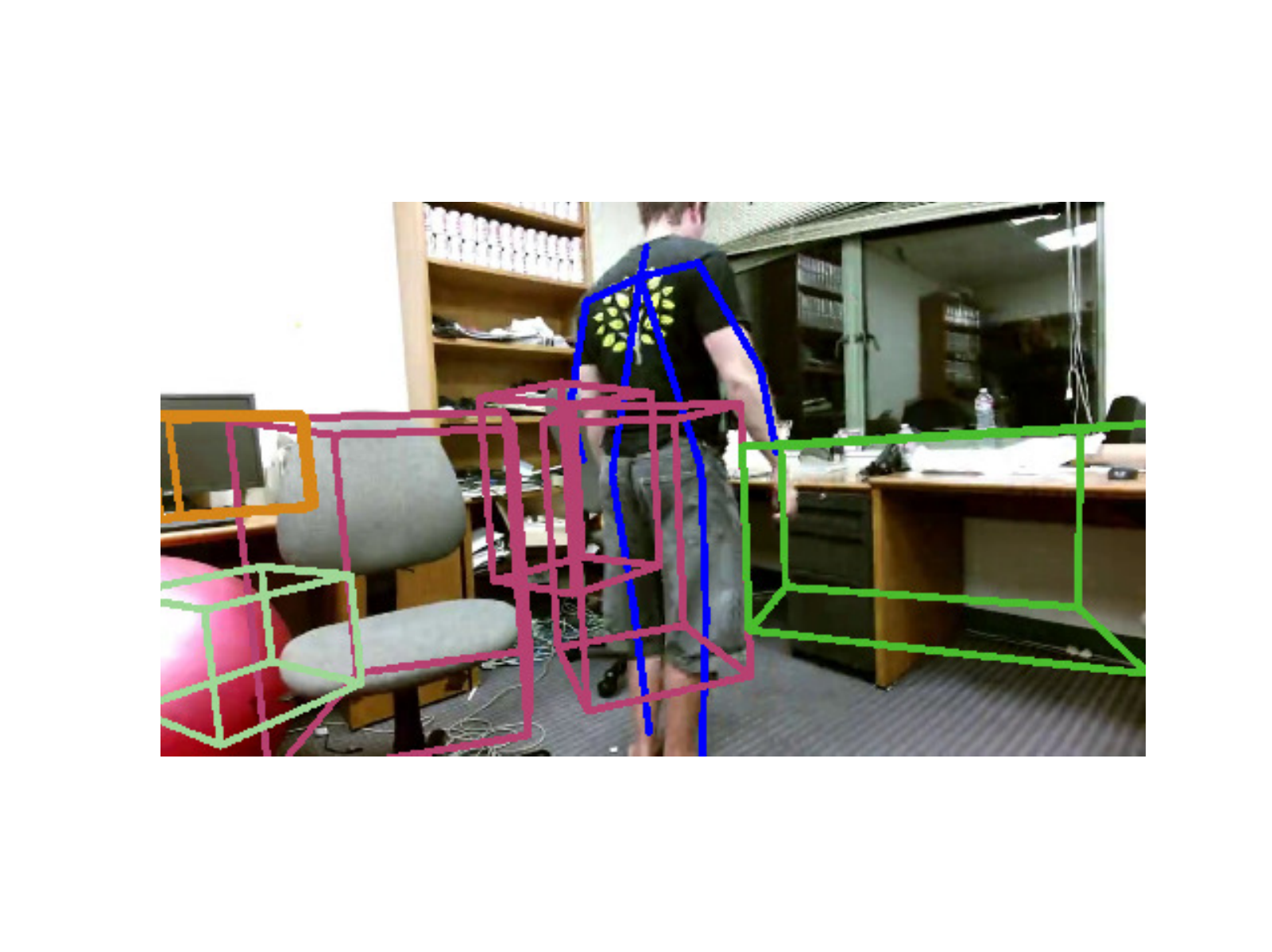}
	\includegraphics[width=\linewidth,trim={2cm 2.6cm 2cm 2.65cm},clip]{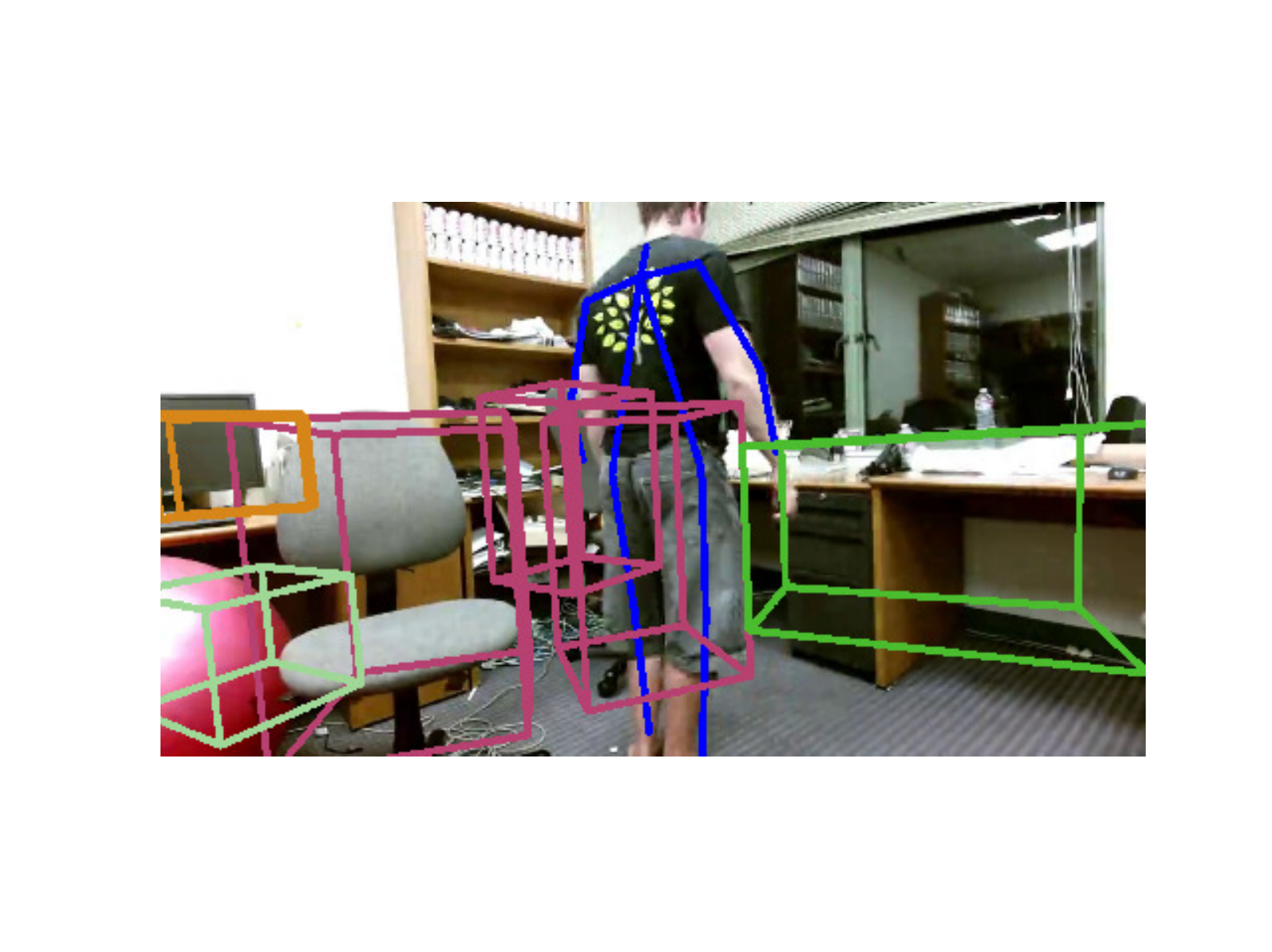}
	\caption{Without physics}
	\end{subfigure}%
	\begin{subfigure}[b]{0.16666\linewidth}
	\includegraphics[width=\linewidth,trim={2cm 4cm 2cm 3cm},clip]{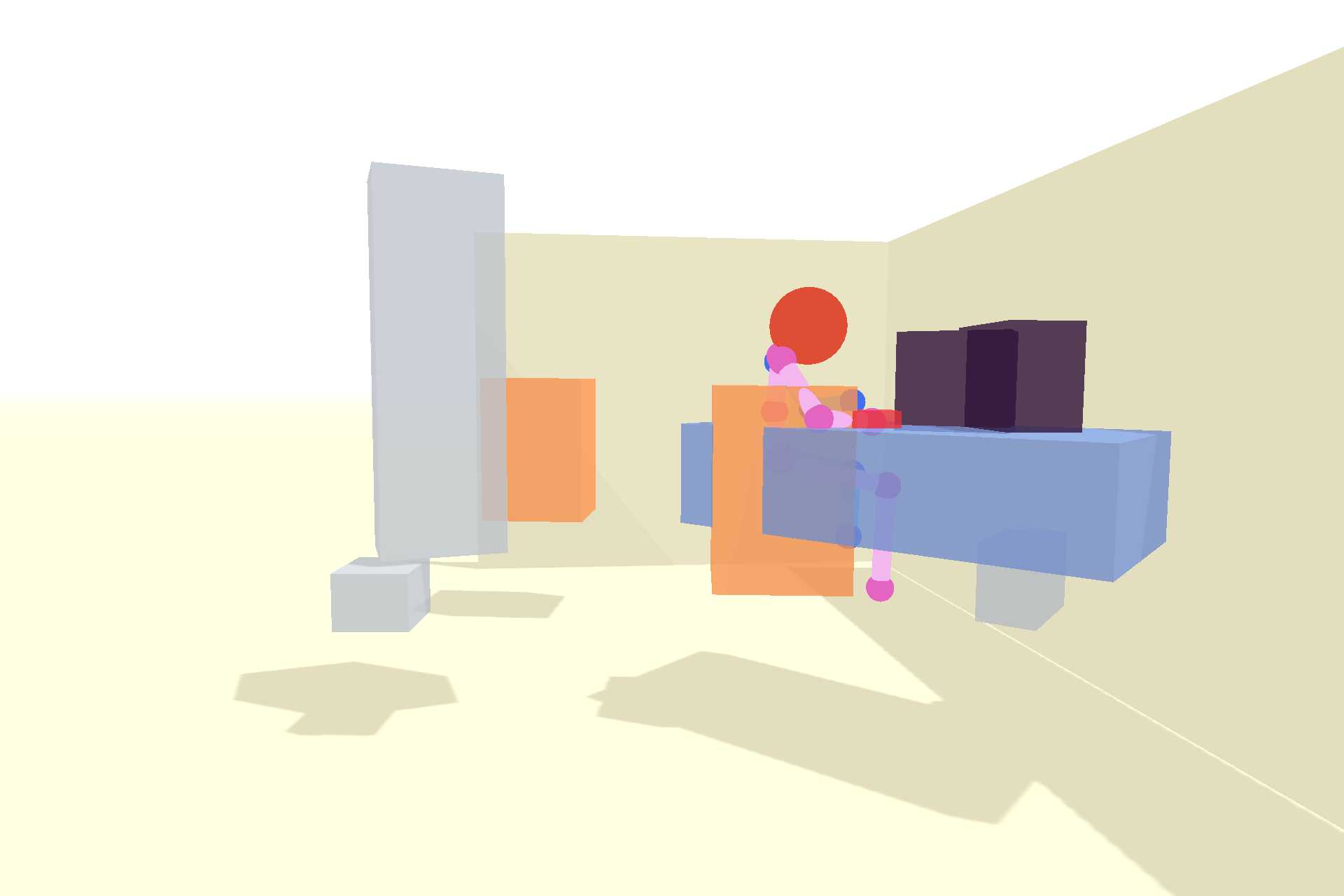}
	\includegraphics[width=\linewidth,trim={6cm 24cm 14cm 29.5cm},clip]{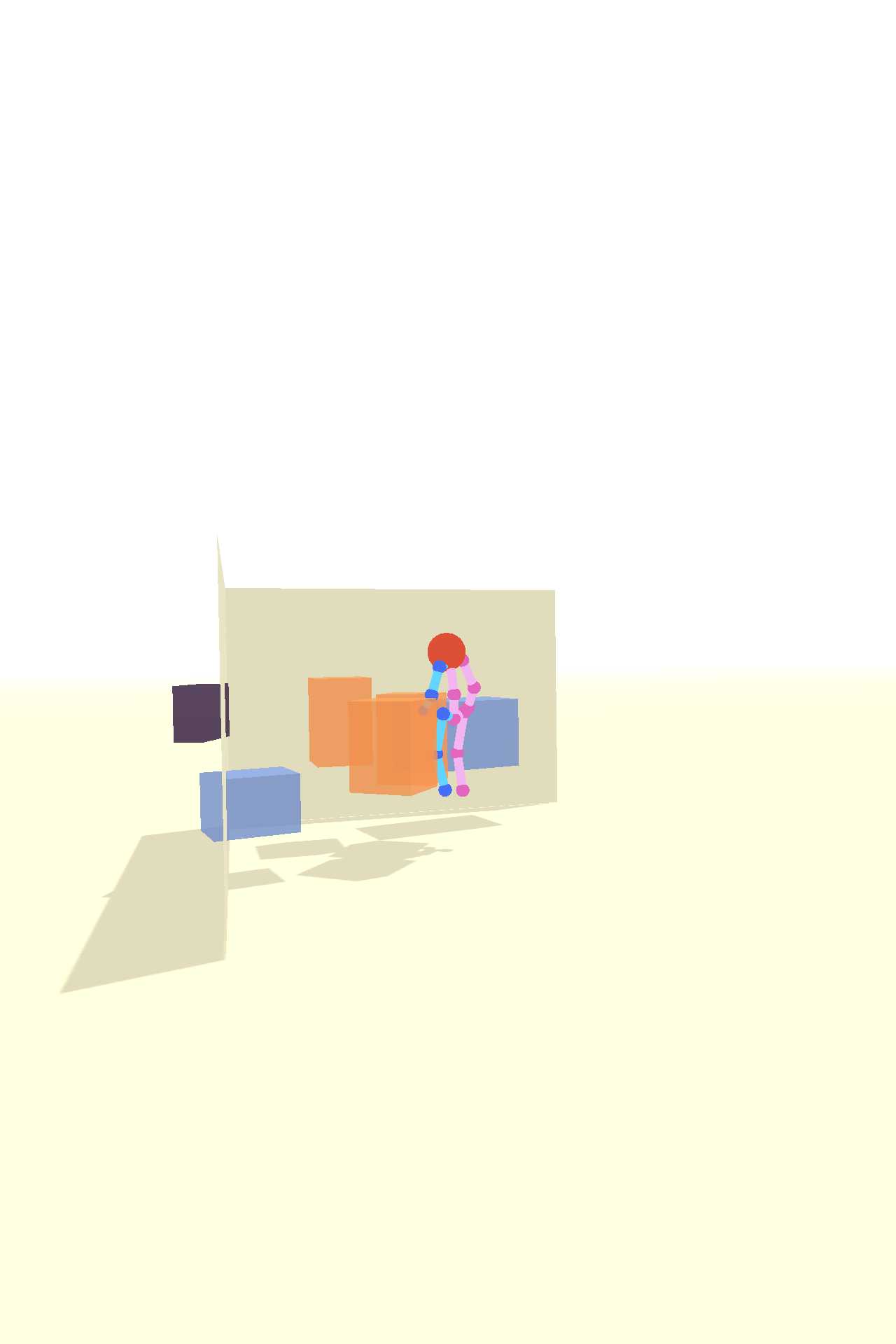}
	\includegraphics[width=\linewidth,trim={10cm 20cm 30cm 12cm},clip]{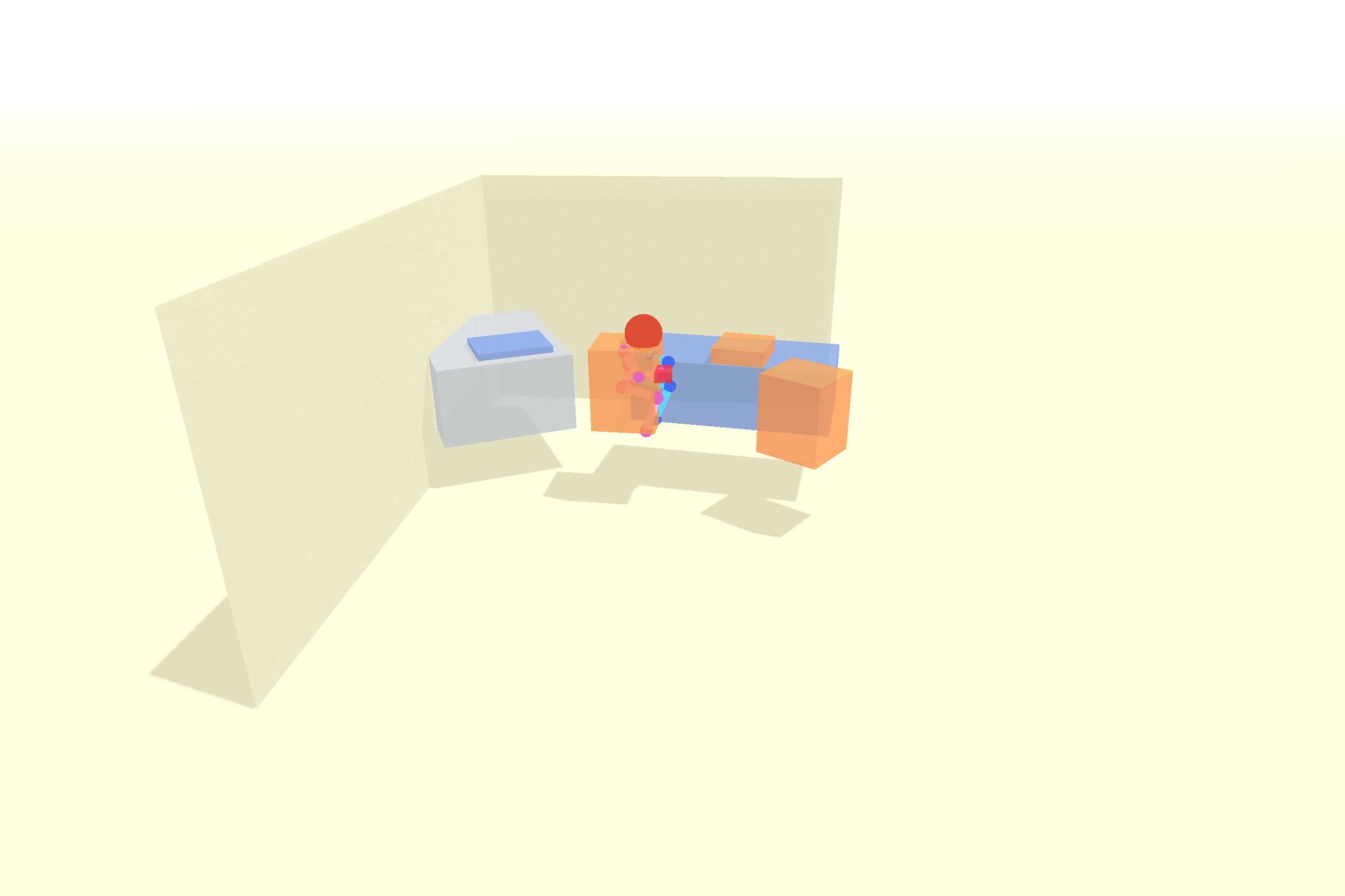}
	\includegraphics[width=\linewidth,trim={14cm 15cm 28cm 18cm},clip]{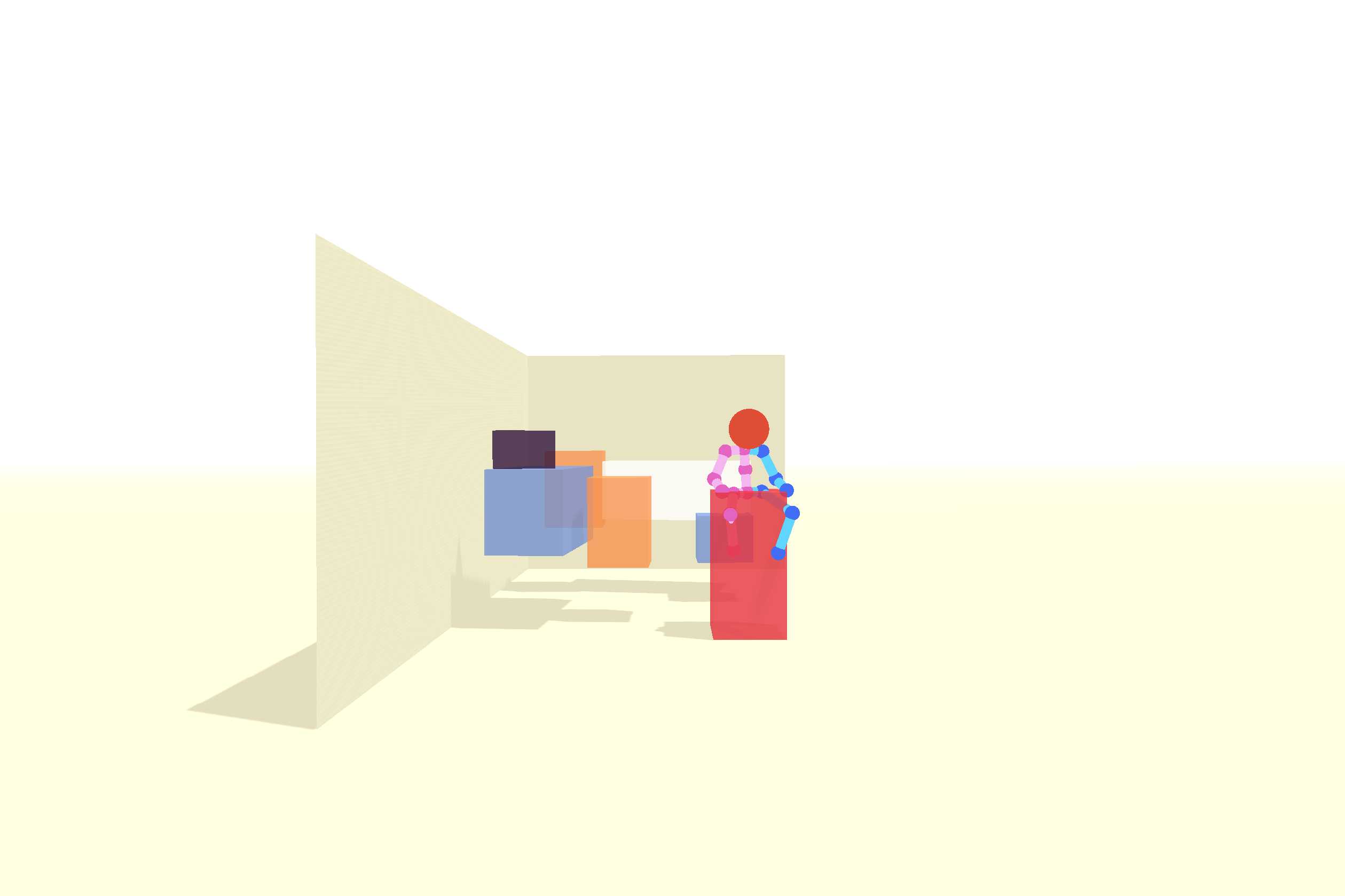}
	\caption{Without physics}
	\end{subfigure}%
	\begin{subfigure}[b]{0.16666\linewidth}
	\includegraphics[width=\linewidth,trim={2cm 2.6cm 2cm 2.65cm},clip]{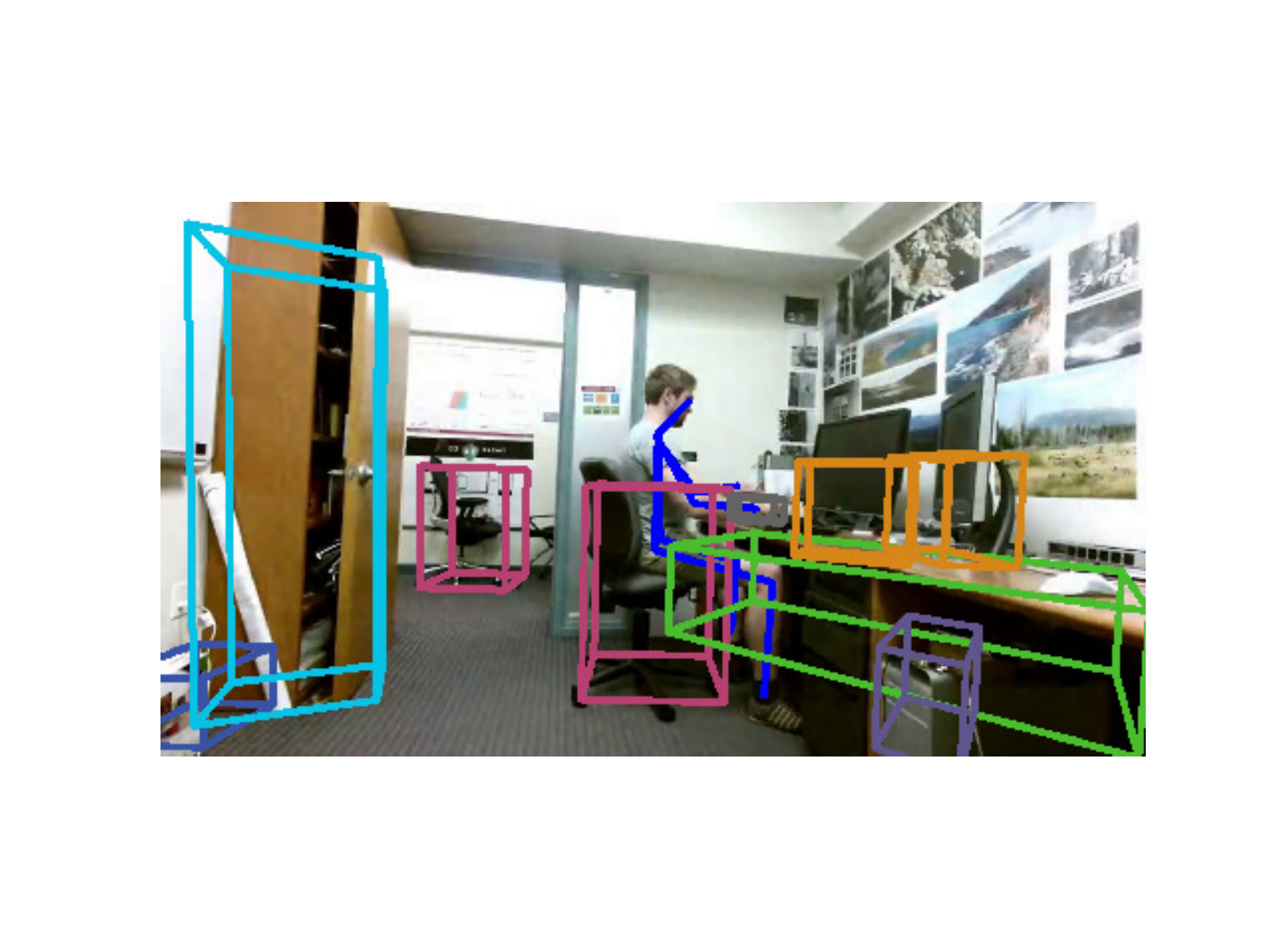}
	\includegraphics[width=\linewidth,trim={2cm 2.6cm 2cm 2.65cm},clip]{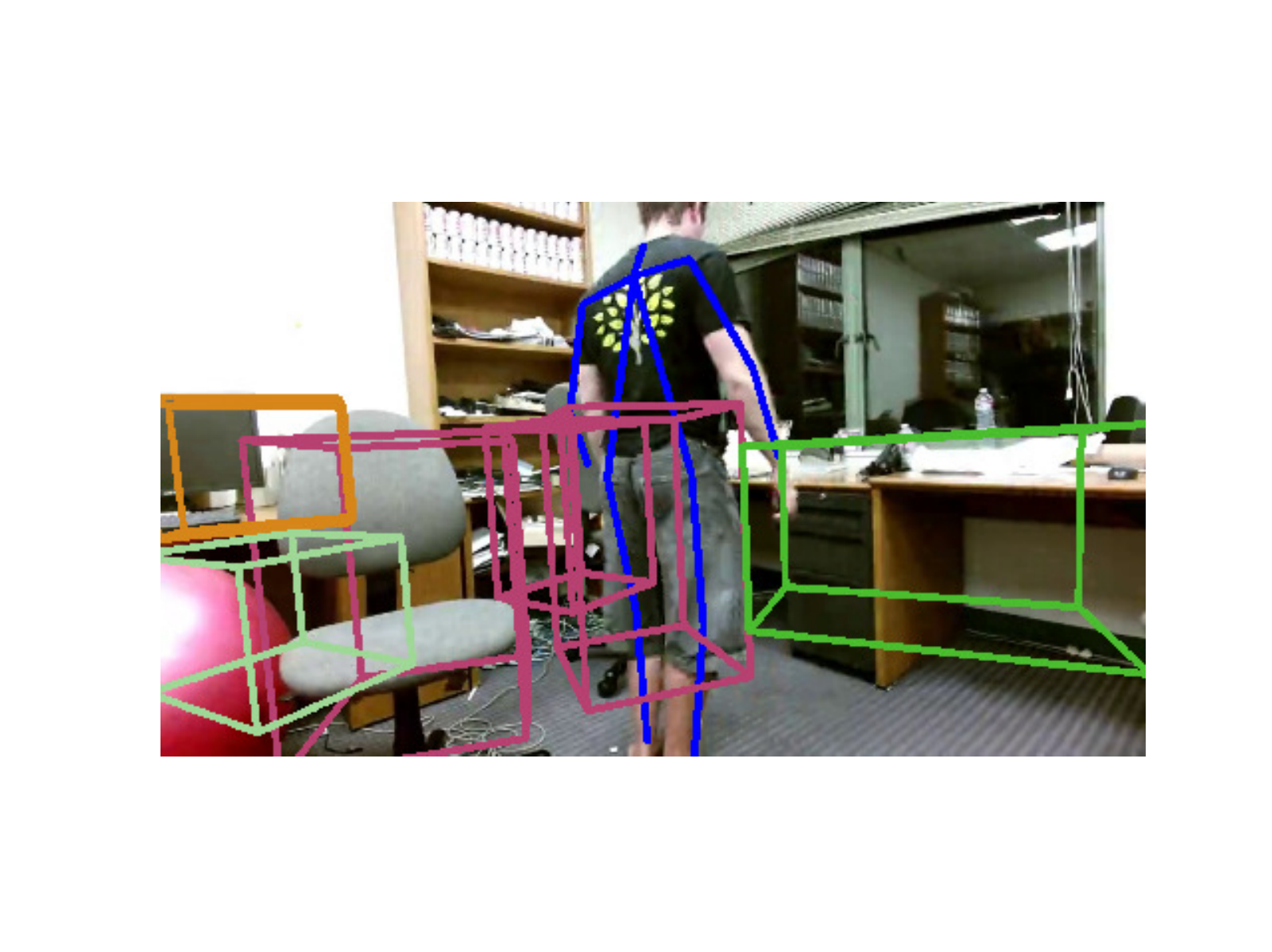}
	\includegraphics[width=\linewidth,trim={2cm 2.6cm 2cm 2.65cm},clip]{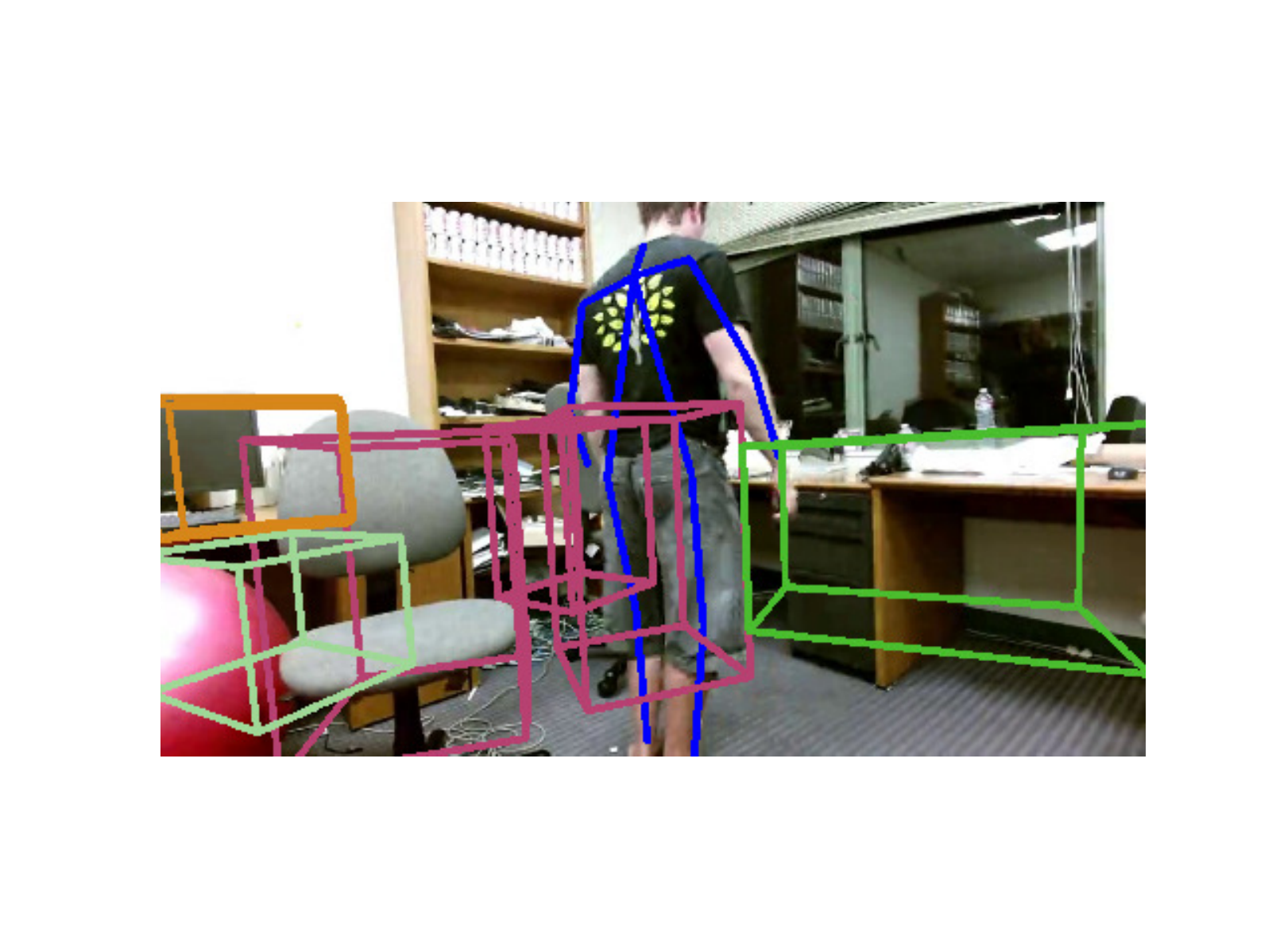}
	\includegraphics[width=\linewidth,trim={2cm 2.6cm 2cm 2.65cm},clip]{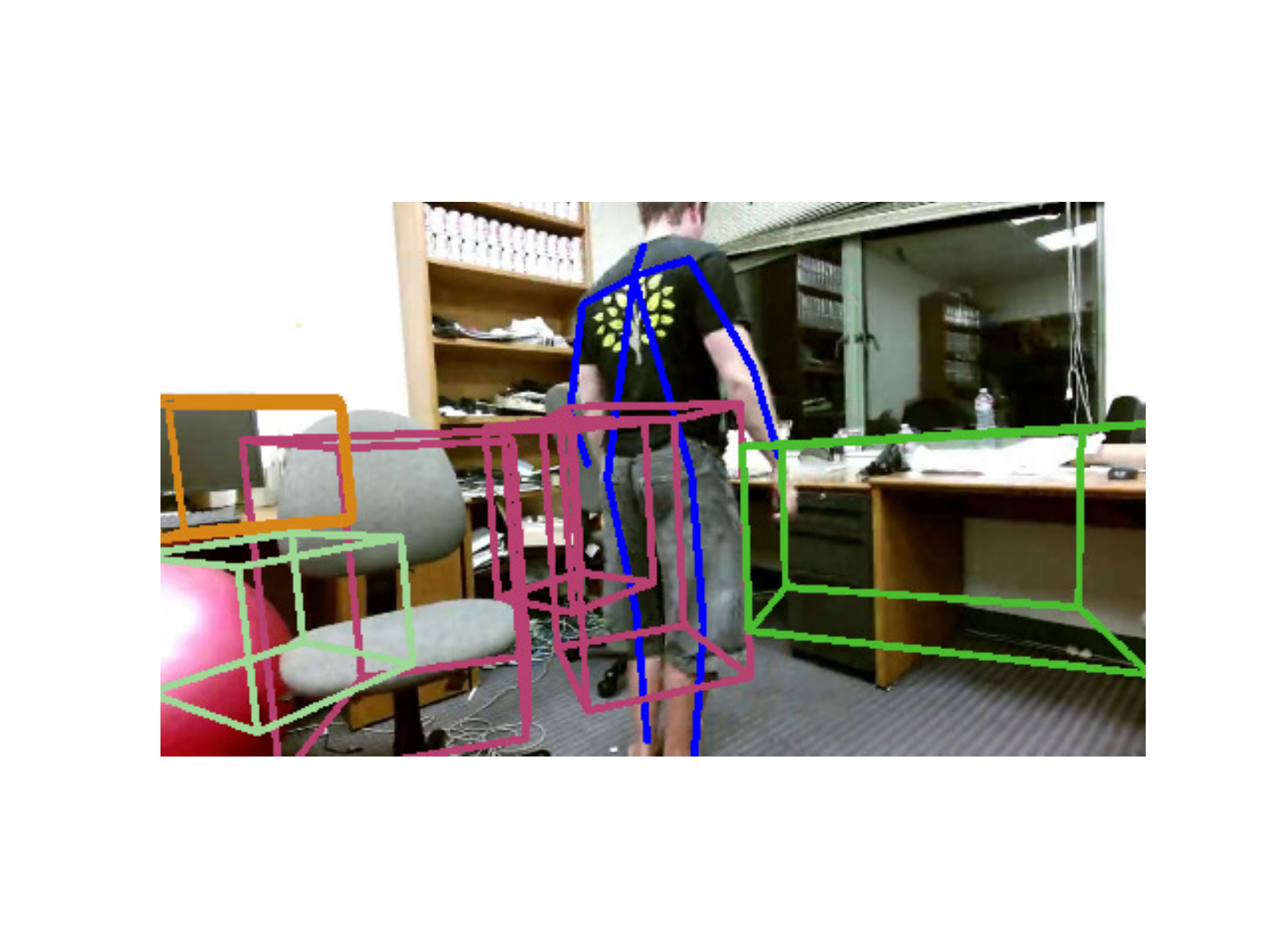}
	\caption{With physics}
	\end{subfigure}%
	\begin{subfigure}[b]{0.16666\linewidth}
	\includegraphics[width=\linewidth,trim={4cm 10cm 4cm 2cm},clip]{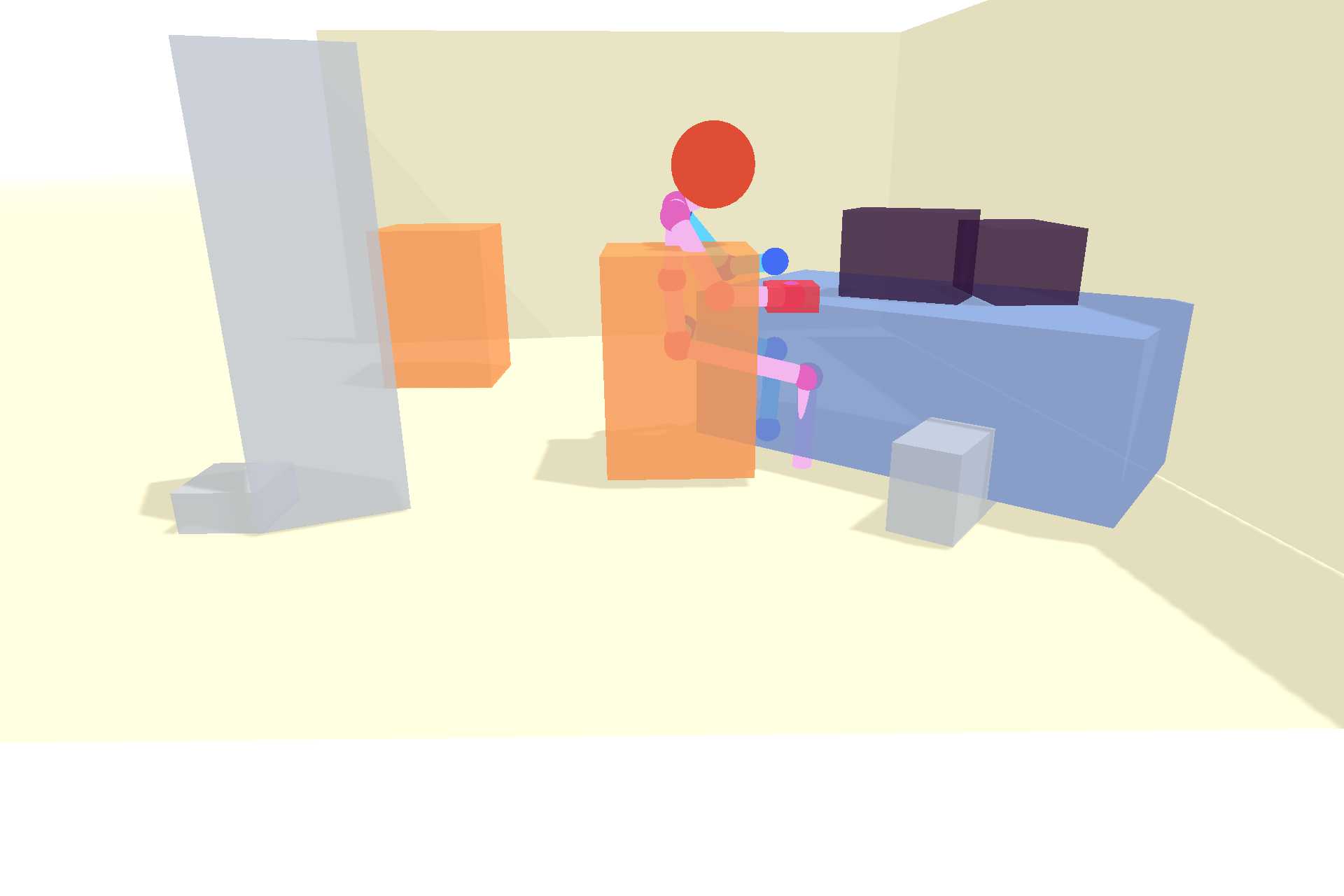}
	\includegraphics[width=\linewidth,trim={8cm 14cm 8cm 2cm},clip]{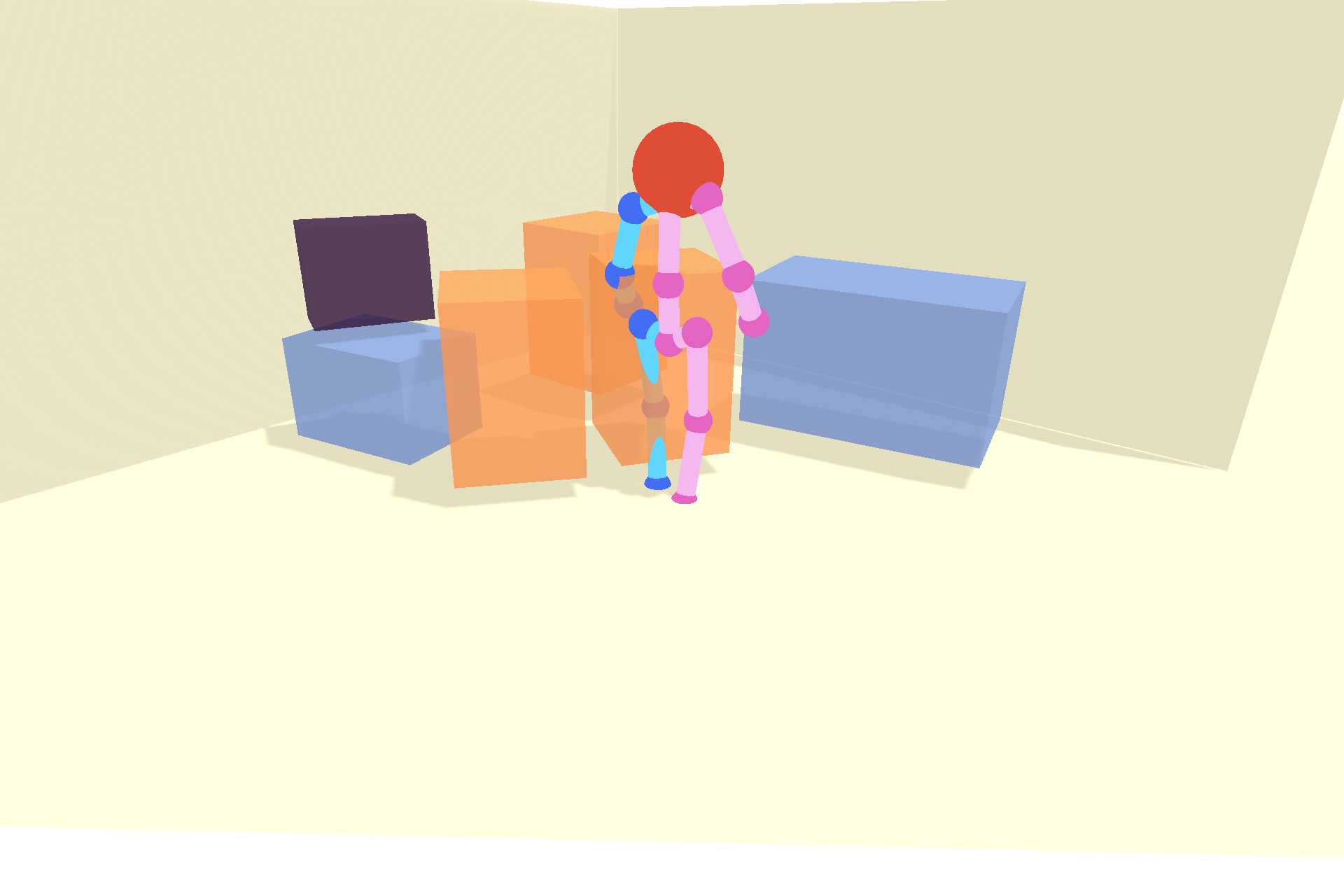}
	\includegraphics[width=\linewidth,trim={12cm 15cm 12cm 5cm},clip]{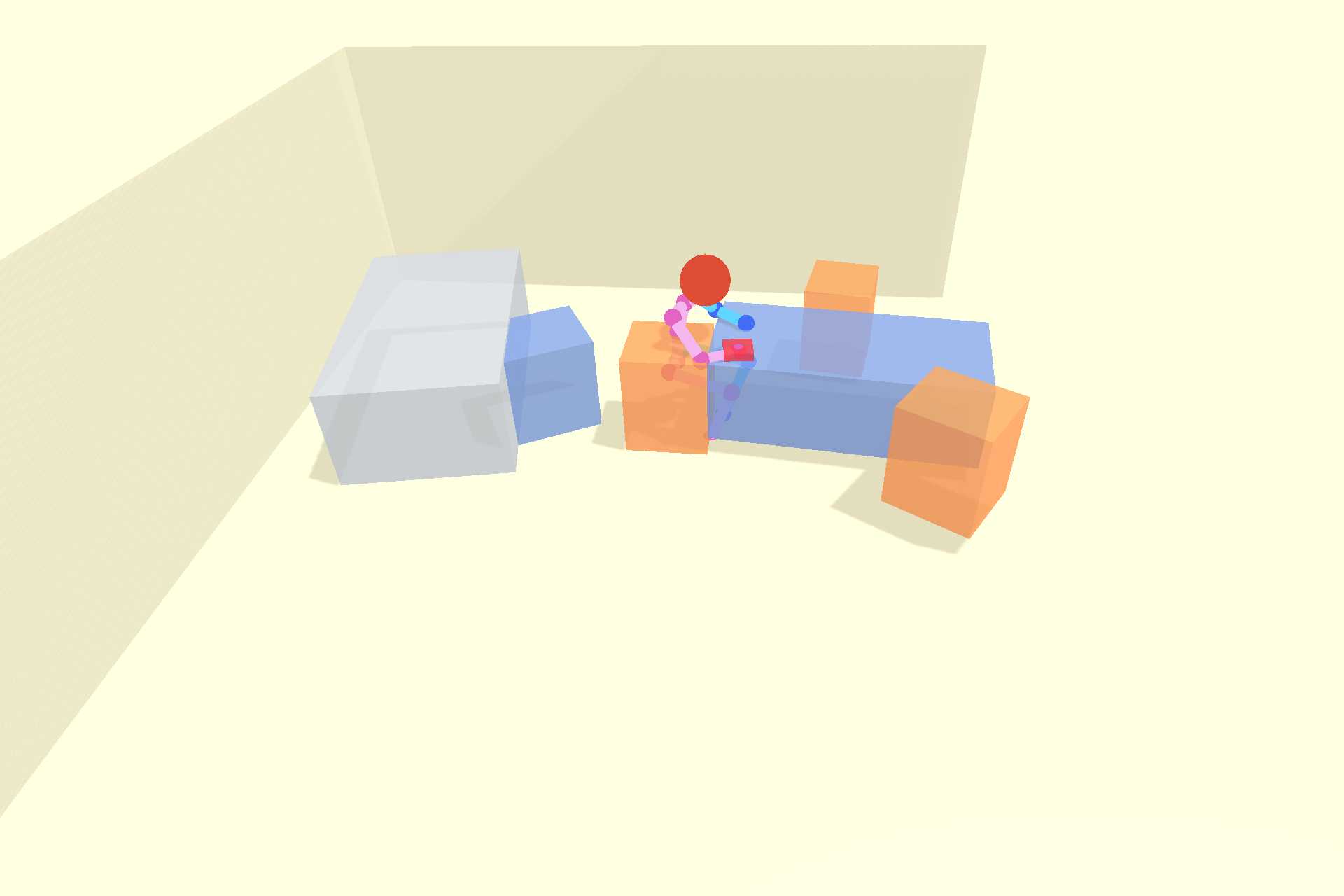}
	\includegraphics[width=\linewidth,trim={6cm 11cm 10cm 5cm},clip]{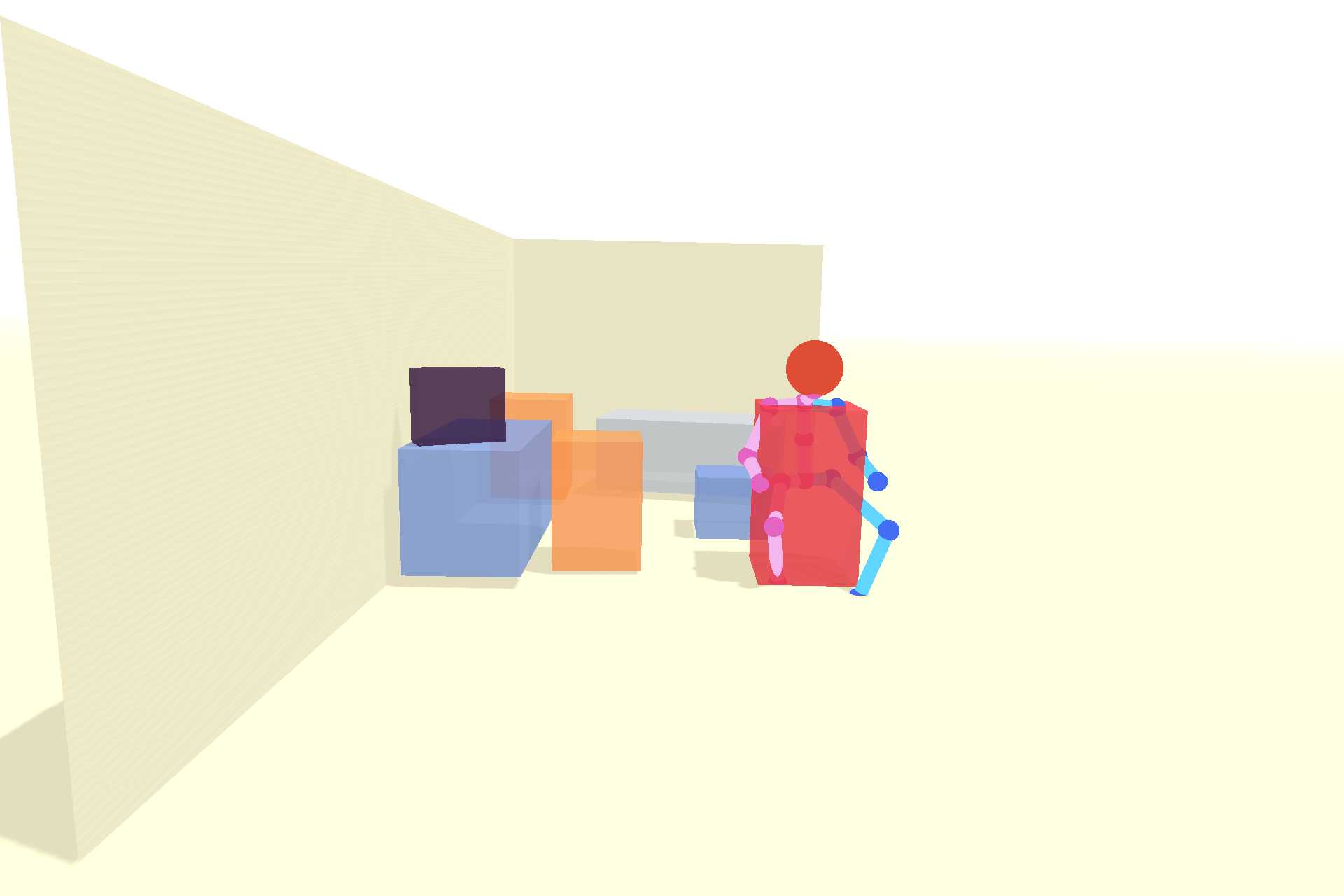}
	\caption{With physics}
	\end{subfigure}
	\caption{Scene parsing and reconstruction by integrating physics and human-object interactions. (a) Input image; (b) ground truth; (c and d) without incorporating physics, the objects might appear to float in the air, resulting in an incorrect parsing; (e and f) after incorporating physics, the parsed 3D scene appears physically stable. The system has been able to perceive the ``dark'' physical stability in which objects must rest on one another to be stable. Reproduced from Ref.~\citep{chen2019holistic} with permission of IEEE, \textcopyright~2019.}
	\label{fig:scene_functionality_physics}
\end{figure*}

Recent breakthroughs in cognitive science provide solid evidence supporting the existence of an intuitive physics model in human scene understanding. This evidence suggests that humans perform physical inferences by running probabilistic simulations in a mental physics engine akin to the 3D physics engines used in video games~\citep{ullman2017mind}; see \cref{fig:falling_tower}~\citep{battaglia2013simulation}. Human intuitive physics can be modeled as an approximated physical engine with a Bayesian probabilistic model~\citep{battaglia2013simulation}, possessing the following distinguishing properties: (i) Physical judgment is achieved by running a coarse and rough forward physical simulation; and (ii) the simulation is stochastic, which is different from the deterministic and precise physics engine developed in computer graphics. For example, in the tower stability task presented in Ref.~\citep{battaglia2013simulation}, there is uncertainty about the exact physical attributes of the blocks; they fall into a probabilistic distribution. For every simulation, the model first samples the blocks' attributes, then generates predicted states by recursively applying elementary physical rules over short-time intervals. This process creates a distribution of simulated results. The stability of a tower is then represented in the results as the probability of the tower not falling. Due to its stochastic nature, this model will judge a tower as stable only when it can tolerate small jitters or other disturbances to its components. This single model fits data from five distinct psychophysical tasks, captures several illusions and biases, and explains core aspects of mental models and commonsense reasoning that are instrumental to how humans understand their everyday world.

More recent studies have demonstrated that intuitive physical cognition is not limited to the understanding of rigid bodies, but also expands to the perception and simulation of the physical properties of liquids~\citep{bates2015humans,kubricht2016probabilistic} and sand~\citep{kubricht2017consistent}. In these studies, the experiments demonstrate that humans do not rely on simple qualitative heuristics to reason about fluid or granular dynamics; instead, they rely on perceived physical variables to make quantitative judgments. Such results provide converging evidence supporting the idea of mental simulation in physical reasoning. For a more in-depth review of intuitive physics in psychology, see Ref.~\citep{kubricht2017intuitive}.

\setstretch{0.98}

\subsection{Physics-based Reasoning in Computer Vision}

\begin{figure*}[t!]
	\centering
	\begin{subfigure}[b]{0.235\linewidth}
	\includegraphics[width=\linewidth]{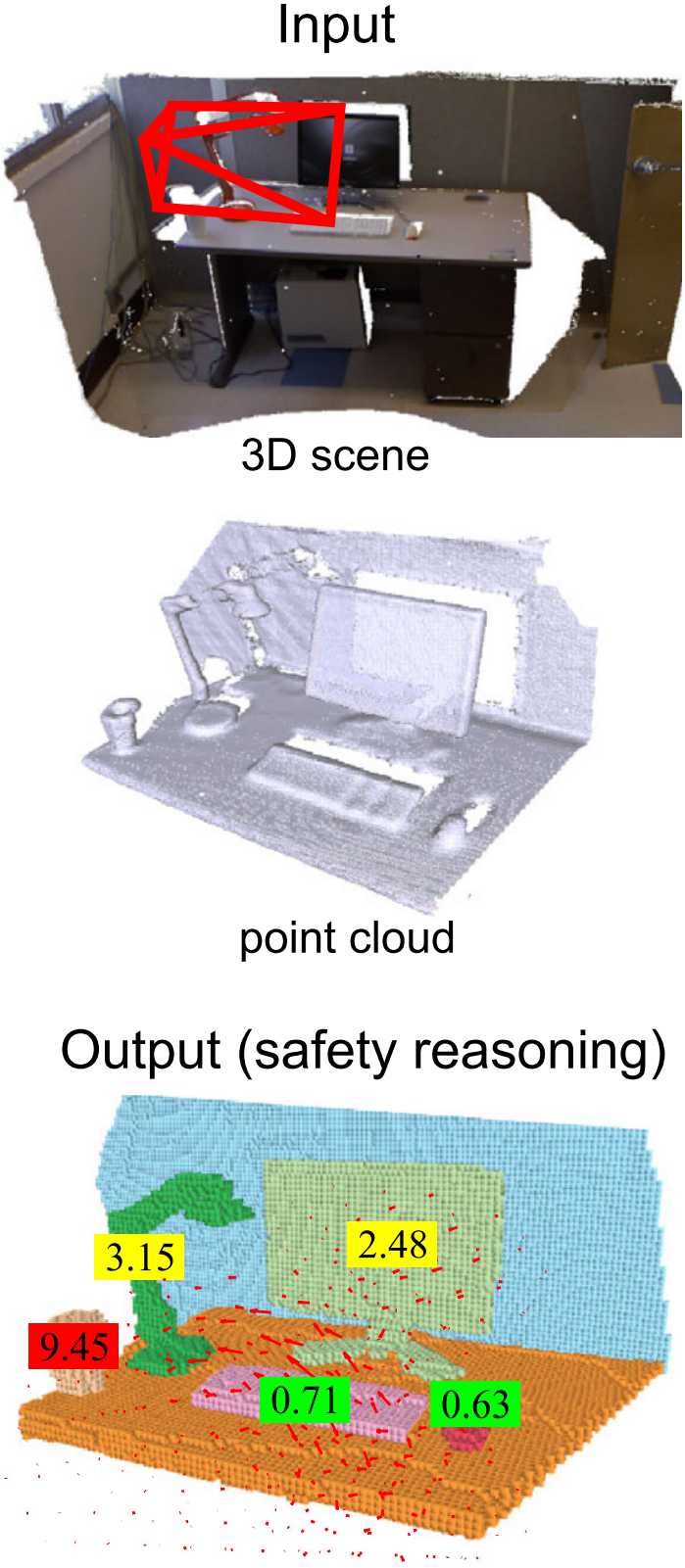}
	\caption{}
	\end{subfigure}%
	\begin{subfigure}[b]{0.765\linewidth}
	\includegraphics[width=\linewidth]{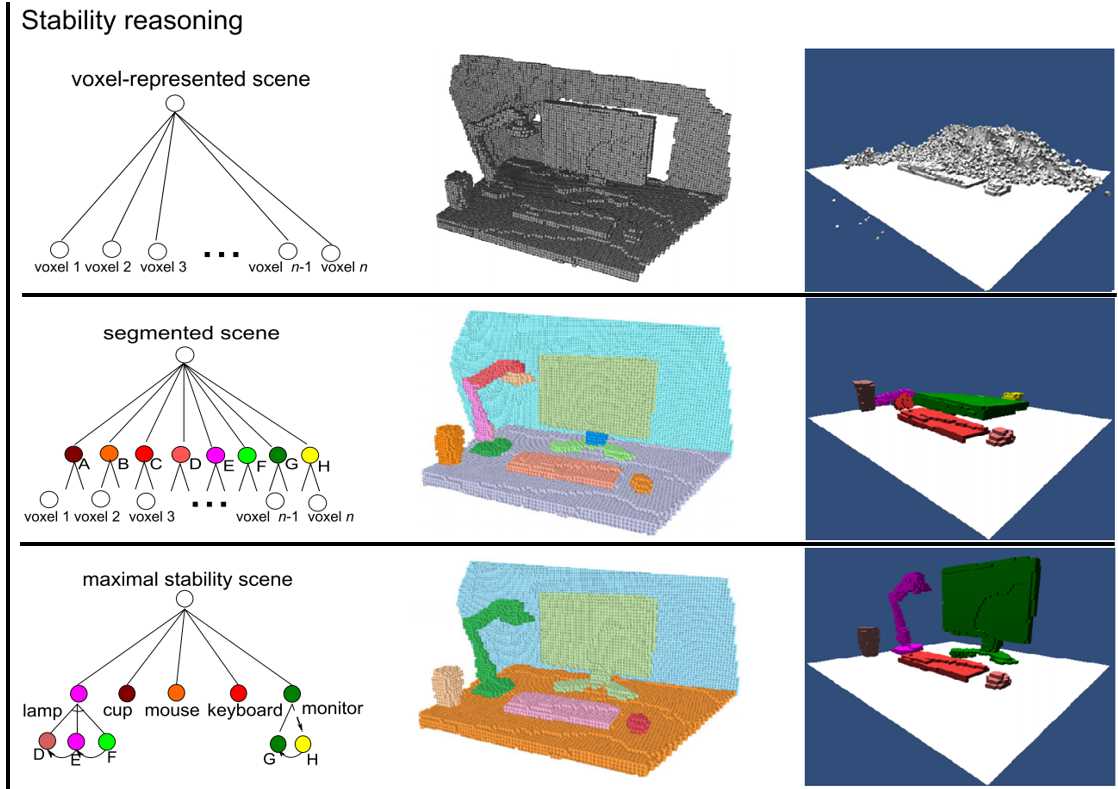}
	\caption{}
	\end{subfigure}
	\caption{An example explicitly exploiting safety and stability in a 3D scene-understanding task. Good performance in this task means that the system can understand the ``dark'' aspects of the image, which include how likely each object is to fall, and where the likely cause of falling will come from. (a) Input: reconstructed 3D scene. Output: parsed and segmented 3D scene comprised of stable objects. The numbers are ``unsafety'' scores for each object with respect to the disturbance field (represented by red arrows). (b) Scene-parsing graphs corresponding to three bottom-up processes: voxel-based representation (top), geometric pre-process, including segmentation and volumetric completion (middle), and stability optimization (bottom). Reproduced from Ref.~\citep{zheng2015scene} with permission of Springer Science+Business Media New York, \textcopyright~2015.}
	\label{fig:stability}
\end{figure*}

Classic computer vision studies focus on reasoning about appearance and geometry---the highly visible, pixel-represented aspects of images. Statistical modeling~\citep{mumford2010pattern} aims to capture the ``patterns generated by the world in any modality, with all their naturally occurring complexity and ambiguity, with the goal of reconstructing the processes, objects and events that produced them~\citep{mumford1994pattern}.'' Marr conjectured that the perception of a 2D image is an \emph{explicit} multiphase information process~\citep{marr1982vision}, involving (i) an early vision system for perceiving~\citep{julesz1962visual,zhu1998filters} and textons~\citep{julesz1981textons,zhu2005textons} to form a primal sketch~\citep{guo2003towards,guo2007primal}; (ii) a mid-level vision system to form 2.1D~\citep{nitzberg19902,wang1993layered,wang1994representing} and 2.5D~\citep{marr1978representation} sketches; and (iii) a high-level vision system in charge of full 3D scene formation~\citep{binford1971visual,brooks1981symbolic,kanade1981recovery}. In particular, Marr highlighted the importance of different levels of organization and the internal representation~\citep{broadbent1985question}.

Alternatively, perceptual organization~\citep{lowe2012perceptual,pentland1987perceptual} and Gestalt laws~\citep{wertheimer1912experimentelle,wagemans2012century,wagemans2012century2,kohler1920physischen,kohler1938physical,wertheimer1923untersuchungen,wertheimer1938laws,koffka2013principles} aim to resolve the 3D reconstruction problem from a single RGB image without considering depth. Instead, they use priors---groupings and structural cues~\citep{waltz1975understanding,barrow1981interpreting} that are likely to be invariant over wide ranges of viewpoints~\citep{lowe1987three}---resulting in feature-based approaches~\citep{lowe2004distinctive,dalal2005histograms}.

However, both appearance~\citep{solso2005cognitive} and geometric~\citep{hartley2003multiple} approaches have well-known difficulties resolving ambiguities. In addressing this challenge, modern computer vision systems have started to account for ``dark'' aspects of images by incorporating physics; as a result, they have demonstrated dramatic improvements over prior works. In certain cases, ambiguities have been shown to be extremely difficult to resolve through current state-of-the-art data-driven classification methods, indicating the significance of ``dark'' physical cues and signals in our ability to correctly perceive and operate within our daily environments; see examples in \cref{fig:scene_functionality_physics}~\citep{chen2019holistic}, where systems perceive which objects must rest on each other in order to be stable in a typical office space.

Through modeling and adopting physics into computer vision algorithms, the following two problems have been broadly studied:
\begin{enumerate}[leftmargin=*,noitemsep,nolistsep]
	\item Stability and safety in scene understanding. As demonstrated in Ref.~\citep{zheng2015scene}, this line of work is mainly based on a simple but crucial observation in human-made environments: by human design, objects in static scenes should be stable in the gravity field and be safe with respect to various physical disturbances. Such an assumption poses key constraints for physically plausible interpretation in scene understanding.
	\item Physical relationships in 3D scenes. Humans excel in reasoning about the physical relationships in a 3D scene, such as which objects support, attach, or hang from one another. As shown in Ref.~\citep{huang2018holistic}, those relationships represent a deeper understanding of 3D scenes beyond observable pixels that could benefit a wide range of applications in robotics, virtual reality (VR), and augmented reality (AR).
\end{enumerate}

The idea of incorporating physics to address vision problems can be traced back to Helmholtz and his argument for the ``unconscious inference'' of probable causes of sensory input as part of the formation of visual impressions~\citep{dayan1995helmholtz}. The very first such formal solution in computer vision dates back to Roberts' solutions for the parsing and reconstruction of a 3D block world in 1963~\citep{roberts1963machine}. This work inspired later researchers to realize the importance of both the violation of physical laws for scene understanding~\citep{biederman1982scene} and stability in generic robot manipulation tasks~\citep{blum1970stability,brand1995seeing}.

Integrating physics into scene parsing and reconstruction was revisited in the 2010s, bringing it into modern computer vision systems and methods. From a single RGB image, Gupta \etal proposed a qualitative physical representation for indoor~\citep{gupta2010estimating,gupta20113d} and outdoor~\citep{gupta2010blocks} scenes, where an algorithm infers the volumetric shapes of objects and relationships (such as occlusion and support) in describing 3D structure and mechanical configurations. In the next few years, other work~\citep{hedau2009recovering,lee2009geometric,hedau2012recovering,silberman2012indoor,schwing2012efficient,jia20133d,schwing2013box,guo2013support,shao2014imagining,zhao2013scene} also integrated the inference of physical relationships for various scene understanding tasks. In the past two years, Liu \etal~\citep{liu2018single} inferred physical relationships in joint semantic segmentation and 3D reconstruction of outdoor scenes. Huang \etal~\citep{huang2018holistic} modeled support relationships as edges in a human-centric scene graphical model, inferred the relationships by minimizing supporting energies among objects and the room layout, and enforced physical stability and plausibility by penalizing the intersections among reconstructed 3D objects and room layout~\citep{huang2018cooperative,chen2019holistic}.

\setstretch{1}

The aforementioned recent work mostly adopts simple physics cues; that is, very limited (if any) physics-based simulation is applied. The first recent work that utilized an actual physics simulator in modern computer vision methods was proposed by Zheng \etal in 2013~\citep{zheng2013beyond,zheng2014detecting,zheng2015scene}. As shown in \cref{fig:stability}~\citep{zheng2015scene}, the proposed method first groups potentially unstable objects with stable ones by optimizing for stability in the scene prior. Then, it assigns an ``unsafety'' prediction score to each potentially unstable object by inferring hidden potential triggers of instability (the disturbance field). The result is a physically plausible scene interpretation (voxel segmentation). This line of work has been further explored by Du \etal~\citep{du2018learning} by integrating an end-to-end trainable network and synthetic data.

\begin{figure}[t!]
	\centering
	\begin{subfigure}[b]{0.58\linewidth}
	\includegraphics[width=\linewidth]{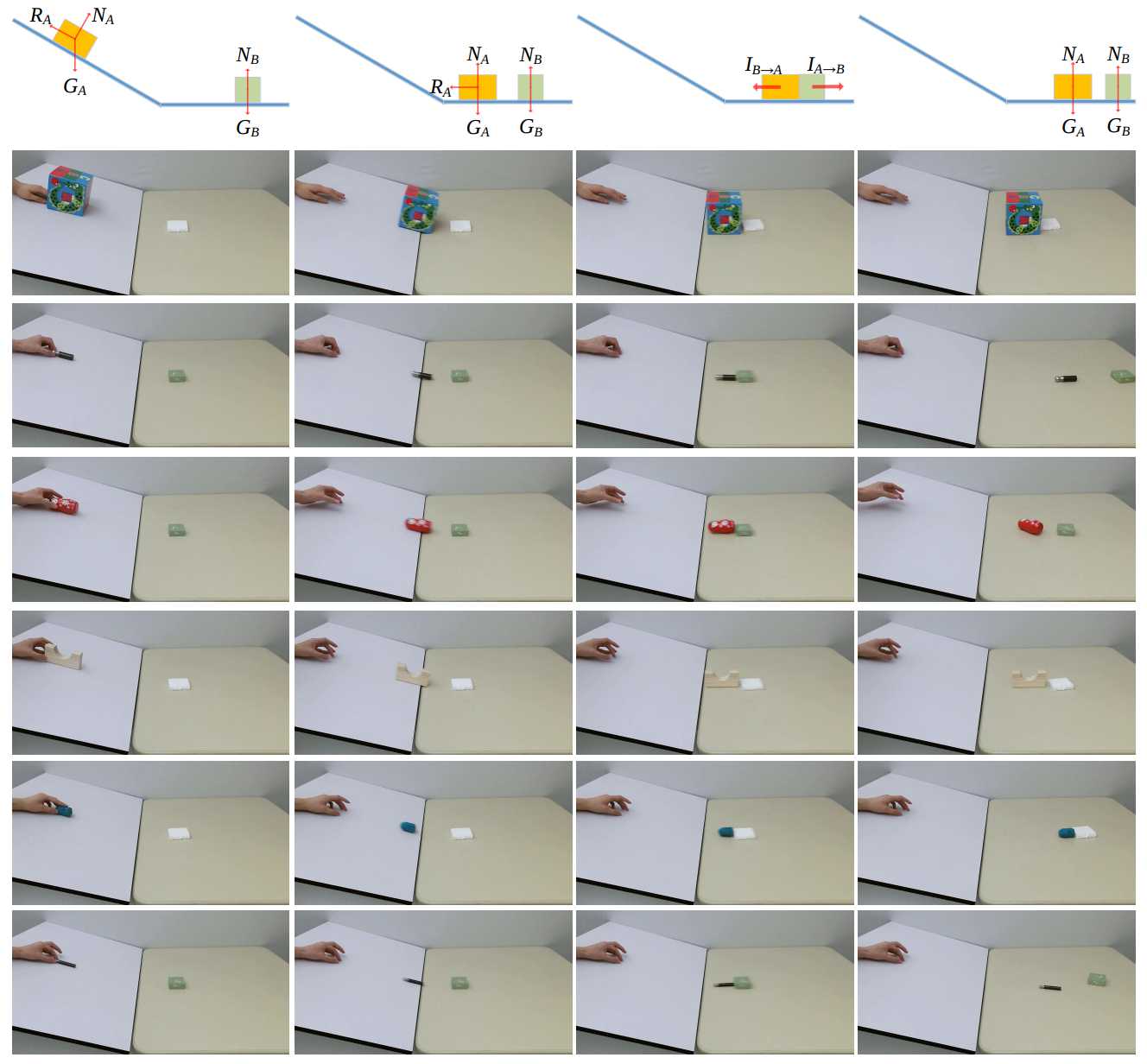}
	\caption{Snapshots of datasets}
	\end{subfigure}%
	\begin{subfigure}[b]{0.42\linewidth}
	\includegraphics[width=\linewidth]{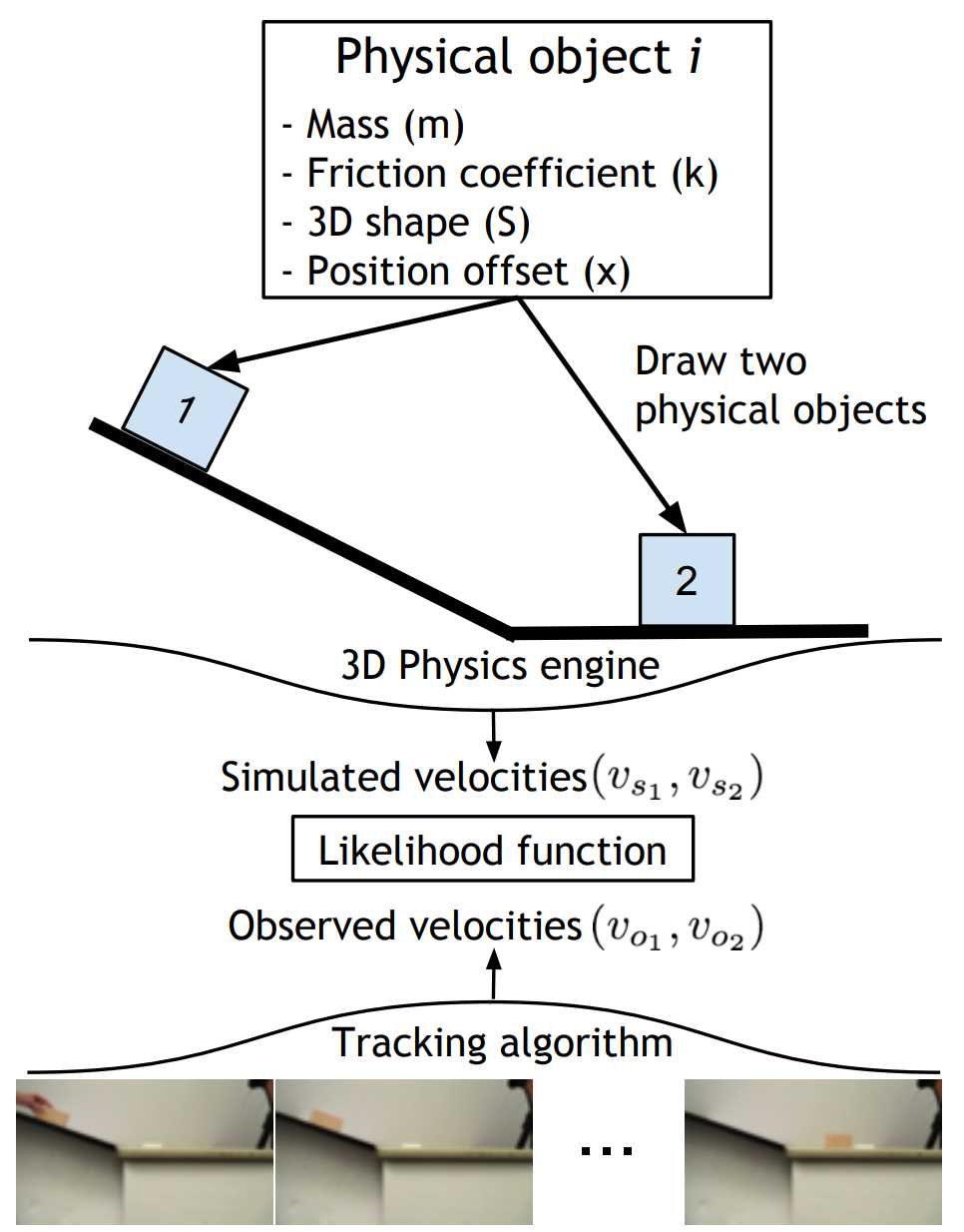}
	\caption{Galileo model}
	\end{subfigure}%
	\caption{Inferring the dynamics of the scenes. (a) Snapshots of the dataset; (b) overview of the Galileo model that estimates the physical properties of objects from visual inputs by incorporating the feedback of a physics engine in the loop. Reproduced from Ref.~\citep{wu2015galileo} with permission of Neural Information Processing Systems Foundation, Inc., \textcopyright~2015}
	\label{fig:galileo}
\end{figure}

Going beyond stability and support relationships, Wu \etal~\citep{wu2015galileo} integrated physics engines with deep learning to predict the future dynamic evolution of static scenes. Specifically, a generative model named Galileo was proposed for physical scene understanding using real-world videos and images. As shown in \cref{fig:galileo}, the core of the generative model is a 3D physics engine, operating on an object-based representation of physical properties including mass, position, 3D shape, and friction. The model can infer these latent properties using relatively brief runs of \ac{mcmc}, which drive simulations in the physics engine to fit key features of visual observations. Wu \etal~\citep{wu2016physics} further explored directly mapping visual inputs to physical properties, inverting a part of the generative process using deep learning. Object-centered physical properties such as mass, density, and the coefficient of restitution from unlabeled videos could be directly derived across various scenarios. With a new dataset named \emph{Physics 101} containing 17 408 video clips and 101 objects of various materials and appearances (\ie, shapes, colors, and sizes), the proposed unsupervised representation learning model, which explicitly encodes basic physical laws into the structure, can learn the physical properties of objects from videos.

\begin{figure}[t!]
	\centering
	\includegraphics[width=\linewidth]{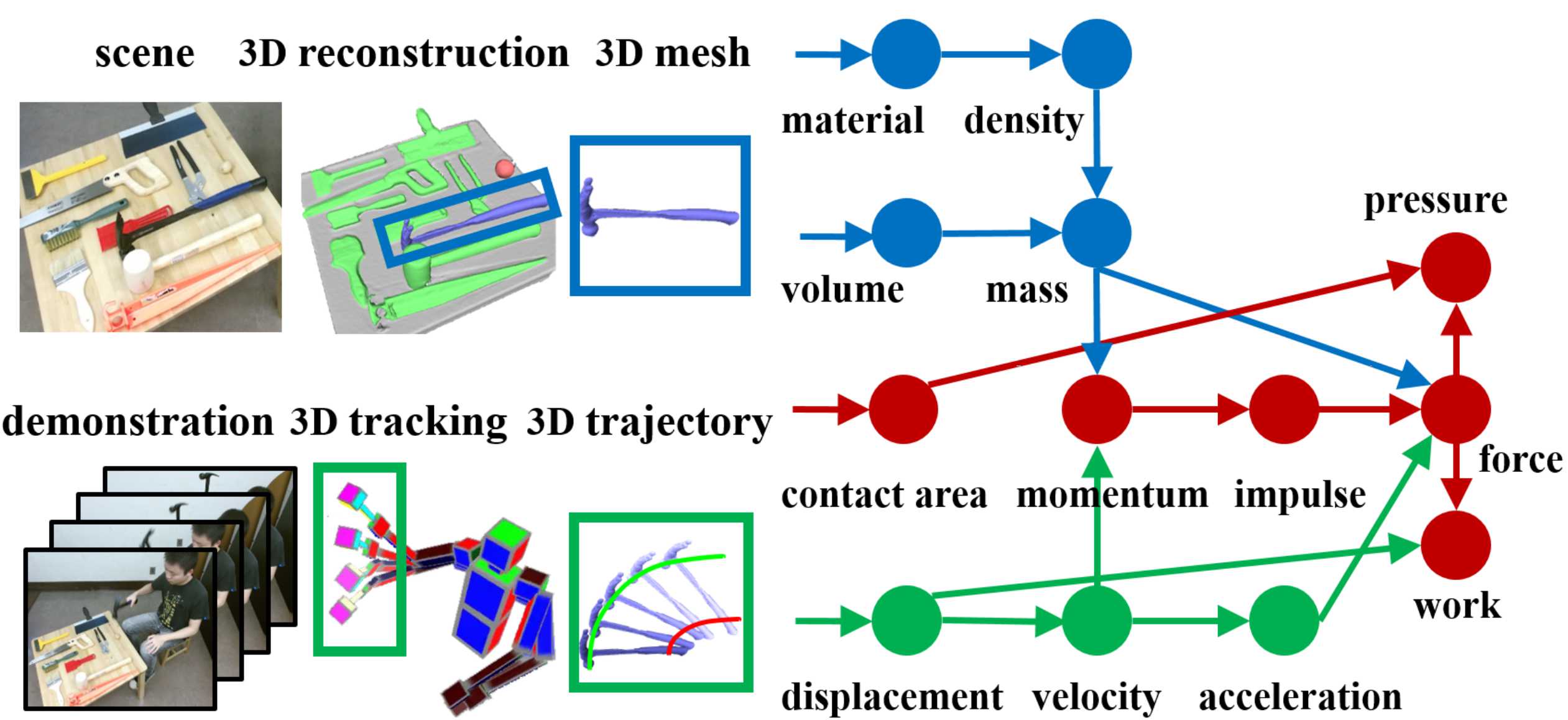}
	\caption{Thirteen physical concepts involved in tool use and their compositional relationships. By parsing a human demonstration, the physical concepts of material, volume, concept area, and displacement are estimated from 3D meshes of tool attributes (blue), trajectories of tool use (green), or both together (red). Higher level physical concepts can be further derived recursively. Reproduced from Ref.~\citep{zhu2015understanding} with permission of the authors, \textcopyright~2015.}
	\label{fig:physics_tool}
\end{figure}

Integrating physics and predicting future dynamics opens up quite a few interesting doors in computer vision. For example, given a human motion or task demonstration presented as a RGB-D image sequence, \etal~\citep{zhu2015understanding} built a system that calculated various physical concepts from just a single example of tool use (\cref{fig:physics_tool}), enabling it to reason about the essential physical concepts of the task (\eg, the force required to crack nuts). As the fidelity and complexity of the simulation increased, Zhu \etal~\citep{zhu2016inferring} were able to infer the forces impacting a seated human body, using a \ac{fem} to generate a mesh estimating the force on various body parts; \cref{fig:physics_chair}.

Physics-based reasoning can not only be applied to scene understanding tasks, as above, but have also been applied to pose and hand recognition and analysis tasks. For example, Brubaker \etal~\citep{brubaker2008kneed,brubaker2009estimating,brubaker2010physics} estimated the force of contacts and the torques of internal joints of human actions using a mass-spring system. Pham \etal~\citep{pham2015towards} further attempted to infer the forces of hand movements during human-object manipulation. In computer graphics, soft-body simulations based on video observation have been used to jointly track human hands and calculate the force of contacts~\citep{wang2013video,zhao2013robust}. Altogether, the laws of physics and how they relate to and among objects in a scene are critical ``dark'' matter for an intelligent agent to perceive and understand; some of the most promising computer vision methods outlined above have understood and incorporated this insight.

\section{Functionality and Affordance: The Opportunity for Task and Action}\label{sec:function}

Perception of an environment inevitably leads to a course of action~\citep{gibson1950perception,gibson1966senses}; Gibson argued that clues indicating opportunities for action in a nearby environment are perceived in a \emph{direct}, \emph{immediate} way with no sensory processing. This is particularly true for human-made objects and environments, as ``an object is first identified as having important functional relations'' and ``perceptual analysis is derived of the functional concept''~\citep{nelson1974concept}; for example, switches are clearly for flipping, buttons for pushing, knobs for turning, hooks for hanging, caps for rotating, handles for pulling, and so forth. This idea is the core of affordance theory~\citep{gibson1977theory}, which is based on Gestalt theory and has had a significant influence on how we consider visual perception and scene understanding.

\begin{figure}[t!]
	\centering
	\begin{subfigure}[b]{\linewidth}
	\includegraphics[width=\linewidth]{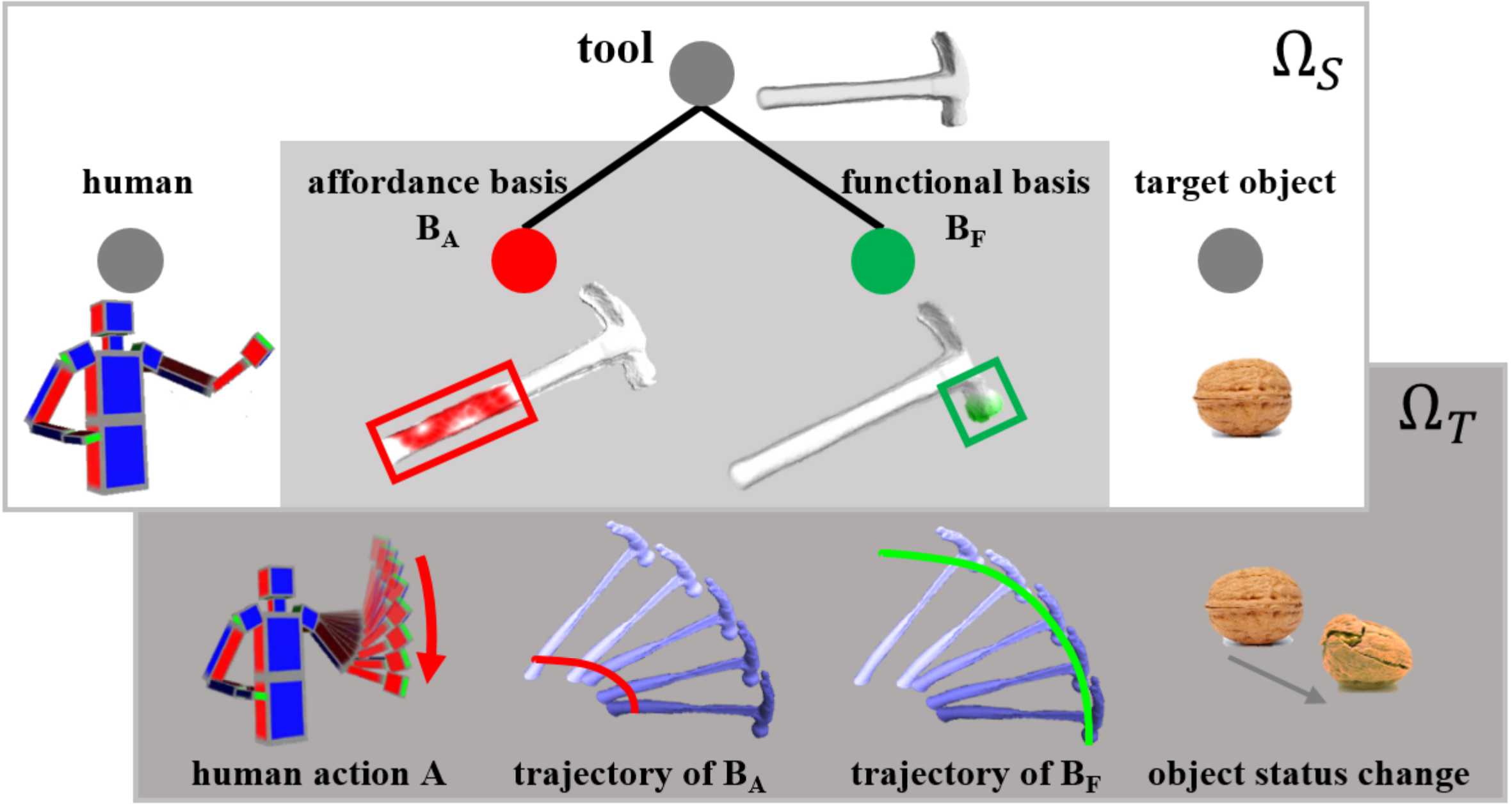}
	\caption{Functional basis and affordance basis in a tool-use example.}
	\end{subfigure}%
	\\
	\begin{subfigure}[b]{\linewidth}
	\includegraphics[width=\linewidth,trim={2cm 0.8cm 1cm 1cm},clip]{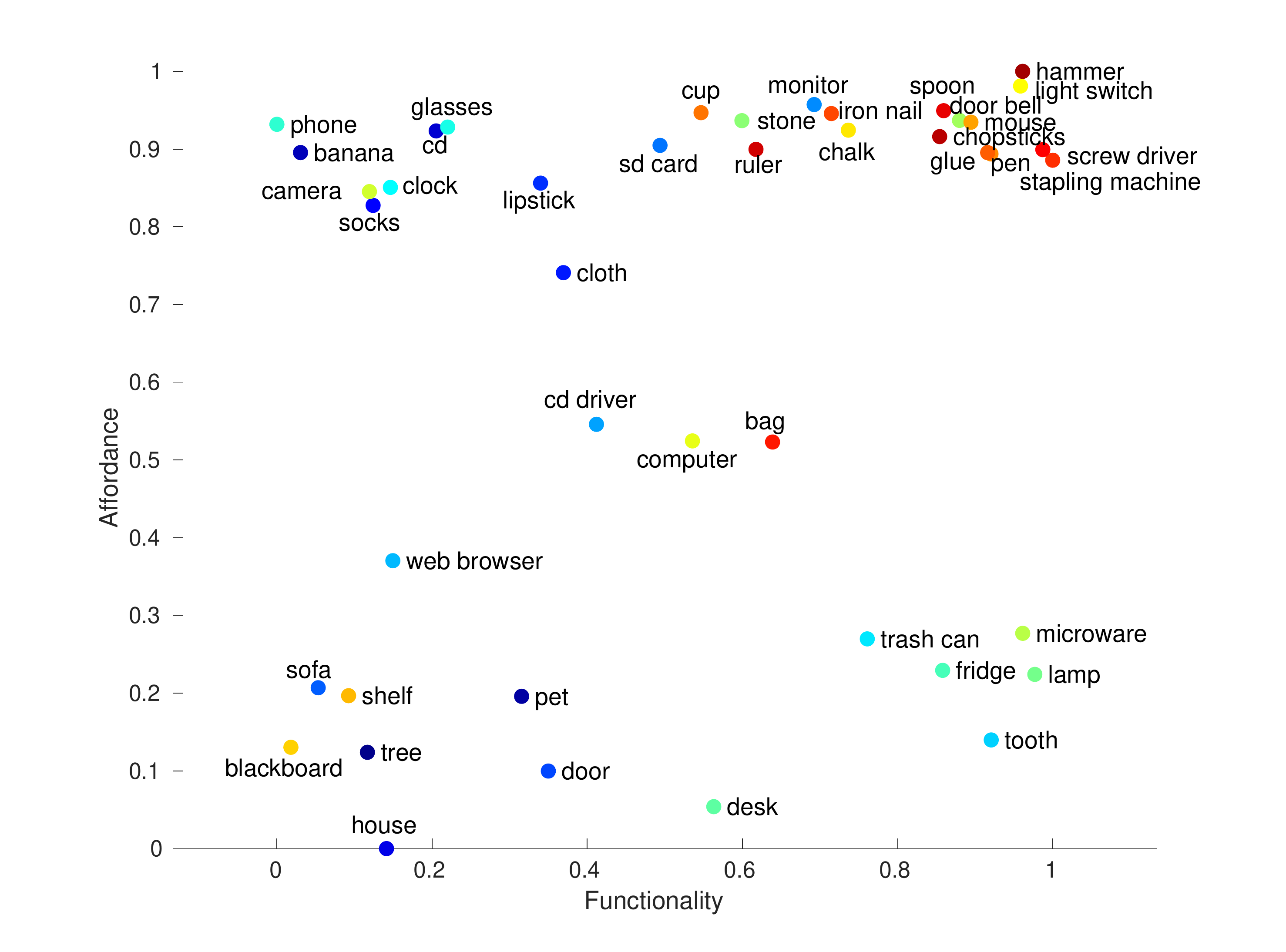}
	\caption{Examples of objects in the space spanned by functionality and affordance.}
	\end{subfigure}%
	\caption{(a) The task-oriented representation of a hammer and its use in cracking a nut in a joint spatiotemporal space. In this example, an object is decomposed into a functional basis and an affordance basis for a given task. (b) The likelihood of a common object being used as a tool based on its functionality and affordance. The warmer the color, the higher the probability. The functionality score is the average response to the question ``Can it be used to change the status of another object?'', and the affordance score is the average response to ``Can it be manipulated by hand?''}
	\label{fig:affordance_vs_functionality}
\end{figure}

Functional understanding of objects and scenes is rooted in identifying possible tasks that can be performed with an object~\citep{hassanin2018visual}. This is deeply related to the perception of causality, as covered in \cref{sec:causal}; to understand how an object can be used, an agent must understand what change of state will result if an object is interacted with in any way. While affordances depend directly on the actor, functionality is a permanent property of an object independent of the characteristics of the user; see an illustration of this distinction in \cref{fig:affordance_vs_functionality}. These two interweaving concepts are more invariant for object and scene understanding than their geometric and appearance aspects. Specifically, we argue that:
\begin{enumerate}[leftmargin=*,noitemsep,nolistsep]
	\item Objects, especially human-made ones, are defined by their functions, or by the actions they are associated with;
	\item Scenes, especially human-made ones, are defined by the actions than can be performed within them.
\end{enumerate}

Functionality and affordance are interdisciplinary topics and have been reviewed from different perspectives in the literature (\eg, Ref.~\citep{min2016affordance}). In this section, we emphasize the importance of incorporating functionality and affordance in the field of computer vision and \ac{ai} by starting with a case study of tool use in animal cognition. A review of functionality and affordance in computer vision follows, from both the object level and scene level. At the end, we review some recent literature in robotic manipulation that focuses on identifying the functionality and affordance of objects, which complements previous reviews of data-driven approaches~\citep{bohg2013data} and affordance tasks~\citep{yamanobe2017brief}.

\subsection{Revelation from Tool Use in Animal Cognition}

The ability to use an object as a tool to alter another object and accomplish a task has traditionally been regarded as an indicator of intelligence and complex cognition, separating humans from other animals~\citep{kohler1925mentality,thorpe1956learning}. Researchers commonly viewed tool use as the hallmark of human intelligence~\citep{oakley1968man} until relatively recently, when Dr. Jane Goodall observed wild chimpanzees manufacturing and using tools with regularity~\citep{goodall1986chimpanzees,whiten1999cultures,byrne1990machiavellian}. Further studies have since reported on tool use by other species in addition to chimpanzees. For example, Santos \etal~\citep{sabbatini2014sequential} trained two species of monkeys to choose between two canes to reach food under a variety of conditions involving different types of physical concepts (\eg, materials, connectivity, and gravity). Hunt~\citep{hunt1996manufacture} and Weir \etal~\citep{weir2002shaping} reported that New Caledonian crows can bend a piece of straight wire into a hook and use it to lift a bucket containing food from a vertical pipe. More recent studies also found that New Caledonian crows behave optimistically after using tools~\citep{mccoy2019new}. Effort cannot explain their optimism; instead, they appear to enjoy or be intrinsically motivated by tool use.

These discoveries suggest that some animals have the capability (and possibly the intrinsic motivation) to reason about the functional properties of tools. They can infer and analyze physical concepts and causal relationships of tools to approach a novel task using domain-general cognitive mechanisms, despite huge variety in their visual appearance and geometric features. Tool use is of particular interest and poses two major challenges in comparative cognition~\citep{beck1980animal}, which further challenges the reasoning ability of computer vision and \ac{ai} systems.

\begin{figure}[t!]
	\centering
	\begin{subfigure}[b]{0.475\linewidth}
	\includegraphics[width=\linewidth]{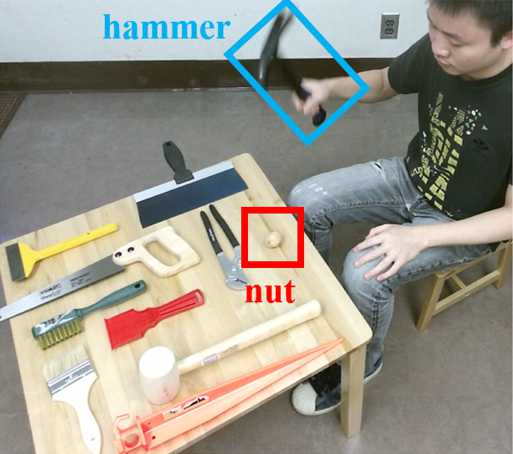}
	\caption{Learning}
	\end{subfigure}%
	\begin{subfigure}[b]{0.521\linewidth}
	\includegraphics[width=\linewidth]{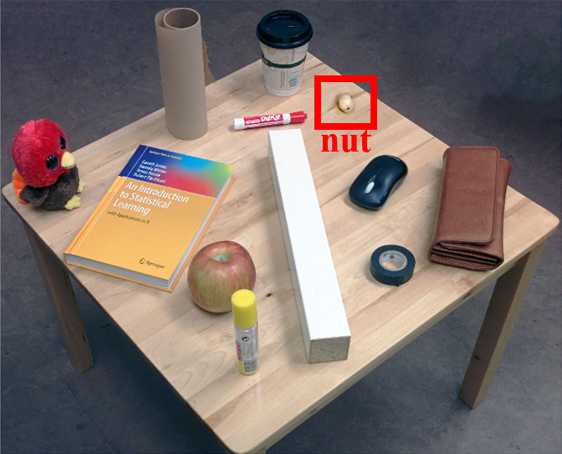}
	\caption{Inference}
	\end{subfigure}%
	\caption{Finding the right tools in novel situations. (a) In a learning phase, a rational human charged with cracking a nut is observed examining a hammer and other tools; (b) in an inference phase, the algorithm is asked to pick the best object on the table (\ie, the wooden leg) for the same task. This generalization entails reasoning about functionality, physics, and causal relationships among objects, actions, and overall tasks. Reproduced from Ref.~\citep{zhu2015understanding} with permission of the authors, \textcopyright~2015.}
	\label{fig:tool}
\end{figure}

First, why can some species devise innovative solutions, while others facing the same situation cannot? Look at the example in \cref{fig:tool}~\citep{zhu2015understanding}: by observing only a single demonstration of a person achieving the complex task of cracking a nut, we humans can effortlessly reason about which of the potential candidates from a new set of random and very different objects is best capable of helping us complete the same task. Reasoning across such large intraclass variance is extremely difficult to capture and describe for modern computer vision and \ac{ai} systems. Without a consistent visual pattern, properly identifying tools for a given task is a long-tail visual recognition problem. Moreover, the very same object can serve multiple functions depending on task context and requirements. Such an object is no longer defined by its conventional name (\ie, a hammer); instead, it is defined by its functionality.

Second, how can this functional reasoning capability emerge if one does not possess it innately? New Caledonian crows are well-known for their propensity and dexterity at making and using tools; meanwhile, although a crow's distant cousin, the rook, is able to reason and use tools in a lab setting, even \emph{they} do not use tools in the wild~\citep{bird2009insightful}. These findings suggest that the ability to represent tools may be more of a domain-general cognitive capacity based on functional reasoning than an adaptive specialization.

\subsection{Perceiving Functionality and Affordance}

\begin{quote}
	``\emph{The theory of affordances rescues us from the philosophical muddle of assuming fixed classes of objects, each defined by its common feature and then give a name \ldots You do not have to classify and label things in order to perceive what they afford \ldots It is never necessary to distinguish all the features of an object and, in fact, it would be impossible to do so.}''
	
	\hfill --- J. J. Gibson, 1977~\citep{gibson1977theory}
\end{quote}

The idea to incorporate functionality and affordance into computer vision and \ac{ai} can be dated back to the second International Joint Conference on Artificial Intelligence (IJCAI) in 1971, where Freeman and Newell~\citep{freeman1971model} argued that available structures should be described in terms of functions provided and functions performed. The concept of affordance was later coined by Gibson~\citep{gibson1977theory}. Based on the classic geometry-based ``arch-learning'' program~\citep{winston1970learning}, Winston \etal~\citep{winston1983learning} discussed the use of function-based descriptions of object categories. They pointed out that it is possible to use a single functional description to represent all possible cups, despite there being an infinite number of individual physical descriptions of cups or many other objects. In their ``mechanic's mate'' system~\citep{brady1984advances}, Connell and Brady~\citep{connell1987generating} proposed semantic net descriptions based on 2D shapes together with a generalized structural description. ``Chair'' and ``tool,'' exemplary categories researchers used for studies in functionality and affordance, were first systematically discussed alongside a computational method by Ho~\citep{ho1987representing} and DiManzo \etal~\citep{dimanzo1989understanding}, respectively. Inspired by the functional aspect of the ``chair'' category in Minsky's book~\citep{minsky1988society}, the first work that uses a purely functional-based definition of an object category (\ie, no explicit geometric or structural model) was proposed by Stark and Bowyer~\citep{stark1991achieving}. These early ideas of integrating functionality and affordance with computer vision and \ac{ai} systems have been modernized in the past decade; below, we review some representative topics.

\begin{figure}[t!]
	\centering
	\includegraphics[width=\linewidth]{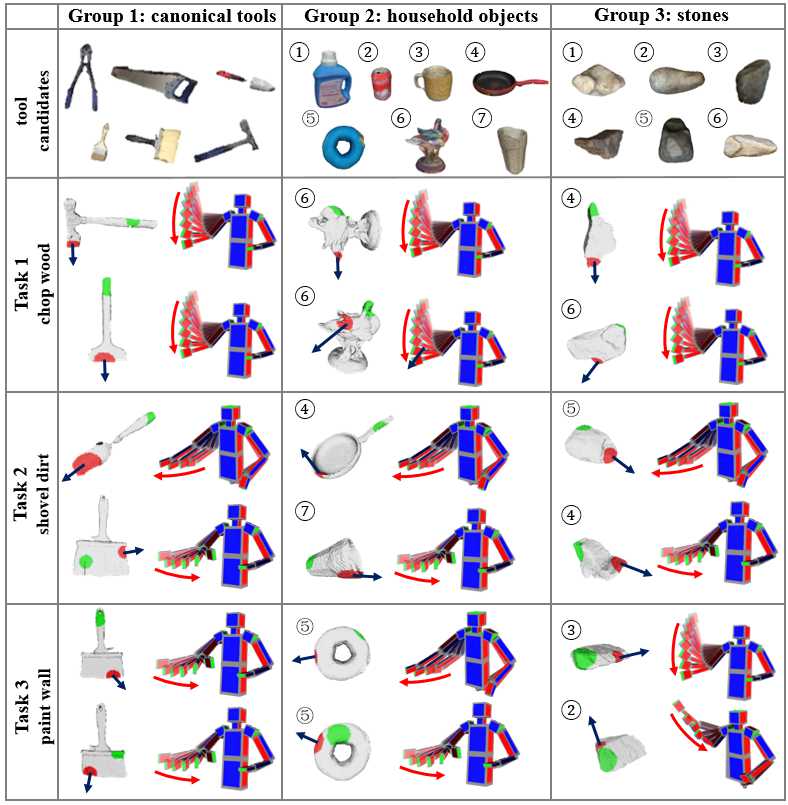}
	\caption{Given the three tasks of chopping wood, shoveling dirt, and painting a wall, an algorithm proposed by Zhu \etal~\citep{zhu2015understanding} picks and ranks objects within groups in terms of which object in each group is the best fit for task performance: conventional tools, household objects, and stones. Second, the algorithm outputs the imagined use of each tool, providing an affordance basis (the green spot indicating where the tool would be grasped by hand), a functional basis (the red area indicating the part of the tool that would make contact with the object), and the imagined sequence of poses constituting the movement of the action itself. Reproduced from Ref.~\citep{zhu2015understanding} with permission of the authors, \textcopyright~2015.}
	\label{fig:tool_inference}
\end{figure}

\emph{``Tool''} is of particular interest in computer vision and robotics, partly due to its nature as an object for changing \emph{other} objects' status. Motivated by the studies of tool use in animal cognition, Zhu \etal~\citep{zhu2015understanding} cast the tool understanding problem as a \emph{task-oriented} object-recognition problem, the core of which is understanding an object's underlying functions, physics, and causality. As shown in \cref{fig:tool_inference}~\citep{zhu2015understanding}, a tool is a physical object (\eg, a hammer or a shovel) that is used through action to achieve a task. From this new perspective, any object can be viewed as a hammer or a shovel. This generative representation allows computer vision and \ac{ai} algorithms to reason about the underlying mechanisms of various tasks and generalize object recognition across novel functions and situations. This perspective goes beyond memorizing examples for each object category, which tends to prevail among traditional appearance-based approaches in the literature. Combining both physical and geometric aspects, Liu \etal~\citep{liu2018physical} took the decomposition of physical primitives for tool recognition and tower stability further.

\begin{figure}[t!]
	\centering
	\begin{subfigure}[b]{\linewidth}
	\includegraphics[width=\linewidth]{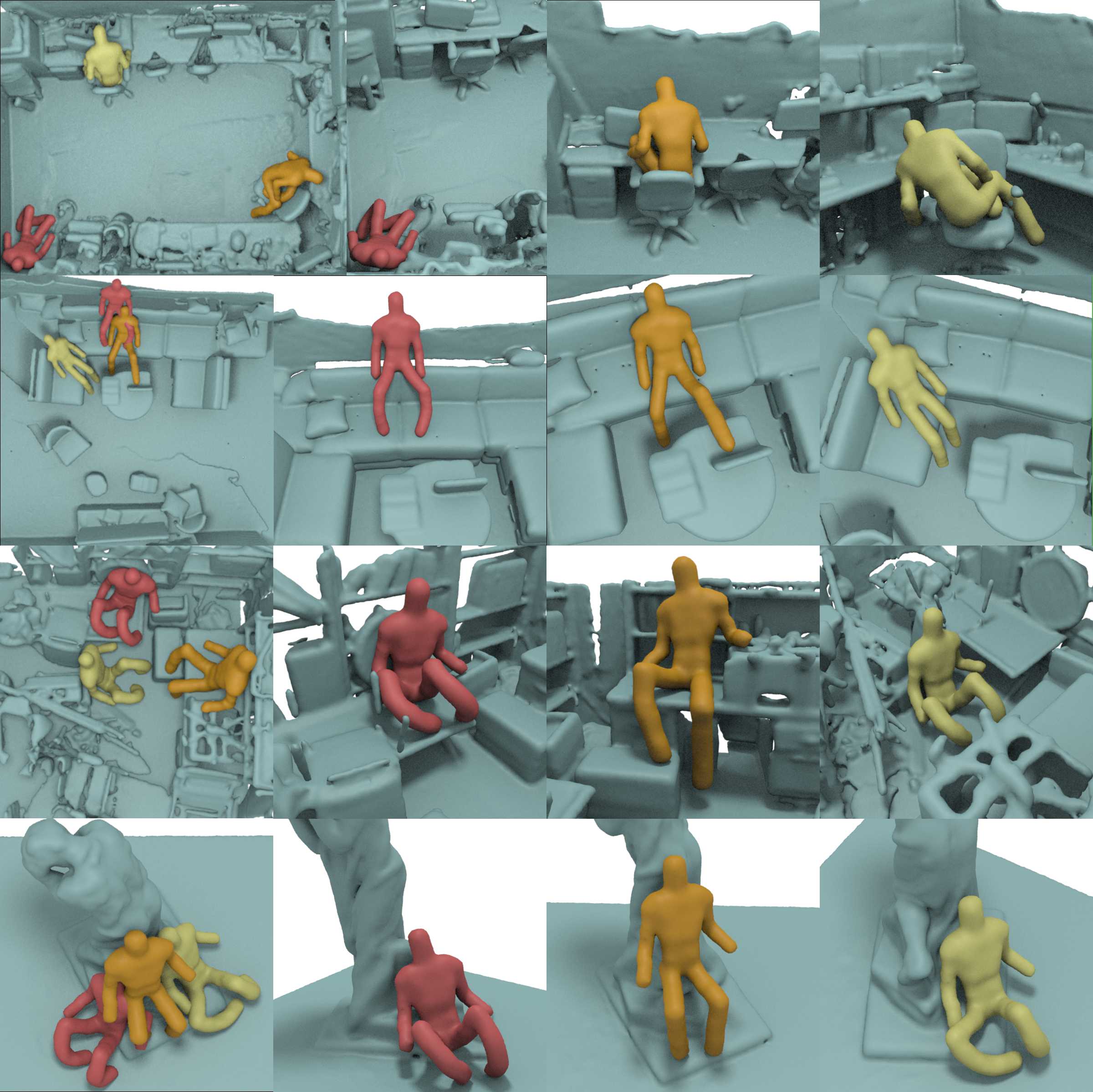}
	\end{subfigure}%
	\\
	\begin{subfigure}[b]{0.25\linewidth}
	\includegraphics[width=\linewidth]{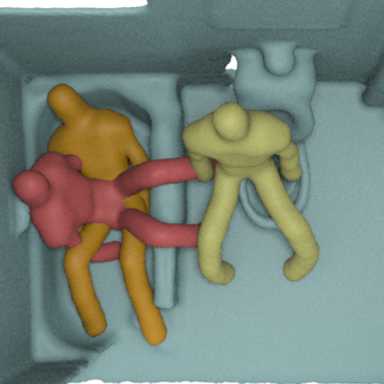}
	\caption{}
	\end{subfigure}%
	\begin{subfigure}[b]{0.25\linewidth}
	\includegraphics[width=\linewidth]{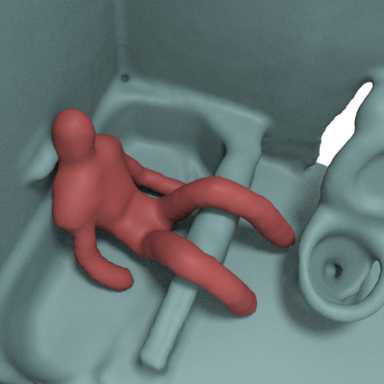}
	\caption{}
	\end{subfigure}%
	\begin{subfigure}[b]{0.25\linewidth}
	\includegraphics[width=\linewidth]{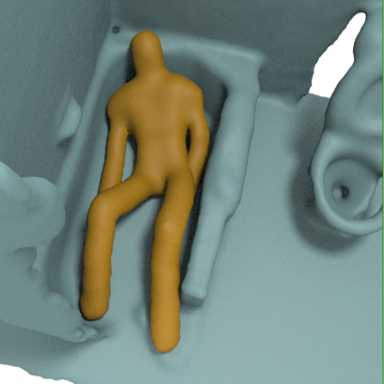}
	\caption{}
	\end{subfigure}%
	\begin{subfigure}[b]{0.25\linewidth}
	\includegraphics[width=\linewidth]{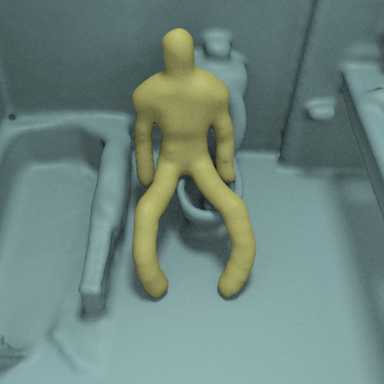}
	\caption{}
	\end{subfigure}%
	\hfill
	\caption{(a) Top three poses in various scenes for affordance (sitting) recognition. The zoom-in shows views of the (b) best, (c) second-best, and (d) third-best choice of sitting poses. The top two rows are canonical scenarios, the middle row is a cluttered scenario, and the bottom two rows are novel scenarios that demonstrated significant generalization and transfer capability. Reproduced from Ref.~\citep{zhu2016inferring} with permission of the authors, \textcopyright~2016.}
	\label{fig:force_sitting}
\end{figure}

\emph{``Container''} is ubiquitous in daily life and is considered a half-tool~\citep{baber2003cognition}. The study of containers can be traced back to a series of studies by Inhelder and Piaget in 1958~\citep{inhelder1958growth}, in which they showed six-year-old children could still be confused by the complex phenomenon of pouring liquid into containers. Container and containment relationships are of particular interest in \ac{ai}, computer vision, and psychology due to the fact that it is one of the earliest spatial relationships to be learned, preceding other common ones \eg, occlusions~\citep{strickland2015visual} and support relationships~\citep{casasola2002infant}). As early as two and a half months old, infants can already understand containers and containment~\citep{hespos2001reasoning,wang2005detecting,hespos2007precursors}. In the \ac{ai} community, researchers have been adopting commonsense reasoning~\citep{davis2017commonsense,davis2011does,davis2008pouring} and qualitative representation~\citep{cohn1997qualitative,cohn2001qualitative} for reasoning about container and containment relationships, mostly focusing on ontology, topology, first-order logic, and knowledge base.

\begin{figure}[t!]
	\centering
	\includegraphics[width=\linewidth]{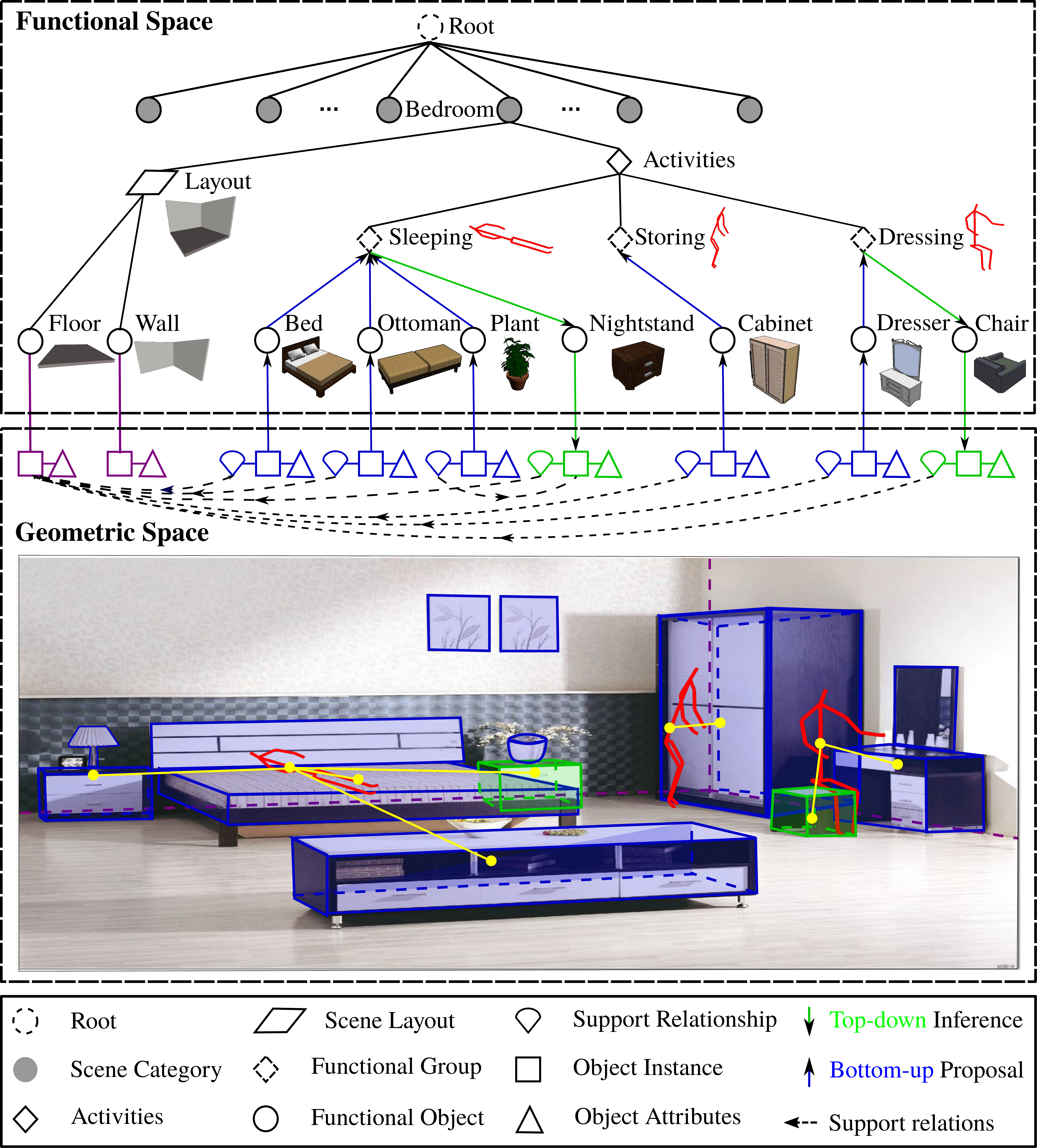}
	\caption{Task-centered representation of an indoor scene. The functional space exhibits a hierarchical structure, and the geometric space encodes the spatial entities with contextual relationships. The objects are grouped by their hidden activity, \ie, by latent human context or action. Reproduced from Ref.~\citep{huang2018holistic} with permission of the authors, \textcopyright~2018.}
	\label{fig:scene_functionality}
\end{figure}

More recently, physical cues and signals have been demonstrated to strongly facilitate reasoning about functionality and affordance in container and containment relationships. For example, Liang \etal~\citep{liang2015evaluating} demonstrated that a physics-based simulation is robust and transferable for identifying containers in response to three questions: ``What is a container?'', ``Will an object contain another?'', and ``How many objects will a container hold?'' Liang's approach performed better than approaches using features extracted from appearance and geometry for the same problem. This line of research aligns with the recent findings of intuitive physics in psychology~\citep{battaglia2013simulation,smith2013consistent,bates2015humans,kubricht2016probabilistic,kubricht2017consistent,kubricht2017intuitive}, and enabled a few interesting new directions and applications in computer vision, including reasoning about liquid transfer~\citep{yu2015fill,mottaghi2017see}, container and containment relationships~\citep{liang2016inferring}, and object tracking by utilizing containment constraints~\citep{liang2018tracking}.

\emph{``Chair''} is an exemplar class for affordance; the latest studies on object affordance include reasoning about both geometry and function, thereby achieving better generalizations for unseen instances than conventional, appearance-based, or geometry-based machine learning approaches. In particular, Grabner \etal~\citep{grabner2011makes} designed an ``affordance detector'' for chairs by fitting typical human sitting poses onto 3D objects. Going beyond visible geometric compatibility, through physics-based simulation, Zhu \etal~\citep{zhu2016inferring} inferred the forces/pressures applied to various body parts while sitting on different chairs; see \cref{fig:force_sitting}~\citep{zhu2016inferring} for more information. Their system is able to ``feel,'' in numerical terms, discomfort when the forces/pressures on body parts exceed certain comfort intervals.

\setstretch{0.98}

\begin{figure}[t!]
	\centering
	\includegraphics[width=\linewidth]{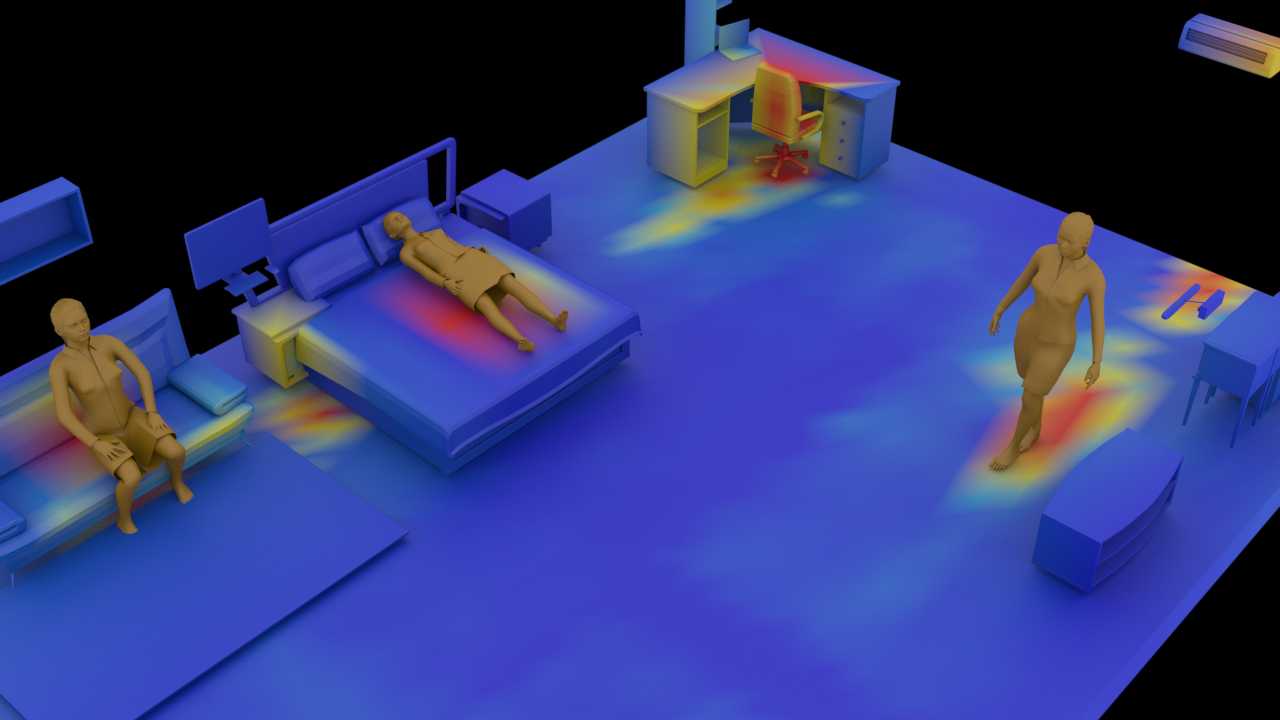}
	\caption{An example of a synthesized human-centric indoor scene (a bedroom) with an affordance heat map generated by Refs.~\citep{qi2018human,jiang2018configurable}. The joint sampling of the scene was achieved by alternatively sampling humans and objects according to a joint probability distribution.}
	\label{fig:scene_synthesis}
\end{figure}

\emph{``Human''} context has proven to be a critical component in modeling the constraints on possible usage of objects in a scene. In approaching this kind of problem, all methods imagine different potential human positioning relative to objects to help parse and understand the visible elements of the scene. The fundamental reason for this approach is that human-made scenes are functional spaces that serve human activities, whose objects exist primarily to assist human actions~\citep{gibson1977theory}. Working at the object level, Jiang \etal proposed methods that use human context to learn object arrangement~\citep{jiang2012learning} and object labeling~\citep{jiang2013hallucinated}. At the scene level, Zhao and Zhu~\citep{zhao2013scene} modeled functionality in 3D scenes through the compositional and contextual relationships among objects within them. To further explore the hidden human context pervading 3D scenes, Huang \etal~\citep{huang2018holistic} proposed a stochastic method to parse and reconstruct scenes with a \acf{hsg}. \ac{hsg} describes a functional, task-centered representation of scenes. As shown in \cref{fig:scene_functionality}~\citep{huang2018holistic}, the descriptor was composed of functional scene categories, task-centered activity groups, and individual objects. In a reversal of the process of parsing scenes using human context, scene functionality could also be used to synthesize \emph{new} scenes with humanlike object arrangements: Qi \etal~\citep{qi2018human} and Jiang \etal\citep{jiang2018configurable} proposed using human-centric representations to synthesize 3D scenes with a simulation engine. As illustrated in \cref{fig:scene_synthesis}~\citep{qi2018human,jiang2018configurable}, they integrated human activities with functional grouping/support relationships to build natural and fitting activity spaces.

\subsection{Mirroring: Causal-equivalent Functionality \& Affordance}\label{sec:functionality_robotics}

\begin{figure}[t!]
	\centering
	\begin{subfigure}[b]{0.38\linewidth}
	\includegraphics[width=\linewidth]{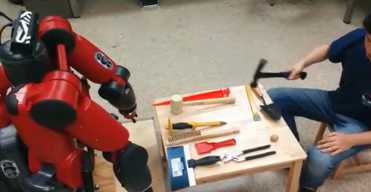}
	\caption{Demonstration}
	\end{subfigure}%
	\begin{subfigure}[b]{0.62\linewidth}
	\includegraphics[width=0.5\linewidth]{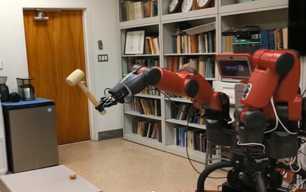}%
	\includegraphics[width=0.5\linewidth]{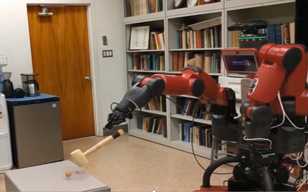}
	\caption{Failure by pure imitation}
	\end{subfigure}%
	\caption{(a) Given a successful human demonstration, (b) the robot may fail to accomplish the same task by imitating the human demonstration due to different embodiments. In this case, a two-finger gripper cannot firmly hold a hammer while swinging; the hammer slips, and the execution fails.}
	\label{fig:hammering}
\end{figure}

It is difficult to evaluate a computer vision or \ac{ai} system's facility at reasoning with functionality and affordance; unlike with causality and physics, not all systems will see functionality and affordance in the same way. Indeed, humans and robots have different morphology; therefore, the same object or environment does not necessarily introduce the same functionality and affordance to both robots and humans. For example, a human with five fingers can firmly grasp a hammer that a robot gripper with the typical two or three fingers might struggle to wield, as shown in \cref{fig:hammering}. In these cases, a system must reason about the underlying mechanisms of affordance, rather than simply mimicking the motions of a human demonstration. This common problem is known as the ``correspondence problem''~\citep{dautenhahn2002imitation} in \acf{lfd}; more details have been provided in two previous surveys~\citep{argall2009survey,osa2018algorithmic}.

Currently, the majority of work in \ac{lfd} uses a one-to-one mapping between human demonstration and robot execution, restricting the \ac{lfd} to mimicking the human's low-level motor controls and replicating a nearly identical procedure. Consequently, the ``correspondence problem'' is insufficiently addressed, and the acquired skills are difficult to adapt to new robots or new situations; thus, more robust solutions are necessary. To tackle these problems, we argue that the robot must obtain deeper understanding in functional and causal understanding of the manipulation, which demands more explicit modeling of knowledge about physical objects and forces. The key to imitating manipulation is using functionality and affordance to create causal-equivalent manipulation; in other words, replicating task execution by reasoning about contact forces, instead of simply repeating the precise trajectory of motion.

However, measuring human manipulation forces is difficult due to the lack of accurate instruments; there are constraints imposed on devices aimed at measuring natural hand motions. For example, a vision-based force-sensing method~\citep{pham2015towards} often cannot handle self-occlusions and occlusions caused during manipulations. Other force-sensing systems, such as strain gauge FlexForce~\citep{gu2015fine} or the liquid metal-embedded elastomer sensor~\citep{hammond2014toward} can be used in glove-like devices; but even they can be too rigid to conform to the contours of the hand, resulting in limitations on natural motion during attempts at fine manipulative action. Recently, Liu \etal~\citep{liu2017glove} introduced Velostat, a soft piezoresistive conductive film whose resistance changes under pressure. They used this material in an inertial measurement unit (IMU)-based position-sensing glove to reliably record manipulation demonstrations with fine-grained force information. This kind of measurement is particularly important for teaching systems to perform tasks with visually latent changes.

Consider the task of opening a medicine bottle with a child-safety locking mechanism. These bottles require the user to push or squeeze in specific places to unlock the cap. By design, attempts to open these bottles using a standard procedure will result in failure. Even if an agent visually observes a successful demonstration, attempted direct imitation will likely omit critical steps in the procedure, as the visual appearance of opening both medicine and traditional bottles are typically very similar if not identical. By using the Velostat~\citep{liu2017glove} glove in demonstration, the fine forces used to unlock the child-safety mechanism become observable. From these observations, Edmonds \etal~\citep{edmonds2017feeling,edmonds2019tale} taught an action planner through both a top-down stochastic grammar model to represent the compositional nature of the task sequence, and a bottom-up discriminative model using the observed poses and forces. These two inputs were combined during planning to select the next optimal action. An augmented reality (AR) interface was also developed on top of this work to improve system interpretability and allow for easy patching of robot knowledge~\citep{liu2018interactive}.

\begin{figure}[t!]
	\centering
	\includegraphics[width=\linewidth]{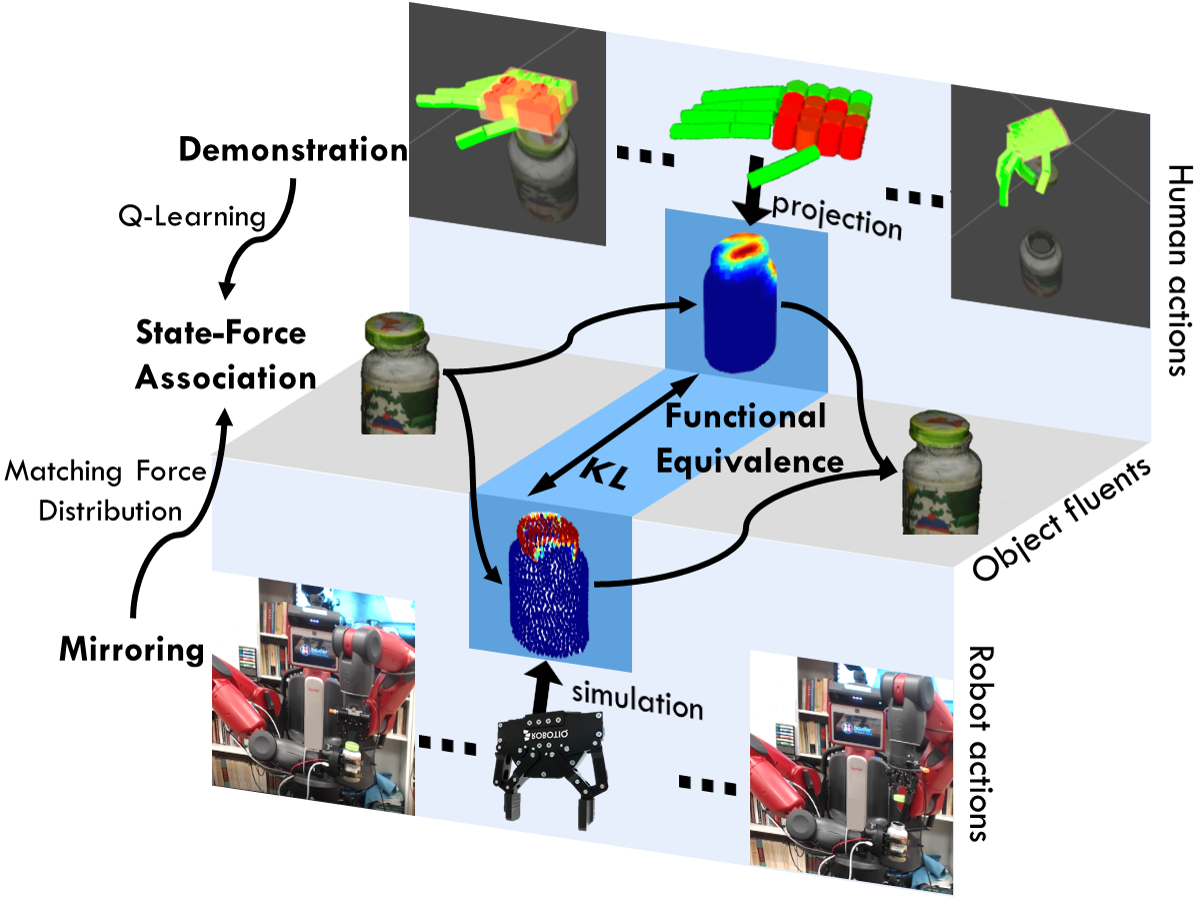}
	\caption{A robot mirrors human demonstrations with functional equivalence by inferring the action that produces similar force, resulting in similar changes in physical states. Q-learning is applied to similar types of forces with categories of object state changes to produce human–object-interaction (\emph{hoi}) units. KL: Kullback–Leibler divergence. Reproduced from Ref.~\citep{liu2019mirroring} with permission of Association for the Advancement of Artificial Intelligence, \textcopyright~2019.}
	\label{fig:mirroring}
\end{figure}

One major limitation of the above work is that the robot's actions are predefined, and the underlying structure of the task is not modeled. Recently, Liu \etal~\citep{liu2019mirroring} proposed a \emph{mirroring} approach and a concept of \emph{functional manipulation} that extends the current \ac{lfd} through a physics-based simulation to address the correspondence problem; see \cref{fig:mirroring}~\citep{liu2019mirroring} for more details. Rather than over-imitating the motion trajectories of the demonstration, the robot is encouraged to seek \emph{functionally equivalent} but possibly visually different actions that can produce the same effect and achieve the same goal as those in the demonstration. This approach has three characteristics distinguishing it from the standard \ac{lfd}. First, it is \emph{force-based}: these tactile perception-enabled demonstrations capture a deeper understanding of the physical world that a robot interacts with beyond visually observable space, providing an extra dimension that helps address the correspondence problem. Second, it is \emph{goal-oriented}: a ``goal'' is defined as the desired state of the target object and is encoded in a grammar model. The terminal node of the grammar model comprises the state changes caused by forces, independent of embodiments. Finally, this method uses \emph{mirroring without overimitation}: in contrast to the classic \ac{lfd}, a robot does not necessarily mimic every action in a human demonstration; instead, the robot reasons about the motion to achieve the goal states based on the learned grammar and simulated forces.

\setstretch{1}

\section{Perceiving Intent: The Sense of Agency}\label{sec:intention}

In addition to inanimate physical objects, we live in a world with a plethora of animate and goal-directed agents, whose agency implies the ability to perceive, plan, make decisions, and achieve goals. Crucially, such a sense of agency further entails (i) the \emph{intentionality}~\citep{dennett1989intentional} to represent a future goal state and equifinal variability~\citep{heider2013psychology} to be able to achieve the intended goal state with different actions across contexts; and (ii) the\emph{rationality of actions} in relation to goals~\citep{gergely1995taking} to devise the most efficient possible action plan. The perception and comprehension of intent enable humans to better understand and predict the behavior of other agents and engage with others in cooperative activities with shared goals. The construct of intent, as a basic organizing principle guiding how we interpret one another, has been increasingly granted a central position within accounts of human cognitive functioning, and thus should be an essential component of future \ac{ai}.

In \cref{sec:agency}, we start with a brief introduction to what constitutes the concepts of ``agency,'' which are deeply rooted in humans as young as six months old. Next, in \cref{sec:rationality}, we explain the \emph{rationality} principle as the mechanism with which both infants and adults perceive animate objects as intentional beings. We then describe how intent prediction is related to action prediction in modern computer vision and machine learning, but is in fact much more than predicting action labels; see \cref{sec:beyond_action} for a philosophical perspective. In \cref{sec:building_intention}, we conclude this section by providing a brief review of the building blocks for intent in computer vision and \ac{ai}.

\subsection{The Sense of Agency}\label{sec:agency}

In the literature, theory of mind (ToM) refers to the ability to attribute mental states, including beliefs, desires, and intentions, to oneself and others~\citep{premack1978does}. Perceiving and understanding an agent's intent based on their \emph{belief} and \emph{desire} is the ultimate goal, since people largely act to fulfill intentions arising from their beliefs and desires~\citep{baldwin2001discerning}.

Evidence from developmental psychology shows that six-month-old infants see human activities as goal-directed behavior~\citep{woodward1998infants}. By the age of 10 months, infants segment continuous behavior streams into units that correspond to what adults would see as separate goal-directed acts, rather than mere spatial or muscle movements~\citep{meltzoff2001like,baldwin2001infants}. After their first birthday, infants begin to understand that an actor may consider various plans to pursue a goal, and choose one to intentionally enact based on environmental reality~\citep{tomasello2005understanding}. Eighteen-month-old children are able to both \emph{infer} and \emph{imitate} the intended goal of an action even if the action repeatedly fails to achieve the goal~\citep{biro2007becoming}. Moreover, infants can imitate actions in a rational, efficient way based on an evaluation of the action's situational constraints instead of merely copying movements, indicating that infants have a deep understanding of relationships among the environment, action, and underlying intent~\citep{gergely2002developmental}. Infants can also perceive intentional relationships at varying levels of analysis, including concrete action goals, higher order plans, and collaborative goals~\citep{woodward2009emergence}.

Despite the complexity of the behavioral streams we actually witness, we readily process action in intentional terms from infancy onward~\citep{baldwin2001discerning}. It is underlying \emph{intent}, rather than surface behavior, that matters when we observe motions. One latent intention can make several highly dissimilar movement patterns conceptually cohesive. Even an identical physical movement could have a variety of different meanings depending on the intent motivating it; for example, the underlying intent driving a reach for a cup could be to either fill the cup or clean it. Thus, inference about others' intentions is what gives an observer the ``gist'' of human actions. Research has found that we do not encode the complete details of human motion in space; instead, we perceive motions in terms of intent. It is the constructed understanding of actions in terms of the actors' goals and intentions that humans encode in memory and later retrieve~\citep{baldwin2001discerning}. Reading intentions has even led to species-unique forms of cultural learning and cognition~\citep{tomasello2005understanding}. From infants to complex social institutions, our world is constituted of the intentions of its agents~\citep{tomasello1999developing,bloom1996intention,tomasello2005understanding}.

\subsection{From Animacy to Rationality}\label{sec:rationality}

\begin{figure}[t!]
	\centering
	\includegraphics[width=\linewidth]{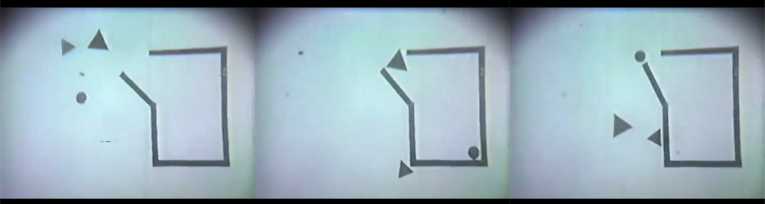}
	\caption{The seminal Heider–Simmel experiment~\citep{heider1944experimental}. Adults can perceive and attribute mental states from nothing but the motion of simple geometric shapes.}
	\label{fig:hs}
\end{figure}

Human vision has the uniquely social function of extracting latent mental states about goals, beliefs, and intentions from nothing but visual stimuli. Surprisingly, such visual stimuli do not need to contain rich semantics or visual features. An iconic illustration of this is the seminal Heider-Simmel display created in the 1940s~\citep{heider1944experimental}; see \cref{fig:hs} for more detail. Upon viewing the 2D motion of three simple geometric shapes roaming around a space, human participants acting without any additional hints automatically and even irresistibly perceive ``social agents,'' with a set of rich mental states such as goals, emotions, personalities, and coalitions. These mental states come together to form a story-like description of what is happening in the display, such as a hero saving a victim from a bully. Note that in this experiment, where no specific directions regarding perception of the objects were provided, participants still tended to describe the objects as having different sexes and dispositions. Another crucial observation is that human participants always reported the animated objects as ``opening'' or ``closing'' the door, similar to in Michotte's ``entrance'' display~\citep{michotte1963perception}; the movement of the animated object is imparted to the door through prolonged contact rather than through sudden impact. This interpretation of simple shapes as animated beings was a remarkable demonstration of how human vision is able to extract rich social relationships and mental states from sparse, symbolized inputs with extremely minimal visual features.

\setstretch{0.97}

In the original Heider-Simmel display, it is unclear whether the demonstrated visual perception of social relationships and mental states was attributable more or less to the dynamic motion of the stimuli, or to the relative attributes (size, shape, \etc) of the protagonists. Berry and Misovich~\citep{berry1994methodological} designed a quantitative evaluation of these two confounding variables by degrading the structural display while preserving its original dynamics. They reported a similar number of anthropomorphic terms as in the original design, indicating that the display's structural features are not the critical factors informing human social perception; this finding further strengthened the original finding that human perception of social relationships goes beyond visual features. Critically, when Berry and Misovich used static frames in both the original and degraded displays, the number of anthropomorphic terms dropped significantly, implying that the dynamic motion and temporal contingency were the crucial factors for the successful perception of social relationships and mental states. This phenomenon was later further studied by Bassili~\citep{bassili1976temporal} in a series of experiments.

Similar simulations of biologically meaningful motion sequences were produced by Dittrich and Lea~\citep{dittrich1994visual} in simple displays of moving letters. Participants were asked to identify one letter acting as a ``wolf'' chasing another ``sheep'' letter, or a ``lamb'' letter trying to catch up with its mother. These scholars' findings echoed the Heider-Simmel experiment; motion dynamics played an important factor in the perception of intentional action. Specifically, intentionality appeared stronger when the ``wolf/lamb'' path was closer to its target, and was more salient when the speed difference between the two was significant. Furthermore, Dittrich and Lea failed to find significantly different effects when the task was described in neutral terms (letters) in comparison with when it was described in intentional terms (\ie, wolf/sheep).

\begin{figure}[t!]
	\centering
	\includegraphics[width=\linewidth]{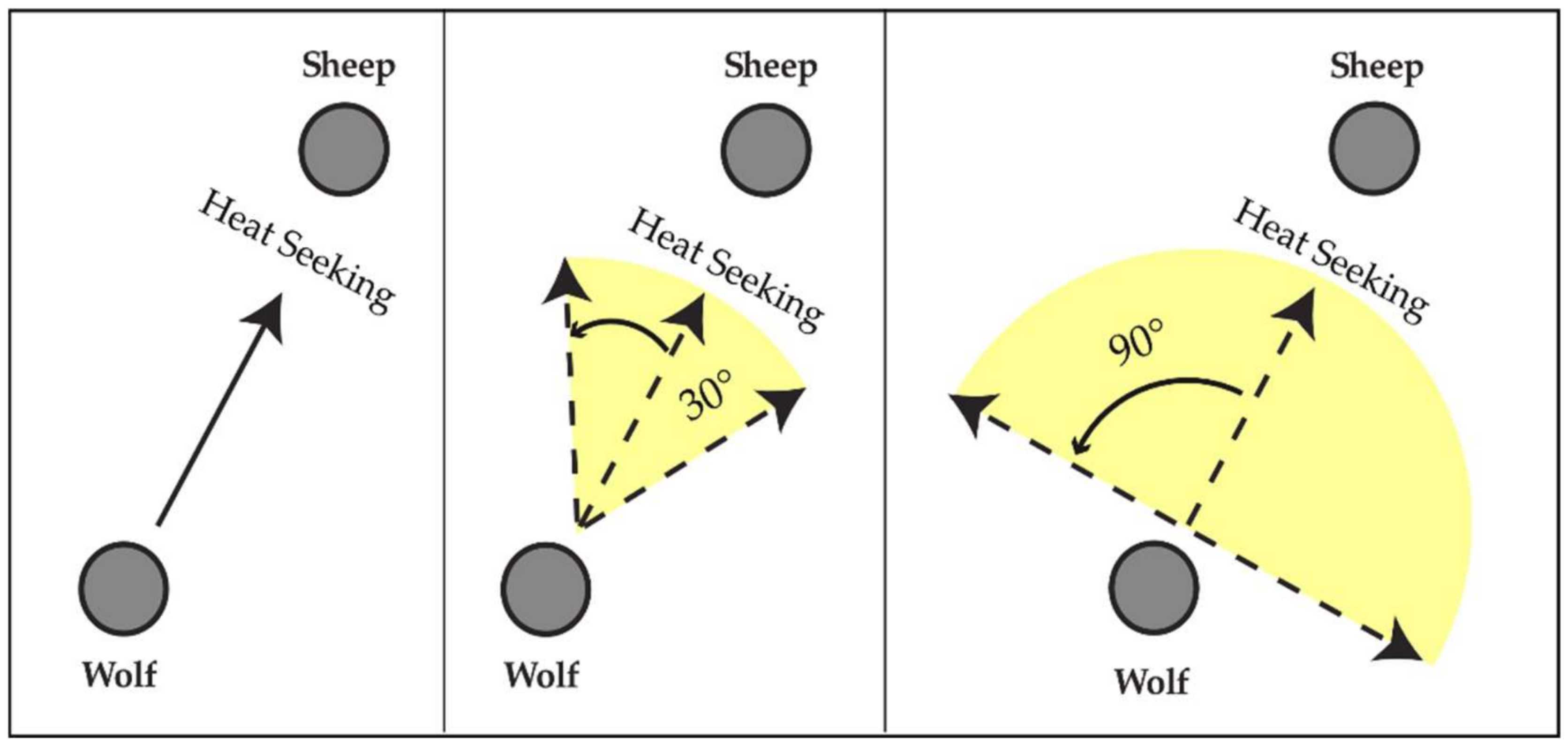}
	\caption{An illustration of \emph{chasing subtlety} manipulation in the ``Don't Get Caught'' experiment. When chasing subtlety is set to zero, the wolf always heads directly toward the (moving) sheep in a ``heat-seeking'' manner. When the chasing subtlety is set to 30, the wolf always moves in the general direction of the sheep, but is not on a perfect, heat-seeking trajectory; instead, it can move in any direction within a 60-degree window that is always centered on the moving sheep. When the chasing subtlety is set to 90, the wolf's movement is even less directed; now the wolf may head in an orthogonal direction to the (moving) sheep, though it can still never move away from it. Reproduced from Ref.~\citep{gao2009psychophysics} with permission of Elsevier Inc., \textcopyright~2009}
	\label{fig:chasing}
\end{figure}

\begin{figure*}[t!]
	\centering
	\includegraphics[width=\linewidth]{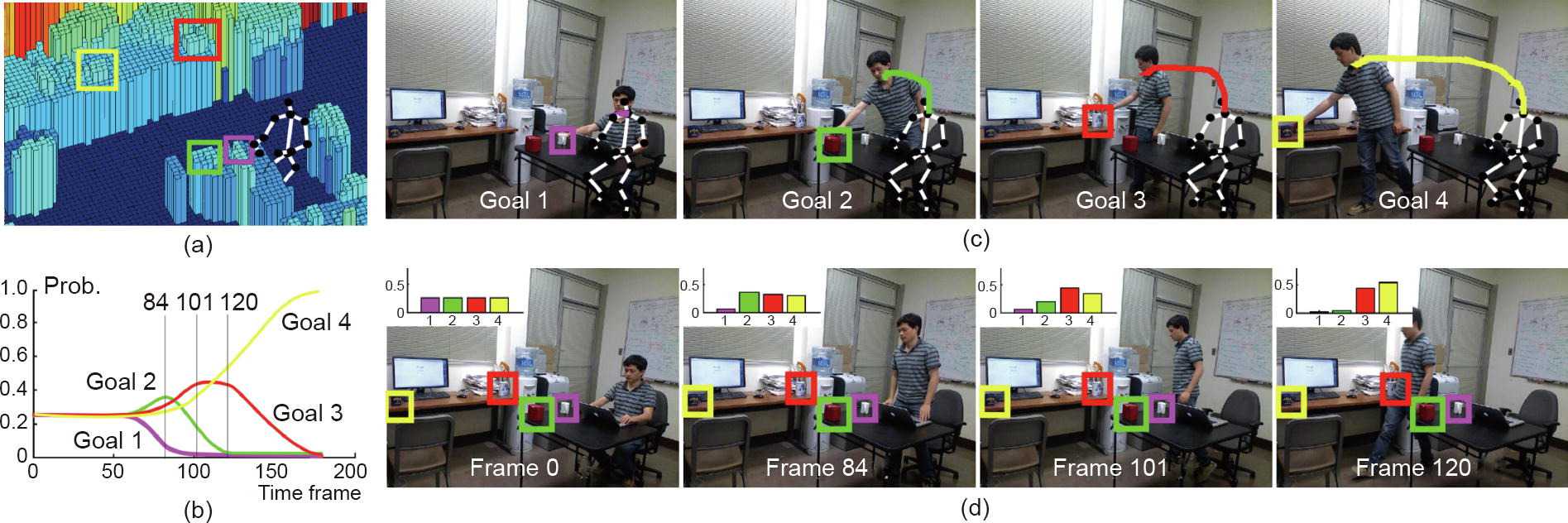}
	\caption{The plan inference task presented in Ref.~\citep{holtzen2016inferring}, seen from the perspective of an observing robot. (a) Four different goals (target objects) in a 3D scene. (b) One outcome of the proposed method: the marginal probability (Prob.) of each terminal action over time. Note that terminal actions are marginal probabilities over the probability density described by the hierarchical graphical model. (c) Four rational hierarchical plans for different goals: Goal 1 is within reach, which does not require standing up; Goal 2 requires standing up and reaching out; Goals 3 and 4 require standing up, moving, and reaching for different objects. (d) A progression of time corresponding to the results shown in (b). The action sequence and its corresponding probability distributions for each of these four goals are visualized in the bar plots in the upper left of each frame. Reproduced from Ref.~\citep{holtzen2016inferring} with permission of IEEE, \textcopyright~2016.}
	\label{fig:plan_inference}
\end{figure*}

Taken together, these experiments demonstrate that even the simplest moving shapes are irresistibly perceived in an intentional and goal-directed ``social'' way---through a holistic understanding of the events as an unfolding story whose characters have goals, beliefs, and intentions. A question naturally arises: what is the underlying mechanism with which the human visual system perceives and interprets such a richly social world? One possible mechanism governing this process that has been proposed by several philosophers and psychologists is the intuitive agency theory, which embodies the so-called ``rationality principle.'' This theory states that humans view themselves and others as \emph{causal} agents: (i) they devote their \emph{limited} time and resources only to those actions that change the world in accordance with their intentions and desires; and (ii) they achieve their intentions \emph{rationally} by maximizing their \emph{utility} while minimizing their \emph{costs}, given their \emph{beliefs} about the world~\citep{dennett1988precis,gergely1995taking,liu2019origins}.

Guided by this principle, Gao \etal~\citep{gao2009psychophysics} explored the psychophysics of chasing, one of the most salient and evolutionarily important types of intentional behavior. In an interactive ``Don't Get Caught'' game, a human participant pretended to be a sheep. The task was to detect a hidden ``wolf'' and keep away from it for 20 s. The effectiveness of the wolf's chasing was measured by the percentage of the human's escape attempts that failed. Across trials, the wolf's pursuit strategy was manipulated by a variable called \emph{chasing subtlety}, which controlled the maximum deviation from the perfect heat-seeking trajectory; see \cref{fig:chasing}~\citep{gao2009psychophysics} for more details. The results showed that humans can effectively detect and avoid wolves with small subtlety values, whereas wolves with modest subtlety values turned out to be the most ``dangerous.'' A dangerous wolf can still approach a sheep relatively quickly; meanwhile, deviation from the most efficient heat-seeking trajectory severely disrupts a human's perception of being chased, leaving the crafty wolf undetected. In other words, they can effectively stalk the human-controlled ``sheep'' without being noticed. This result is consistent with the ``rationality principle,'' where human perception assumes that an agent's intentional action will be one that maximizes its efficiency in reaching its goal.

Not only are adults sensitive to the cost of actions, as demonstrated above, but 6-to-12-month-old infants have also shown similar behavior measured in terms of habituation; they tend to look longer when an agent takes a long, circuitous route to a goal than when a shorter route is available~\citep{liu2017six,gergely2003teleological}. Crucially, infants interpret actions as directed toward goal objects, looking longer when an agent reaches for a new object, even if the reach follows a familiar path~\citep{woodward1998infants}. Recently, Liu \etal~\citep{liu2019origins} performed five looking-time experiments in which three-month-old infants viewed object-directed reaches that varied in efficiency (following the shortest physically possible path vs. a longer path), goals (lifting an object vs. causing a change in its state), and causal structures (action on contact vs. action at a distance and after a delay). Their experiments verified that infants interpret actions they cannot yet perform as causally efficacious: when people reach for and cause state changes in objects, young infants interpret these actions as goal-directed, and look longer when they are inefficient than when they are efficient. Such an early-emerging sensitivity to the causal powers of agents engaged in costly and goal-directed actions may provide one important foundation for the rich causal and social learning that characterizes our species.

The rationality principle has been formally modeled as inverse planning governed by Bayesian inference~\citep{baker2007goal,baker2009action,baker2011bayesian}. Planning is a process by which intent causes action. Inverse planning, by inverting the rational planning model via Bayesian inference that integrates the likelihood of observed actions with prior mental states, can infer the latent mental intent. Based on inverse planning, Baker \etal~\citep{baker2007goal} proposed a framework for goal inference, in which the bottom-up information of behavior observations and the top-down prior knowledge of goal space are integrated to allow inference of underlying intent. In addition, Bayesian networks, with their flexibility in representing probabilistic dependencies and causal relationships, as well as the efficiency of inference methods, have proven to be one of the most powerful and successful approaches for intent recognition~\citep{pereira2009intention,narang2019inferring,nakahashi2016modeling,baker2009action}.

Moving from the symbolic input to real video input, Holtzen \etal~\citep{holtzen2016inferring} presented an inverse planning method to infer human hierarchical intentions from partially observed RGB-D videos. Their algorithm is able to infer human intentions by reverse-engineering decision-making and action planning processes in human minds under a Bayesian probabilistic programming framework; see \cref{fig:plan_inference}~\citep{holtzen2016inferring} for more details. The intentions are represented as a novel hierarchical, compositional, and probabilistic graph structure that describes the relationships between actions and plans.

\begin{figure}[t!]
	\centering
	\includegraphics[width=\linewidth]{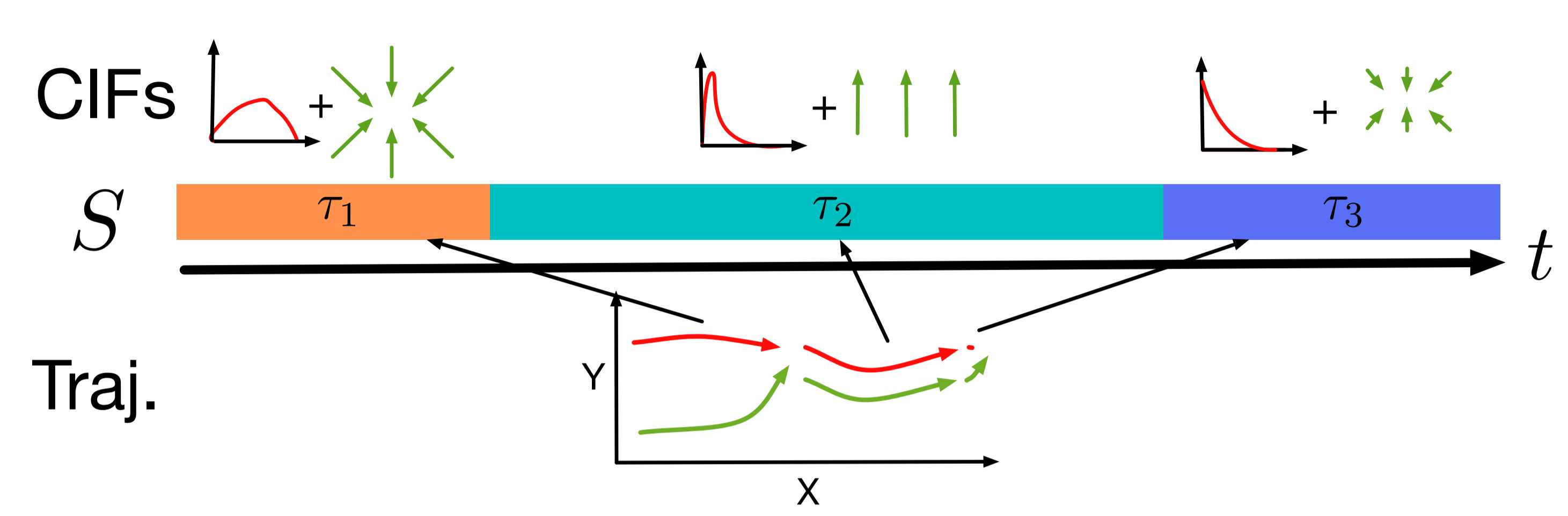}
	\caption{Inference of human interaction from motion trajectories. The top row demonstrates change within a conditional interactive field (CIF) in sub-interactions as the interaction proceeds, where the CIF models the expected relative motion pattern conditioned on the reference agent's motion. The bottom illustrates the change in interactive behaviors in terms of motion trajectories (Traj.). The colored bars in the middle depict the types of sub-interactions (S). Reproduced from Ref.~\citep{shu2018perception} with permission of Cognitive Science Society, Inc., \textcopyright~2017.}
	\label{fig:motion_trajectory}
\end{figure}

By bridging from the abstract Heider-Simmel display to aerial videos, Shu \etal~\citep{shu2018perception} proposed a method to infer humans' intentions with respect to interaction by observing motion trajectories (\cref{fig:motion_trajectory}). A non-parametric exponential potential function is taught to derive ``social force and fields'' through the calculus of variations (as in Landau physics); such force and fields explain human motion and interaction in the collected drone videos. The model's results fit well with human judgments of propensity or inclination to interact, and demonstrate the ability to synthesize decontextualized animations that have a controlled level of interactiveness.

In outdoor scenarios, Xie \etal~\citep{xie2018learning} jointly inferred object functionality and human intent by reasoning about human activities. Based on the rationality principle, the people in the observed videos are expected to intentionally take the shortest possible paths toward functional objects, subject to obstacles, that allow the people to satisfy certain of their needs (\eg, a vending machine can quench thirst); see \cref{fig:physical_social_foodtruck}. Here, the functional objects are ``dark matter'' since they are typically difficult to detect in low-resolution surveillance videos and have the functionality to ``attract'' people. Xie \etal formulated agent-based Lagrangian mechanics wherein human trajectories are probabilistically modeled as motions in many layers of ``dark energy'' fields, and wherein each agent can choose to allow a particular force field to affect its motions, thus defining the minimum-energy Dijkstra path toward the corresponding ``dark matter'' source. Such a model is effective in predicting human intentional behaviors and trajectories, localizing functional objects, and discovering distinct functional classes of objects by clustering human motion behavior in the vicinity of functional objects and agents' intentions.

\setstretch{1.02}

\begin{figure}[t!]
	\centering
	\includegraphics[width=\linewidth]{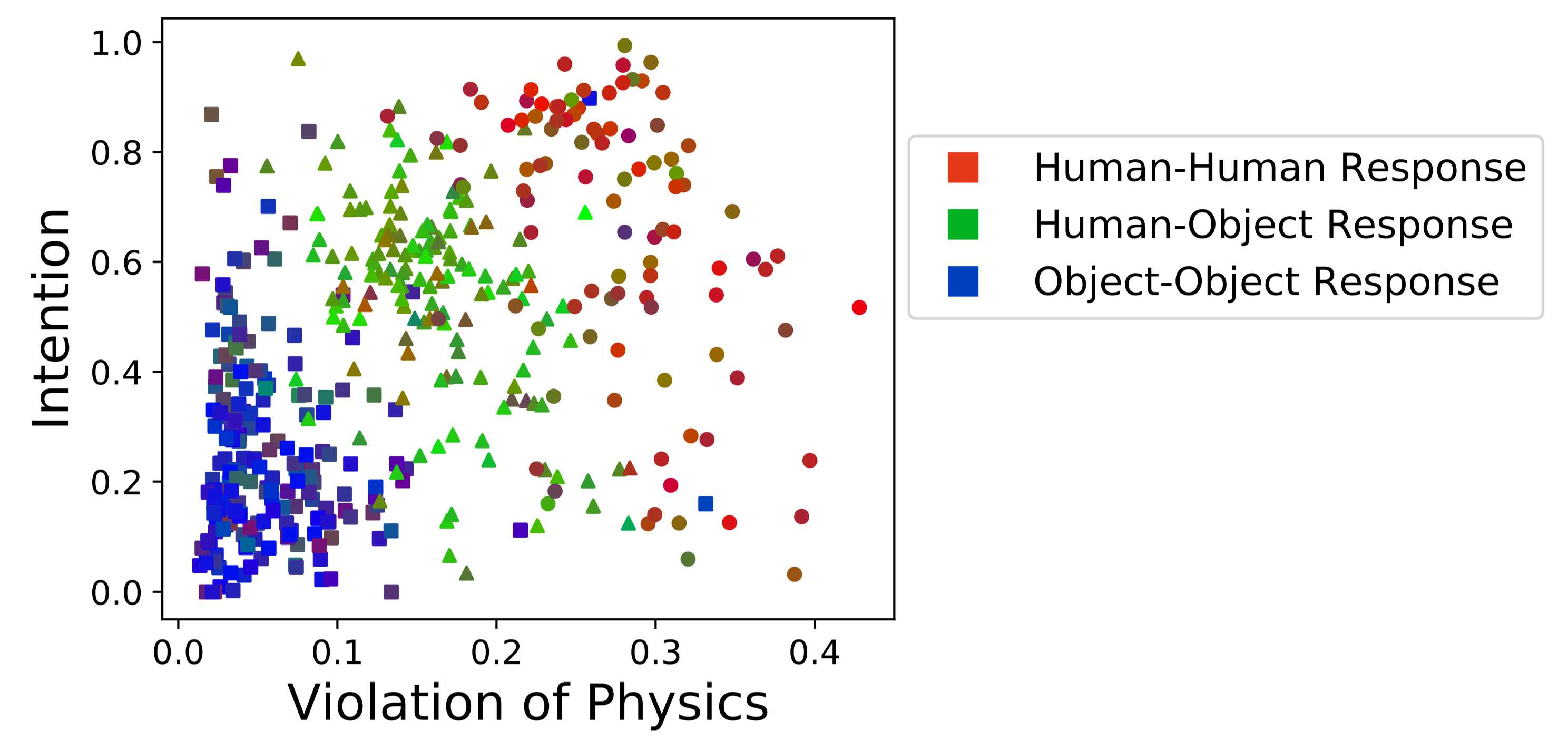}
	\caption{Constructed psychological space including human-human (HH) animations with 100\% animacy degree, human–object (HO) animations, and object-object (OO) animations. Here, a stimulus is depicted by a data point with coordinates derived by the model, and the colors of the data points indicate the average human responses to this stimulus. The two variables in the space are the average of the measures of the degree of violation of physical laws and the values indicating the presence of intent between two entities. The shapes of data points correspond to the interaction types used in the simulation for generating the corresponding
	stimuli (circle: HH, triangle: HO, square: OO). Reproduced from Ref.~\citep{shu2019partitioning} with permission of Cognitive Science Society, Inc., \textcopyright~2019.}
	\label{fig:pshyspace}
\end{figure}

\subsection{Beyond Action Prediction}\label{sec:beyond_action}

In modern computer vision and \ac{ai} systems~\citep{kong2018human}, intent is related to action prediction \emph{much} more profoundly than through simply predicting action labels. Humans have a strong and early-emerging inclination to interpret actions in terms of intention as part of a long-term process of \emph{social learning} about novel means and novel goals. From a philosophical perspective, Csibra \etal~\citep{csibra2007obsessed} contrasted three distinct mechanisms: (i) action-effect association, (ii) simulation procedures, and (iii) teleological reasoning. They concluded that action-effect association and simulation could only serve action monitoring and prediction; social learning, in contrast, requires the inferential productivity of teleological reasoning.

Simulation theory claims that the mechanism underlying the attribution of intentions to actions might rely on simulating the observed action and mapping it onto our own experiences and intent representations~\citep{blakemore2001perception}; and that such simulation processes are at the heart of the development of intentional action interpretation~\citep{biro2007becoming}. In order to understand others' intentions, humans subconsciously empathize with the person they are observing and estimate what their own actions and intentions might be in that situation. Here, action-effect association~\citep{elsner2001effect} plays an important role in quick online intent prediction, and the ability to encode and remember these two component associations contributes to infants' imitation skills and intentional action understanding~\citep{elsner2007infants}. Accumulating neurophysiological evidence supports such simulations in the human brain; one example is the mirror neuron~\citep{rizzolatti2004mirror}, which has been linked to intent understanding in many studies~\citep{kaplan2006getting,iacoboni2005grasping}. However, some studies also find that infants are capable of processing goal-directed actions before they have the ability to perform the actions themselves (\eg, Ref.~\citep{reid2007neural}), which poses challenges to the simulation theory of intent attribution.

\begin{figure*}[t!]
	\centering
	\includegraphics[width=\linewidth]{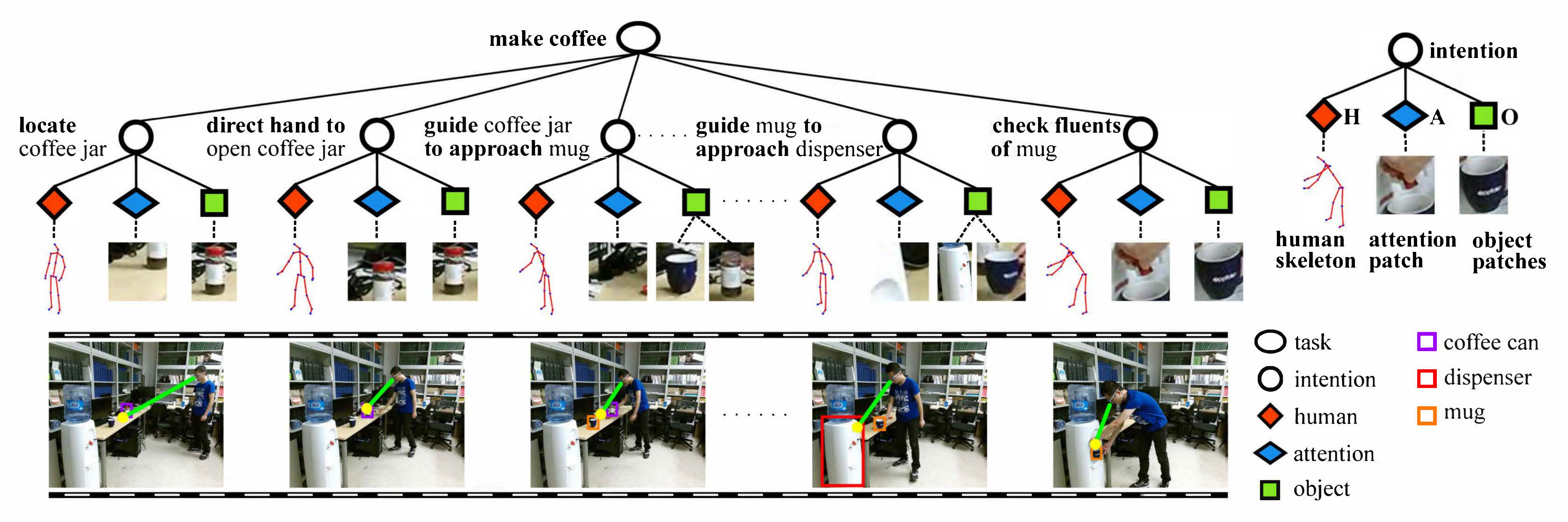}
	\caption{A task is modeled as sequential intentions in terms of hand-eye coordination with a human-attention-object (HAO) graph. Here, an intention is represented through inverse planning, in which human pose, human attention, and a visible object provide context with which to infer an agent's intention. Reproduced from Ref.~\citep{wei2018where} with permission of the authors, \textcopyright~2018.}
	\label{fig:hao}
\end{figure*}

To address social learning, a teleological action interpretational system~\citep{csibra1998teleological} takes a ``functional stance'' for the computational representation of goal-directed action~\citep{csibra2007obsessed}, where such teleological representations are generated by the aforementioned inferential ``rationality principle''~\citep{gergely2002development}. In fact, the very notion of ``action'' implies motor behavior performed by an agent that is conceived in relation to the end state that agent wants to achieve. Attributing a goal to an observed action enables humans to predict the course of future actions, evaluate causal efficacy or certain actions, and justify an action itself. Furthermore, action predictions can be made by breaking down a path toward a goal into a hierarchy of sub-goals, the most basic of which are comprised of elementary motor acts such as grasping.

These three mechanisms do not compete; instead, they complement each other. The fast effect prediction provided by action-effect associations can serve as a starting hypothesis for teleological reasoning or simulation procedure; the solutions provided by teleological reasoning in social learning can also be stored as action-effect associations for subsequent rapid recall.

\subsection{Building Blocks for Intent in Computer Vision}\label{sec:building_intention}

Understanding and predicting human intentions from images and videos is a research topic that is driven by many real-world applications, including visual surveillance, human-robot interaction, and autonomous driving. In order to better predict intent based on pixel inputs, it is necessary and indispensable to fully exploit comprehensive cues such as motion trajectory, gaze dynamics, body posture and movements, human-object relationships, and communicative gestures (\eg, pointing).

Motion trajectory alone could be a strong signal for intent prediction, as discussed in \cref{sec:rationality}. With intuitive physics and perceived intent, humans also demonstrate the ability to distinguish social events from physical events with very limited motion trajectory stimuli, such as the movements of a few simple geometric shapes. Shu \etal~\citep{shu2019partitioning} studied possible underlying computational mechanisms and proposed a unified psychological space that reveals the partition between the perception of physical events involving inanimate objects and the perception of social events involving human interactions with other agents. This unified space consists of two important dimensions: (i) an intuitive sense of whether physical laws are obeyed or violated, and (ii) an impression of whether an agent possesses intent as inferred from the movements of simple shapes; see \cref{fig:pshyspace}~\citep{shu2019partitioning}. Their experiments demonstrate that the constructed psychological space successfully partitions human perception of physical versus social events.

Eye gaze, being closely related to underlying attention, intent, emotion, personality, and anything a human is thinking and doing, also plays an important role in allowing humans to ``read'' other peoples' minds~\citep{kleinke1986gaze}. Evidence from psychology suggests that eyes are a cognitively special stimulus with distinctive, ``hardwired'' pathways in the brain dedicated to their interpretation, revealing humans' unique ability to infer others' intent from eye gazes~\citep{emery2000eyes}. Social eye gaze functions also transcend cultural differences, forming a kind of universal language~\citep{burgoon2016nonverbal}. Computer vision and \ac{ai} systems heavily rely on gazes as cues for intent prediction based on images and videos. For example, the system developed by Wei \etal~\citep{wei2018where} jointly inferred human attention, intent, and tasks from videos. Given an RGB-D video in which a human performs a task, the system answered three questions simultaneously: (i) ``Wwere is the human looking?''---attention/gaze prediction; (ii) ``why is the human looking?''---intent prediction; and (iii) ``what task is the human performing?''---task recognition. Wei \etal~\citep{wei2018where} proposed a hierarchical human-attention-object (HAO) model that represents tasks, intentions, and attention under a unified framework. Under this model, a task is represented as sequential intentions described by hand-eye coordination under a planner represented by a grammar; see \cref{fig:hao} for details~\citep{wei2018where}. 

\begin{figure*}[t]
	\centering
	\includegraphics[width=\linewidth]{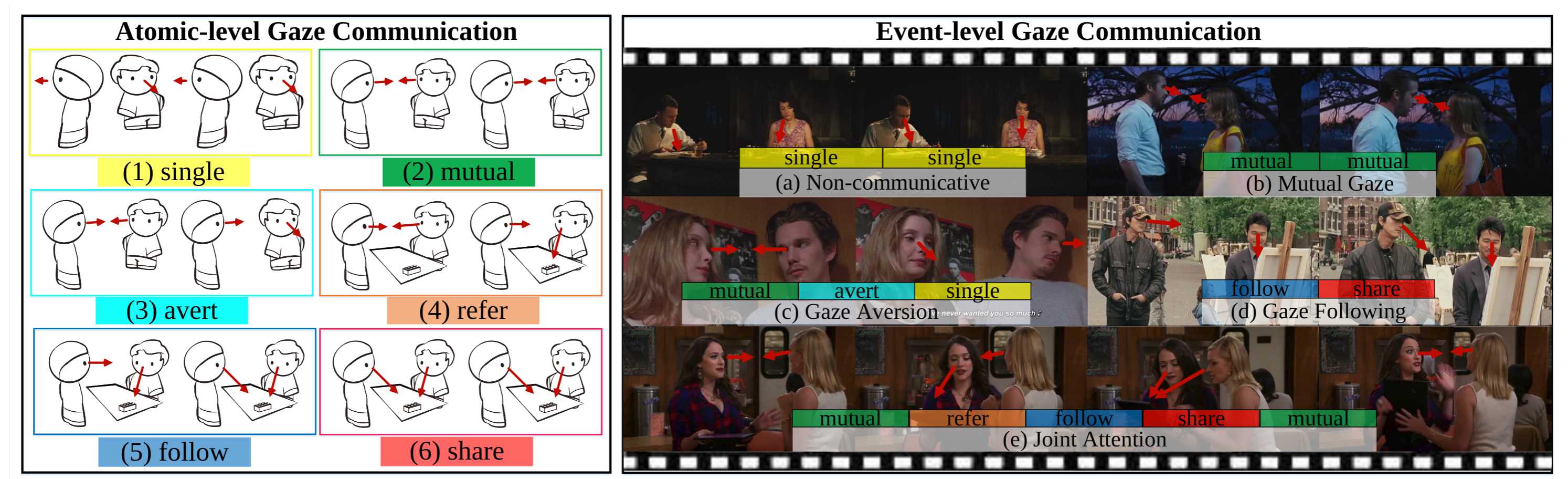}
	\caption{Human gaze communication dynamics on two hierarchical levels: (i) Atomic-level gaze communication describes the fine-grained structures in human gaze interactions; and (ii) event-level gaze communication refers to long-term social communication events temporally composed of atomic-level gaze communications. Reproduced from Ref.~\citep{fan2019understanding} with permission of the authors, \textcopyright~2019.}
	\label{fig:gaze_communication}
\end{figure*}

Communicative gazes and gestures (\eg, pointing) stand out for intent expression and perception in collaborative interactions. Humans need to recognize their partners' communicative intentions in order to collaborate with others and successfully survive in the world. Human communication in mutualistic collaboration often involves agents informing recipients of things they believe will be useful or relevant to them. Melis and Tomasello \etal~\citep{melis2019chimpanzees} investigated whether pairs of chimpanzees were capable of communicating to ensure coordination during collaborative problem-solving. In their experiments, the chimpanzee pairs needed two tools to extract fruit from an apparatus. The communicator in each pair could see the location of the tools (hidden in one of two boxes), but only the recipient could open the boxes. The communicator increasingly communicated the tools' location by approaching the baited box and giving the key needed to open it to the recipients. The recipient used these signals and obtained the tools, transferring one of the tools to the communicator so that the pair could collaborate in obtaining the fruit. As demonstrated by this study, even chimpanzees have obtained the necessary socio-cognitive skills to naturally develop a simple communicative strategy to ensure coordination in a collaborative task. To model such a capability that is demonstrated in both chimpanzees and humans, Fan \etal~\citep{fan2018inferring} studied the problem of human communicative gaze dynamics. They examined the inferring of shared eye gazes in third-person social scene videos, which is a phenomenon in which two or more individuals simultaneously look at a common target in social scenes. A follow-up work~\citep{fan2019understanding} studied various types of gaze communications in social activities from both the atomic level and event level (\cref{fig:gaze_communication}). A spatiotemporal graph network was proposed to explicitly represent the diverse interactions in the social scenes and to infer atomic-level gaze communications.

Humans communicate intentions multimodally; thus, facial expression, head pose, body posture and orientation, arm motion, gesture, proxemics, and relationships with other agents and objects can all contribute to human intent analysis and comprehension. Researchers in robotics try to equip robots with the ability to act ``naturally,'' or to be subject to ``social affordance,'' which represents action possibilities that follow basic social norms. Trick \etal~\citep{trick2019multimodal} proposed an approach for multimodal intent recognition that focuses on uncertainty reduction through classifier fusion, considering four modalities: speech, gestures, gaze directions, and scene objects. Shu \etal~\citep{shu2016learning} presented a generative model for robot learning of social affordance from human activity videos. By discovering critical steps (\ie, latent sub-goals) in interaction, and by learning structural representations of human-human (HH) and human-object-human (HOH) interactions that describe how agents' body parts move and what spatial relationships they should maintain in order to complete each sub-goal, a robot can infer what its own movement should be in reaction to the motion of the human body. Such social affordance could also be represented by a hierarchical grammar model~\citep{shu2017learning}, enabling real-time motion inference for human-robot interaction; the learned model was demonstrated to successfully infer human intent and generate humanlike, socially appropriate response behaviors in robots.

\setstretch{1}

\section{Learning Utility: The Preference of Choices}\label{sec:utility}

Rooted in the field of philosophy, economics, and game theory, the concept of utility serves as one of the most basic principles of modern decision theory: an agent makes rational decisions/choices based on their beliefs and desires to maximize its expected utility. This is known as the principle of maximum expected utility. We argue that the majority of the observational signals we encounter in daily life are driven by this simple yet powerful principle---an invisible ``dark'' force that governs the mechanism that explicitly or implicitly underlies human behaviors. Thus, studying utility could provide a computer vision or \ac{ai} system with a deeper understanding of its visual observations, thereby achieving better generalization.

According to the classic definition of utility, the utility that a decision-maker gains from making a choice is measured with a utility function. A utility function is a mathematical formulation that ranks the preferences of an individual such that $U(a) > U(b)$, where choice $a$ is preferred over choice $b$. It is important to note that the existence of a utility function that describes an agent's preference behavior does not necessarily mean that the agent is \emph{explicitly} maximizing that utility function in its own deliberations. By observing a rational agent's preferences, however, an observer can construct a utility function that represents what the agent is actually trying to achieve, even if the agent does not know it~\citep{russell2016artificial}. It is also worth noting that utility theory is a \emph{positive} theory that seeks to explain the individuals' \emph{observed} behavior and choices, which is different from a \emph{normative} theory that indicates how people \emph{should} behave; such a distinction is crucial for the discipline of economics, and for the devising of algorithms and systems to interpret observational signals.

Although Jeremy Bentham~\citep{bentham1789introduction} is often regarded as the first scholar to systematically study utilitarianism---the philosophical concept that was later borrowed by economics and game theory, the core insight motivating the theory was established much earlier by Francis Hutcheson~\citep{hutcheson1726inquiry} on action choice. In the field of philosophy, utilitarianism is considered a normative ethical theory that places the locus of right and wrong solely on the outcomes (consequences) of choosing one action/policy over others. As such, it moves beyond the scope of one's own interests and takes into account the interests of others~\citep{hutcheson1726inquiry,mill1863utilitarianism}. The term has been adopted by the field of economics, where a utility function represents a consumer's order of preferences given a set of choices. As such, the term ``utility'' is now devoid of its original meaning.

\begin{figure}[t!]
	\centering
	\begin{subfigure}[b]{0.49\linewidth}
	\includegraphics[width=\linewidth]{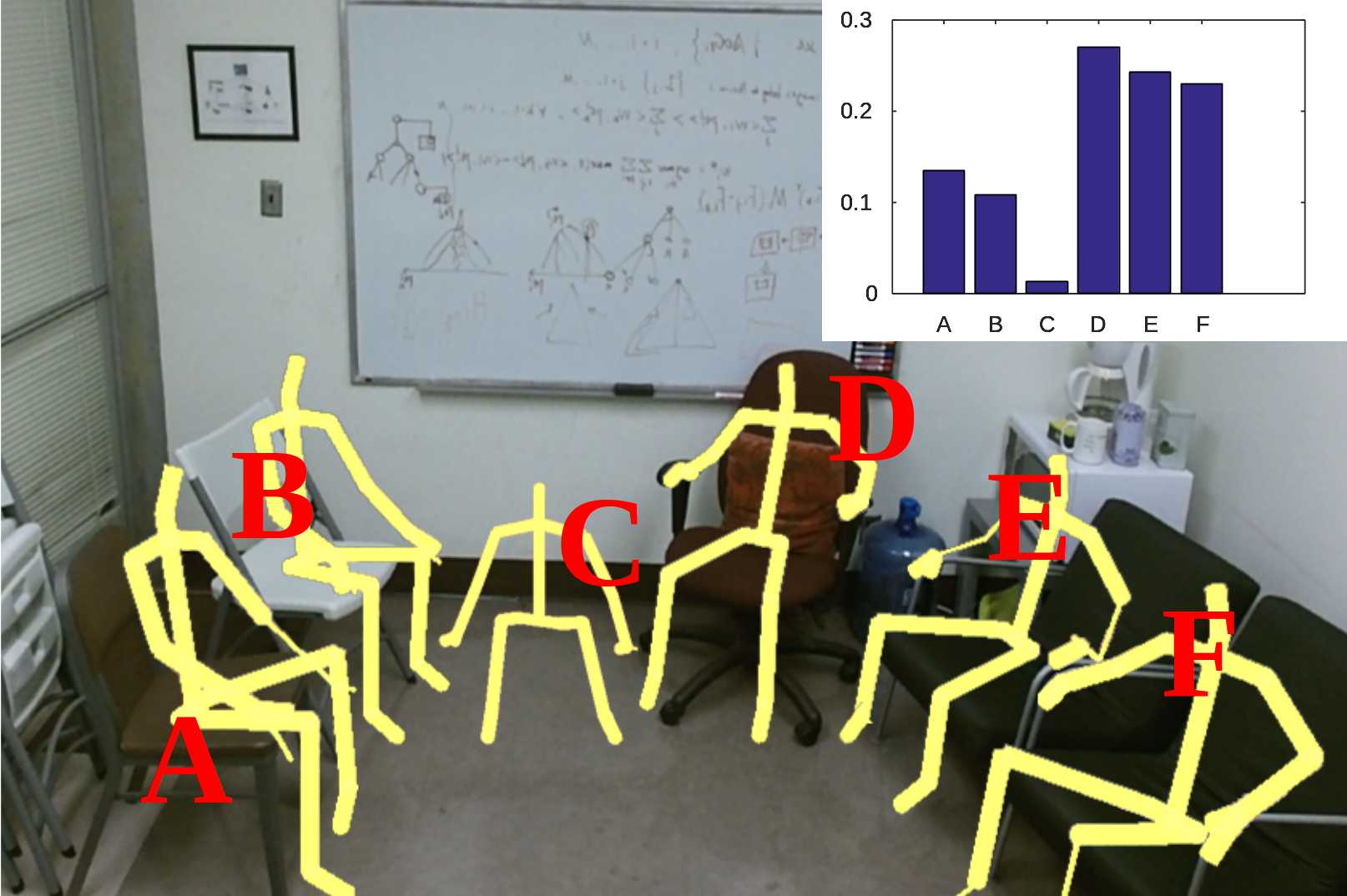}
	\caption{}
	\end{subfigure}%
	\begin{subfigure}[b]{0.51\linewidth}
	\includegraphics[width=\linewidth]{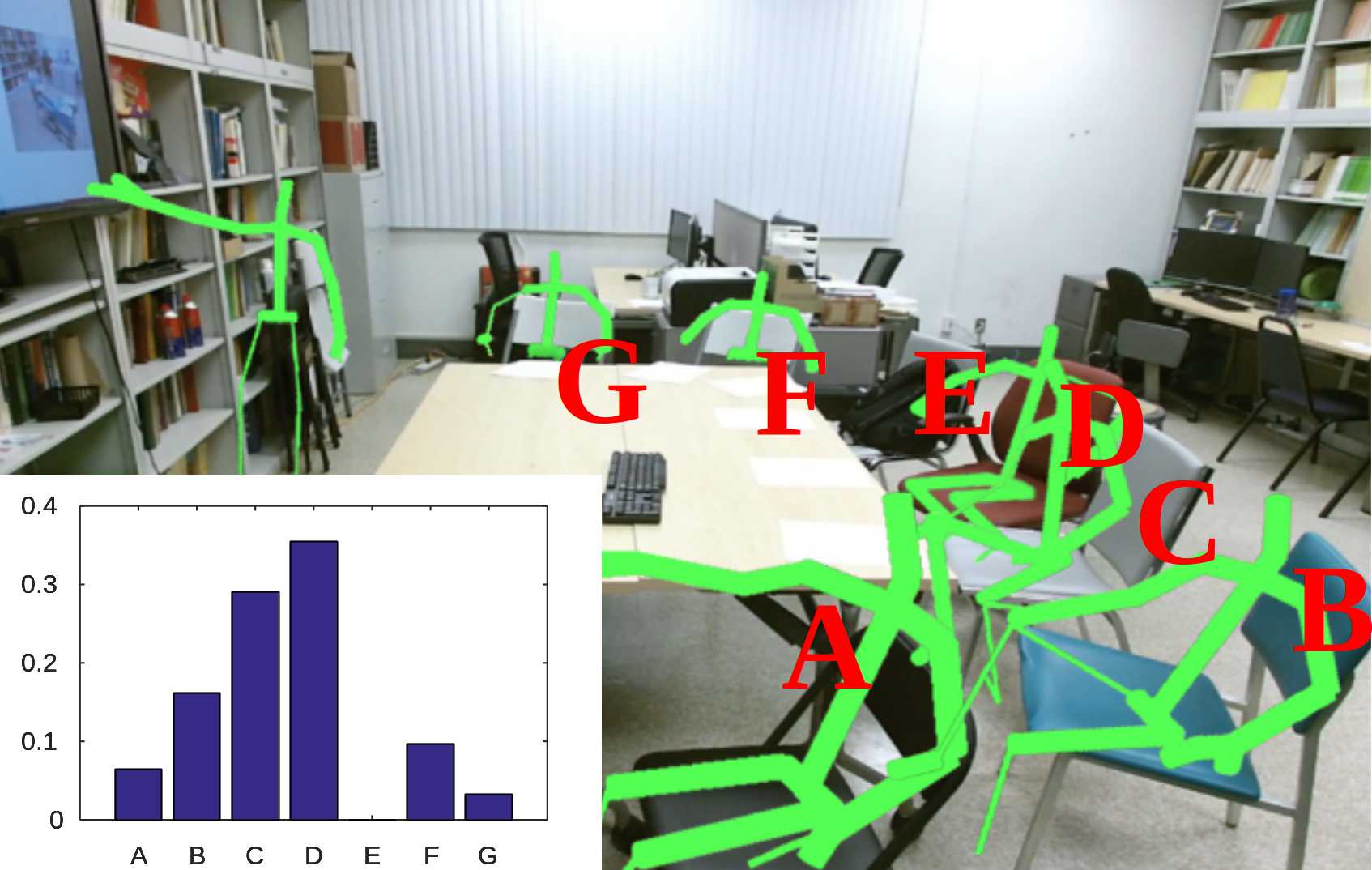}
	\caption{}
	\end{subfigure}%
	\\
	\begin{subfigure}[b]{0.435\linewidth}
	\includegraphics[width=\linewidth]{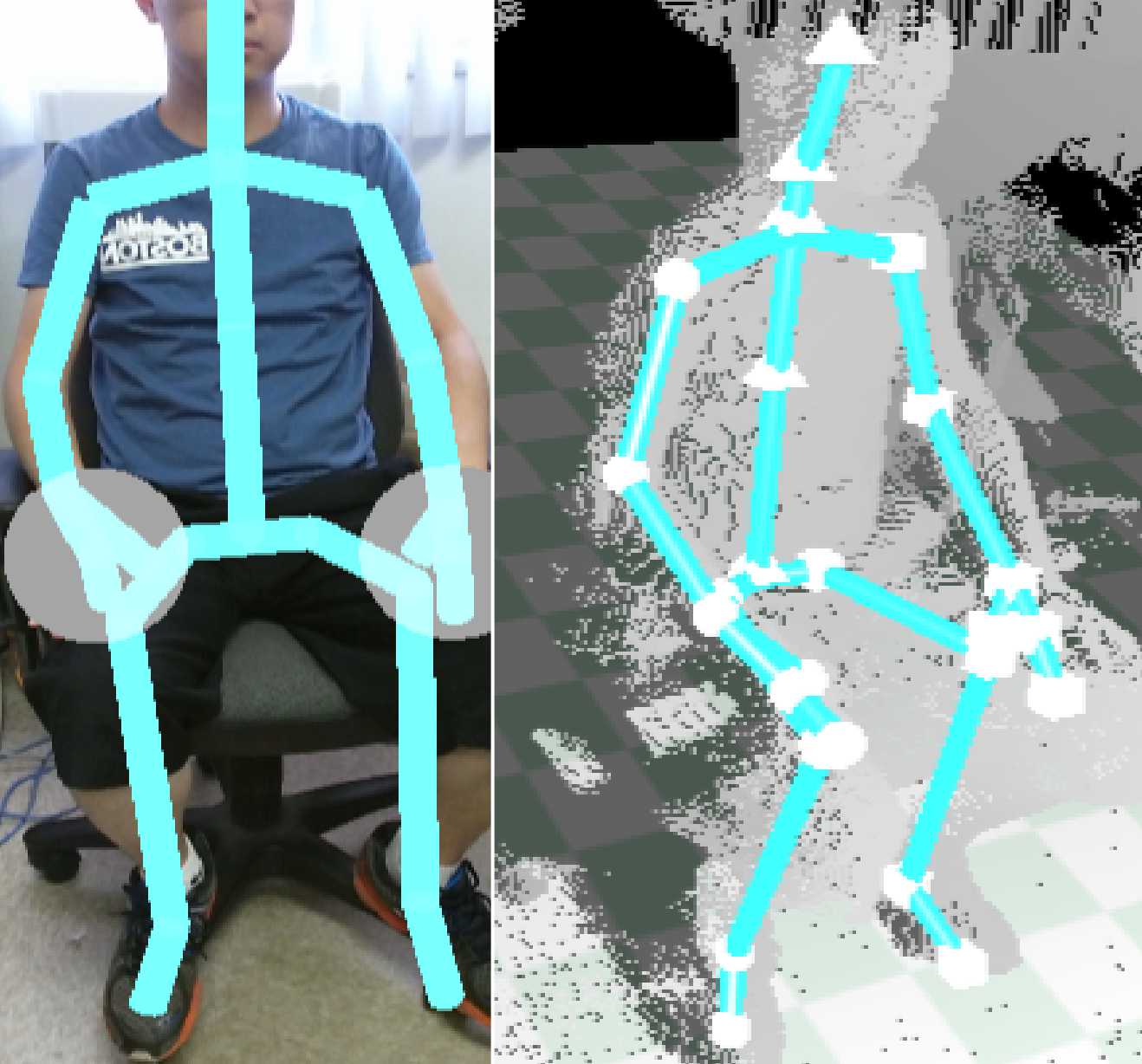}
	\caption{}
	\end{subfigure}%
	\begin{subfigure}[b]{0.565\linewidth}
	\includegraphics[width=\linewidth]{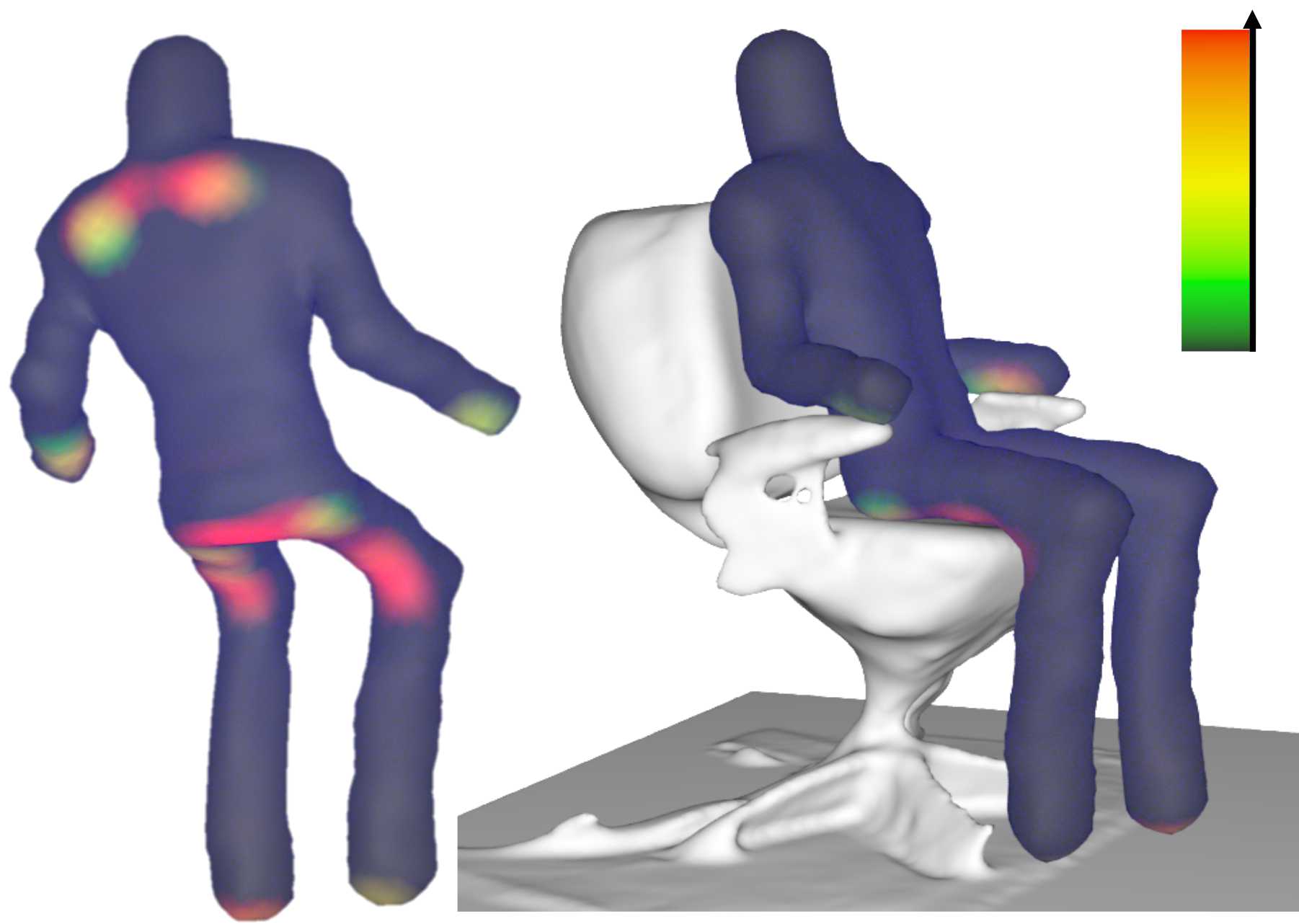}
	\caption{}
	\label{fig:physics_chair}
	\end{subfigure}%
	\caption{Examples of sitting in (a) an office and (b) a meeting room. In addition to geometry and appearance, people consider other important factors when deciding where to sit, including comfort level, reaching cost, and social goals. The histograms indicate human preferences for different candidate chairs. Based on these observations, it is possible to infer human utility during sitting from videos\citep{zhu2016inferring}. (c) The stick-man model captured using a Kinect sensor. It is first converted into a tetrahedralized human model and then segmented into 14 body parts. (d) Using FEM simulation, the forces are estimated at each vertex of the FEM mesh. Reproduced from Ref.~\citep{zhu2016inferring} with permission of the authors, \textcopyright~2016.}
	\label{fig:utility}
\end{figure}

\begin{figure*}[t!]
	\centering
	\begin{subfigure}[b]{0.462\linewidth}
	\includegraphics[width=\linewidth]{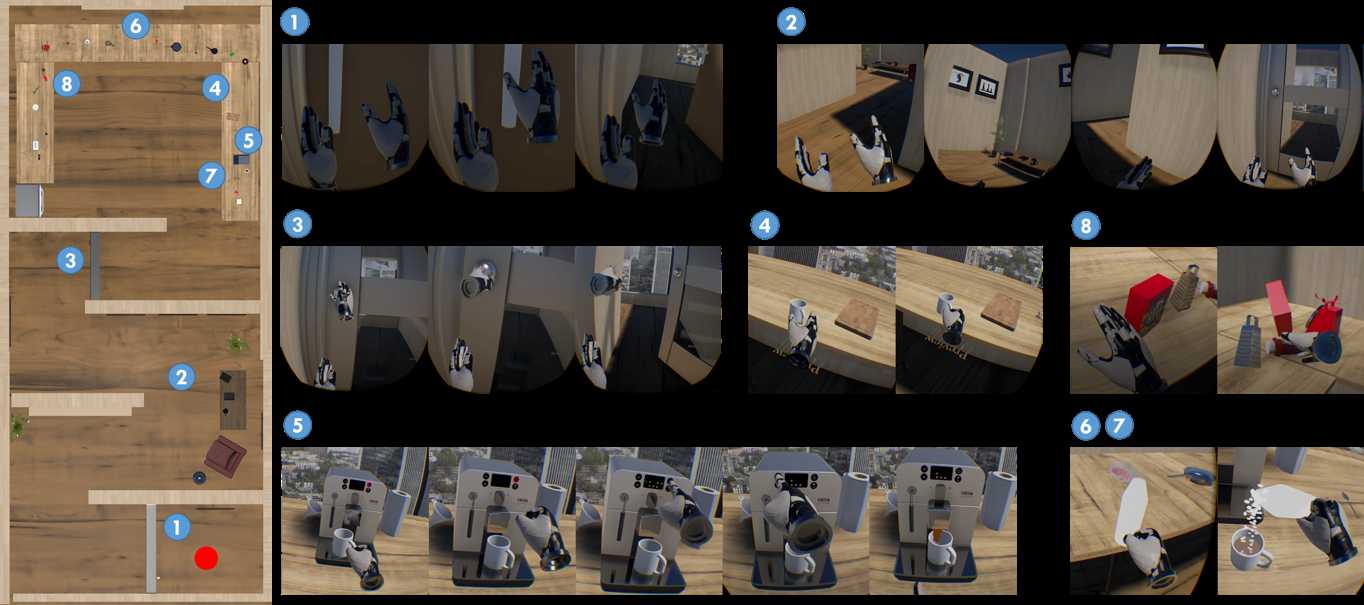}
	\caption{Various task executions in VRGym}
	\end{subfigure}%
	\begin{subfigure}[b]{0.538\linewidth}
	\includegraphics[width=\linewidth]{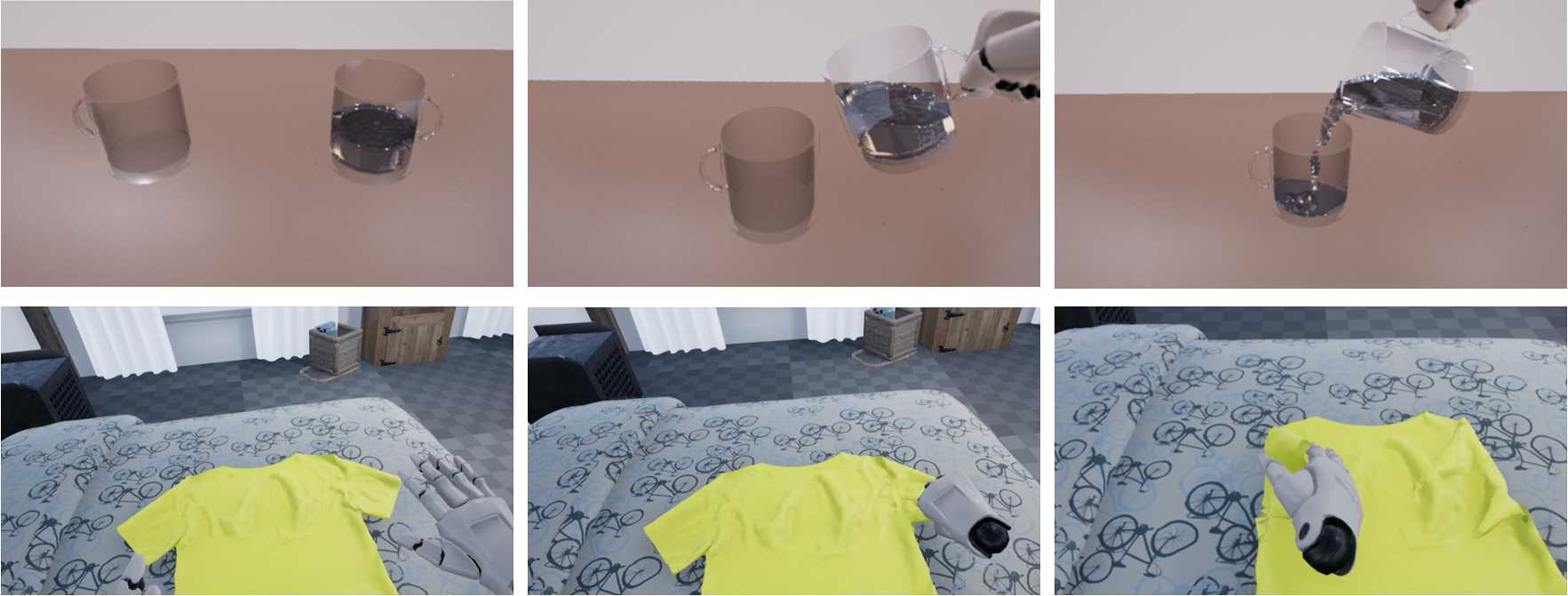}
	\caption{Real-time fluid and cloth simulations in VRGym}
	\end{subfigure}%
	\caption{VRGym, an example of a virtual environment as a large task platform. (a) Inside this platform, either a human agent or a virtual agent can perform various actions in a virtual scene and evaluate the success of task execution; (b) in addition to the rigid-body simulation, VRGym supports realistic real-time fluid and cloth simulations, leveraging state-of-the-art game engines. Reproduced from Ref.~\citep{xie2019vrgym} with permission of Association for Computing Machinery, \textcopyright~2019.}
	\label{fig:vrgym}
\end{figure*}

Formally, the core idea behind utility theory is straightforward: every possible action or state within a given model can be described with a single, uniform value. This value, usually referred to as \emph{utility}, describes the usefulness of that action within the given context. Note that the concept of \emph{utility} is not the same as the concept of \emph{value}: utility measures how much we desire something in a more subjective and context-dependent perspective, whereas value is a measurable quantity (\eg, price), which tends to be more objective. To demonstrate the usefulness of adopting the concept of utility into a computer vision and \ac{ai} system, we briefly review four recent case studies in computer vision, robotics, linguistics, and social learning that use a utility-driven learning approach.

As shown in \cref{fig:utility}~\citep{zhu2016inferring}, by observing the choices people make in videos (particularly in selecting a chair on which to sit), a computer vision system~\citep{zhu2016inferring} is able to learn the comfort intervals of the forces exerted on different body parts while sitting, thereby accounting for people's preferences in terms of human \emph{internal} utility.

Similarly, Shukla \etal~\citep{shukla2017learning} adopted the idea of learning human utility in order to teach a robotics task using human demonstrations. A proof-of-concept work shows a pipeline in which the agent learns the \emph{external} utility of humans and plans a cloth-folding task using this learned utility function. Specifically, under the assumption that the utility of the goal states is higher than that of the initial states, this system learns the \emph{external} utility of humans by ranking pairs of states extracted from images.

In addition, the rationality principle has been studied in the field of linguistics and philosophy, notably in influential work on the theory of implicature by Grice~\citep{grice1975logic}. The core insight of Grice's work is that language use is a form of rational action; thus, technical tools for reasoning about rational action should elucidate linguistic phenomena~\citep{goodman2016pragmatic}. Such a goal-directed view of language production has led to a few interesting language games~\citep{lewis2008convention,sperber1986relevance,wittgenstein1953philosophical,clark1996using,qing2015variations,goodman2013knowledge}, the development of engineering systems for natural language generation~\citep{dale1995computational}, and a vocabulary for formal descriptions of pragmatic phenomena in the field of game theory~\citep{benz2006introduction,jager2008applications}. More recently, by assuming the communications between agents to be helpful yet parsimonious, the ``Rational Speech Act''~\citep{frank2012predicting,goodman2016pragmatic} model has demonstrated promising results in solving some challenging referential games.

By materializing the internal abstract social concepts using external explicit forms, utility theory also plays a crucial role in social learning, and quantizes an actor's belief distribution. Utility, which is analogous to the ``dark'' currency circulating in society, aligns social values better among and within groups. By modeling how people value the decision-making process as permissible or not using utilities, Kleiman-Weiner \etal~\citep{kleiman2015inference} were able to solve challenging situations with social dilemma. Based on how the expected utility influences the distribution, social goals (\eg, cooperation and competition)~\citep{kleiman2016coordinate,shum2019theory} and faireness~\citep{kleiman2017constructing} can also be well explained. On a broader scale, utility can enable individuals to be self-identified in society during the social learning process; for example, when forming basic social concepts and behavior norms during the early stages of the development, children compare their own meta-values with the observed values of others~\citep{kleiman2017learning}.

\section{Summary and Discussions}\label{sec:discussion}

Robots are mechanically capable of performing a wide range of complex activities; however, in practice, they do very little that is useful for humans. Today's robots fundamentally lack physical and social common sense; this limitation inhibits their capacity to aid in our daily lives. In this article, we have reviewed five concepts that are the crucial building blocks of common sense: functionality, physics, intent, causality, and utility (FPICU). We argued that these cognitive abilities have shown potential to be, in turn, the building blocks of cognitive \ac{ai}, and should therefore be the foundation of future efforts in constructing this cognitive architecture. The positions taken in this article are not intended to serve as \emph{the} solution for the future of cognitive \ac{ai}. Rather, by identifying these crucial concepts, we want to call attention to pathways that have been less well explored in our rapidly developing \ac{ai} community. There are indeed many other topics that we believe are also essential \ac{ai} ingredients; for example:
\begin{itemize}[leftmargin=*,noitemsep,nolistsep]
	\item \emph{A physically realistic VR/MR platform: from big data to big tasks.} Since FPICU is ``dark''---meaning that it often does not appear in the form of pixels---it is difficult to evaluate FPICU in traditional terms. Here, we argue that the ultimate standard for validating the effectiveness of FPICU in \ac{ai} is to examine whether an agent is capable of (i) accomplishing the very same task using different sets of objects with different instructions and/ or sequences of actions in different environments; and (ii) rapidly adapting such learned knowledge to entirely new tasks. By leveraging state-of-the-art game engines and physics-based simulations, we are beginning to explore this possibility on a large scale; see \cref{sec:task_platform}.
	\item \emph{Social system: the emergence of language, communication, and morality.} While FPICU captures the core components of a single agent, modeling interaction among and within agents, either in collaborative or competitive situations~\citep{kinney1998learning}, is still a challenging problem. In most cases, algorithms designed for a single agent would be difficult to generalize to a \acf{mas} setting~\citep{lowe2017multi,foerster2016learning,foerster2017stabilising}. We provide a brief review of three related topics in \cref{sec:mas_emergence}.
	\item \emph{Measuring the limits of an intelligence system: IQ tests.} Studying FPICU opens a new direction of analogy and relational reasoning~\citep{holyoak2012analogy}. Apart from the four-term analogy (or proportional analogy), John C. Raven~\citep{raven1938raven} proposed the \acf{rpm} in the image domain. The RAVEN dataset~\citep{zhang2019raven} was recently introduced in the computer vision community, and serves as a systematic benchmark for many visual reasoning models. Empirical studies show that abstract-level reasoning, combined with effective feature-extraction models, could notably improve the performance of reasoning, analogy, and generalization. However, the performance gap between human and computational models calls for future research in this field; see \cref{sec:analogy}.
\end{itemize}

\begin{figure*}[t!]
	\centering
	\includegraphics[width=\linewidth]{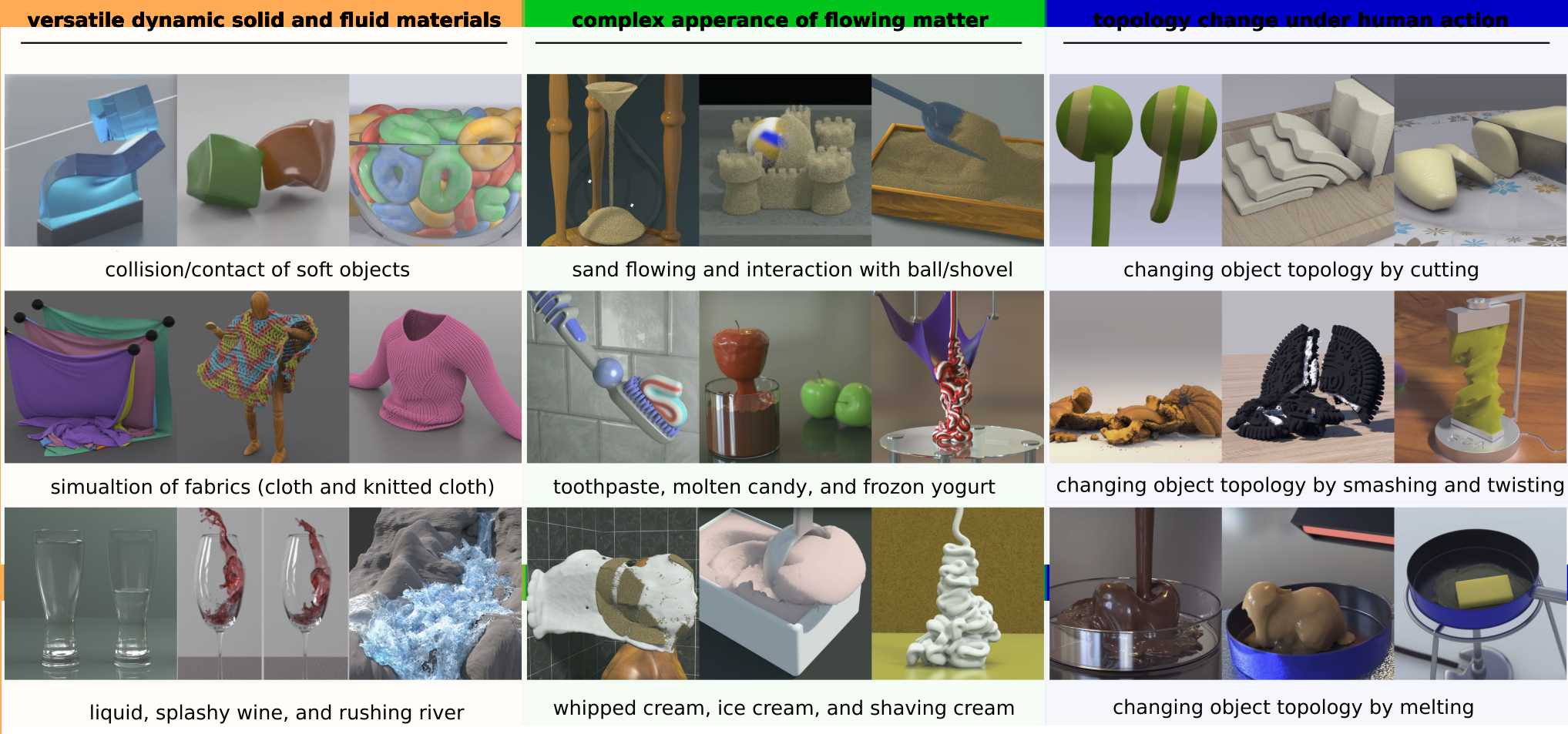}
	\caption{Diverse physical phenomena simulated using the \acf{mpm}.}
	\label{fig:simulations}
\end{figure*}

\subsection{Physically-Realistic VR/MR Platform: From Big-Data to Big-Tasks}\label{sec:task_platform}

A hallmark of machine intelligence is the capability to rapidly adapt to new tasks and ``achieve goals in a wide range of environments''~\citep{legg2007universal}. To reach this goal, we have seen the increasing use of synthetic data and simulation platforms for indoor scenes in recent years by leveraging state-of-the-art game engines and free, publicly available 3D content~\citep{mo2019partnet,chang2015shapenet,jiang2018configurable,feng2016crowd}, including MINOR~\citep{savva2017minos}, HoME~\citep{brodeur2017home}, Gibson~\citep{xia2018gibson}, House3D~\citep{wu2018building}, AI-THOR~\citep{kolve2017ai2}, VirtualHome~\citep{puig2018virtualhome}, VRGym~\citep{xie2019vrgym} (\cref{fig:vrgym}), and VRKitchen~\citep{gao2019vrkitchen}. In addition, the AirSim~\citep{shah2018airsim} open-source simulator was developed for outdoor scenarios. Such synthetic data could be relatively easily scaled up compared with traditional data collection and labeling processes. With increasing realism and faster rendering speeds built on dedicated hardware, synthetic data from the virtual world is becoming increasingly similar to data collected from the physical world. In these realistic virtual environments, it is possible to evaluate any \ac{ai} method or system from a much more holistic perspective. Using a holistic evaluation, whether a method or a system is intelligent or not is no longer measured by the successful performance of a single narrow task; rather, it is measured by the ability to perform well across various tasks: the perception of environments, planning of actions, predictions of other agents' behaviors, and ability to rapidly adapt learned knowledge to new environments for new tasks.

To build this kind of task-driven evaluation, physics-based simulations for multi-material, multi-physics phenomena (\cref{fig:simulations}) will play a central role. We argue that cognitive \ac{ai} needs to accelerate the pace of its adoption of more advanced simulation models from computer graphics, in order to benefit from the capability of highly predictive forward simulations, especially graphics processing unit (GPU) optimizations that allow real-time performance~\citep{gao2019gpu}. Here, we provide a brief review of the recent physics-based simulation methods, with a particular focus on the \acf{mpm}.

The accuracy of physics-based reasoning greatly relies on the fidelity of a physics-based simulation. Similarly, the scope of supported virtual materials and their physical and interactive properties directly determine the complexity of the \ac{ai} tasks involving them. Since the pioneering work of Terzopoulos\etal~\citep{terzopoulos1987elastically,terzopoulos1988modeling} for solids and that of Foster and Metaxas~\citep{foster1996realistic} for fluids, many mathematical and physical models in computer graphics have been developed and applied to the simulation of solids and fluids in a 3D virtual environment.

For decades, the computer graphics and computational physics community sought to increase the robustness, efficiency, stability, and accuracy of simulations for cloth, collisions, deformable, fire, fluids, fractures, hair, rigid bodies, rods, shells, and many other substances. Computer simulation-based engineering science plays an important role in solving many modern problems as an inexpensive, safe, and analyzable companion to physical experiments. The most challenging problems are those involving extreme deformation, topology change, and interactions among different materials and phases. Examples of these problems include hypervelocity impact, explosion, crack evolution, fluid-structure interactions, climate simulation, and ice-sheet movements. Despite the rapid development of computational solid and fluid mechanics, effectively and efficiently simulating these complex phenomena remains difficult. Based on how the continuous physical equations are discretized, the existing methods can be classified into the following categories:
\begin{enumerate}[leftmargin=*,noitemsep,nolistsep]
	\item Eulerian grid-based approaches, where the computational grid is fixed in space, and physical properties advect through the deformation flow. A typical example is the Eulerian simulation of free surface incompressible flow~\citep{stam1999stable,bridson2015fluid}. Eulerian methods are more error-prone and require delicate treatment when dealing with deforming material interfaces and boundary conditions, since no explicit tracking of them is available.
	\item Lagrangian mesh-based methods, represented by \ac{fem}~\citep{bonet1997nonlinear,blemker2005fast,hegemann2013level}, where the material is described with and embedded in a deforming mesh. Mass, momentum, and energy conservation can be solved with less effort. The main problem of ac{fem} is mesh distortion and lack of contact during large deformations~\citep{gast2015optimization,li2019decomposed} or topologically changing events~\citep{wang2014adaptive}.
	\item Lagrangian mesh-free methods, such as smoothed particle hydrodynamics (SPH)~\citep{monaghan1992smoothed} and the reproducing kernel particle method (RKPM)~\citep{liu1995reproducing}. These methods allow arbitrary deformation but require expensive operations such as neighborhood searching~\citep{li2002meshfree}. Since the interpolation kernel is approximated with neighboring particles, these methods also tend to suffer from numerical instability issues.
	\item Hybrid Lagrangian–Eulerian methods, such as the arbitrary Lagrangian–Eulerian (ALE) methods~\citep{donea1982arbitrary} and the \ac{mpm}. These methods (particularly the \ac{mpm}) combine the advantages of both Lagrangian methods and Eulerian grid methods by using a mixed representation.
\end{enumerate}

In particular, as a generalization of the hybrid fluid implicit particle (FLIP) method~\citep{brackbill1986flip,jiang2015affine} from computational fluid dynamics to computational solid mechanics, the \ac{mpm} has proven to be a promising discretization choice for simulating many solid and fluid materials since its introduction two decades ago~\citep{sulsky1994particle,sulsky1995application}. In the field of visual computing, existing work includes snow~\citep{stomakhin2013material,gaume2018dynamic}, foam~\citep{ram2015material,yue2015continuum,fang2019silly}, sand~\citep{klar2016drucker,daviet2016semi}, rigid body~\citep{hu2018moving}, fracture~\citep{wang2019simulation,wolper2019cd}, cloth~\citep{jiang2017anisotropic}, hair~\citep{han2019hybrid}, water~\citep{fu2017polynomial}, and solid-fluid mixtures~\citep{stomakhin2014augmented,tampubolon2017multi,gao2018animating}. In computational engineering science, this method has also become one of the most recent and advanced discretization choices for various applications. Due to its many advantages, it has been successfully applied to tackling extreme deformation events such as fracture evolution~\citep{nairn2003material}, material failure~\citep{chen2005bifurcation,schreyer2002modeling}, hyper-velocity impact~\citep{sulsky1996axisymmetric,huang2008shared}, explosion~\citep{hu2006model}, fluid-structure interaction~\citep{york2000fluid,bandara2015coupling}, biomechanics~\citep{guilkey2006computational}, geomechanics~\citep{huang2010material}, and many other examples that are considerably more difficult when addressed with traditional, non-hybrid approaches. In addition to experiencing a tremendously expanding scope of application, the \ac{mpm}'s discretization scheme has been extensively improved~\citep{fang2018temporally}. To alleviate numerical inaccuracy and stability issues associated with the original \ac{mpm} formulation, researchers have proposed different variations of the \ac{mpm}, including the generalized interpolation material point (GIMP) method~\citep{bardenhagen2004generalized,gao2017adaptive}, the convected particle domain interpolation (CPDI) method~\citep{sadeghirad2011convected}, and the dual domain material point (DDMP) method~\citep{zhang2011material}.

\subsection{Social System: Emergence of Language, Communication, and Morality}\label{sec:mas_emergence}

\begin{figure*}[t!]
	\centering
	\includegraphics[width=\linewidth]{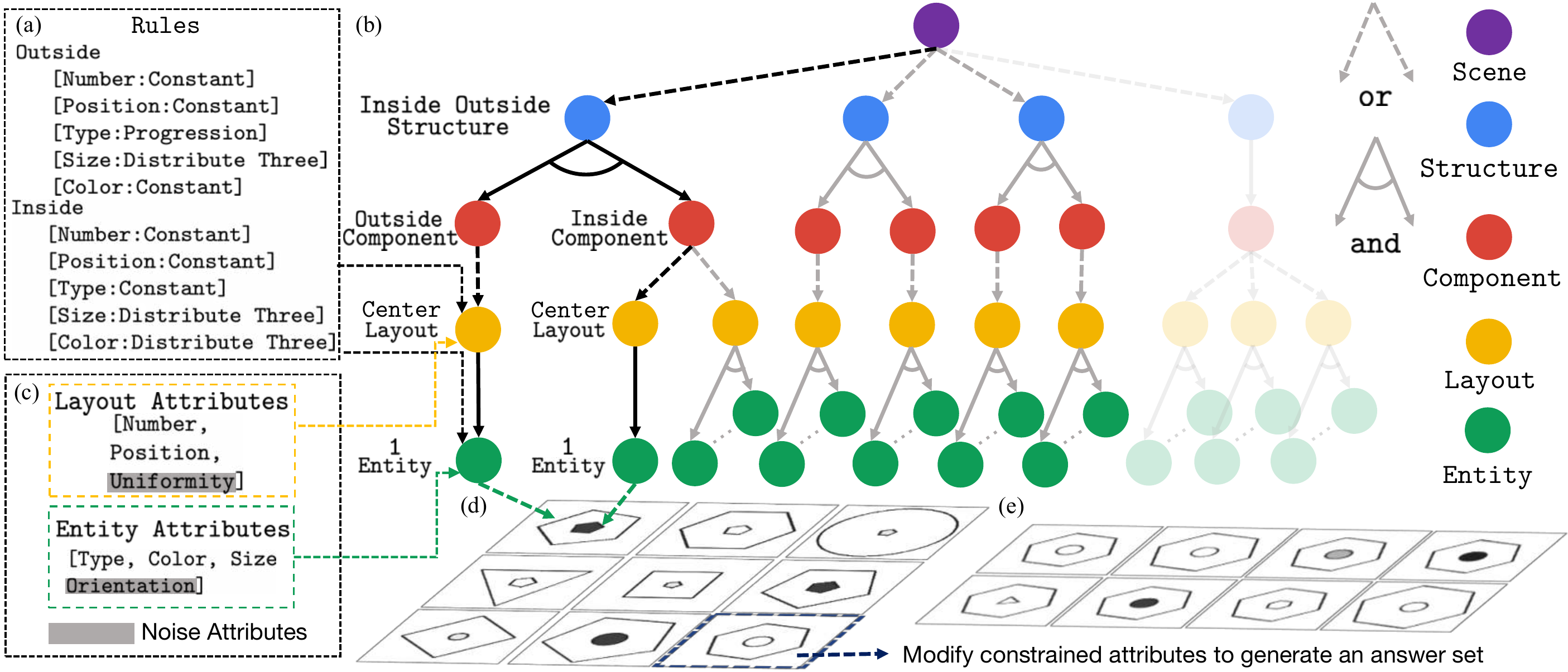}
	\caption{The RAVEN creation process proposed in Ref.~\citep{zhang2019raven}. A graphical illustration of (a) the grammar production rules used in (b) A-SIG. (c) Note that Layout and Entity have associated attributes. (d) A sample problem matrix and (e) a sample candidate set. Reproduced from Ref.~\citep{zhang2019raven} with permission of the authors, \textcopyright~2019.}
	\label{fig:raven}
\end{figure*}

Being able to communicate and collaborate with other agents is a crucial component of \ac{ai}. In classic \ac{ai}, a multi-agent communication strategy is modeled using a predefined rule-based system (\eg, adaptive learning of communication strategies in \ac{mas}~\citep{kinney1998learning}). To scale up from rule-based systems, decentralized partially observable Markov decision processes were devised to model multi-agent interaction, with communication being considered as a special type of action~\citep{bernstein2002complexity,goldman2003optimizing}. As with the success of \ac{rl} in single-agent games~\citep{mnih2013playing}, generalizing Q-learning~\citep{tampuu2017multiagent,foerster2017stabilising} and actor-critic~\citep{lowe2017multi,foerster2018counterfactual}-based methods from single-agent system to \ac{mas} have been a booming topic in recent years.

The emergence of language is also a fruitful topic in multi-agent decentralized collaborations. By modeling communication as a particular type of action, recent research~\citep{foerster2016learning,sukhbaatar2016learning,mordatch2018emergence} has shown that agents can learn how to communicate with continuous signals that are only decipherable within a group. The emergence of more realistic communication protocols using discrete messages has been explored in various types of communication games~\citep{lazaridou2016multi,havrylov2017emergence,evtimova2017emergent,lazaridou2018emergence}, in which agents need to process visual signals and attach discrete tokens to attributes or semantics of images in order to form effective protocols. By letting groups of agents play communication games spontaneously, several linguistic phenomena in emergent communication and language have been studied~\citep{wagner2003progress,ibsen2018language,graesser2019emergent}.

Morality is an abstract and complex concept composed of common principles such as fairness, obligation, and permissibility. It is deeply rooted in the tradeoffs people make every day when these moral principles come into conflict with one another~\citep{dupoux2007universal,mikhail2011elements}. Moral judgment is extremely complicated due to the variability in standards among different individuals, social groups, cultures, and even forms of violation of ethical rules. For example, two distinct societies could hold opposite views on preferential treatment of kin: one might view it as corrupt, the other as a moral obligation~\citep{kleiman2017learning}. Indeed, the same principle might be viewed differently in two social groups with distinct cultures~\citep{blake2015ontogeny}. Even within the same social group, different individuals might have different standards on the same moral principle or event that triggers moral judgment~\citep{henrich2001search,house2013ontogeny,graham2016cultural}. Many works have proposed theoretical accounts for categorizing the different measures of welfare used in moral calculus, including ``base goods'' and ``primary goods''~\citep{hurka2000virtue,rawls1971theory}, ``moral foundations''~\citep{haidt2007new}, and the feasibility of value judgment from an infant's point of view~\citep{hamlin2013moral}. Despite its complexity and diversity, devising a computational account of morality and moral judgment is an essential step on the path toward building humanlike machines. One recent approach to moral learning combines utility calculus and Bayesian inference to distinguish and evaluate different principles~\citep{kleiman2017learning,kim2018computational,kleiman2015inference}.

\subsection{Measuring the Limits of Intelligence System: IQ tests}\label{sec:analogy}

In the literature, we call two cases analogous if they share a common \emph{relationship}. Such a relationship does not need to be among entities or ideas that use the same label across disciplines, such as computer vision and \ac{ai}; rather, ``analogous'' emphasizes commonality on a more abstract level. For example, according to Ref.~\citep{holyoak1997analogical}, the earliest major scientific discovery made through analogy can be dated back to imperial Rome, when investigators analogized waves in water and sound. They posited that sound waves and water waves share similar behavioral properties; for example, their intensities both diminish as they propagate across space. To make a successful analogy, the key is to understand \emph{causes and their effects}~\citep{cheng2012causal}.

The history of analogy can be categorized into three streams of research; see Ref.~\citep{holyoak2012analogy} for a capsule history and review of the literature. One stream is the psychometric tradition of four-term or ``proportional'' analogies, the earliest discussions of which can be traced back to Aristotle~\citep{hesse1966models}. An example in \ac{ai} is the \emph{word2vec} model~\citep{mikolov2013distributed,mikolov2013efficient}, which is capable of making a four-term word analogy; for example, [king:queen::man:woman]. In the image domain, a similar test was invented by John C. Raven~\citep{raven1938raven}---the \acf{rpm}.

\ac{rpm} has been widely accepted and is believed to be highly correlated with real intelligence~\citep{carpenter1990one}. Unlike \ac{vqa}~\citep{antol2015vqa}, which lies at the periphery of the cognitive ability test circle~\citep{carpenter1990one}, \ac{rpm} lies directly at the center: it is diagnostic of abstract and structural reasoning ability~\citep{snow1984the}, and captures the defining feature of high-level cognition---that is, \emph{fluid intelligence}~\citep{jaeggi2008improving}. It has been shown that \ac{rpm} is more difficult than existing visual reasoning tests in the following ways~\citep{zhang2019raven}:
\begin{itemize}[leftmargin=*,noitemsep,nolistsep]
	\item Unlike \ac{vqa}, where natural language questions usually imply what the agent should pay attention to in an image, \ac{rpm} relies merely on visual clues provided in the matrix. The \emph{correspondence problem} itself, that is, the ability to find corresponding objects across frames to determine their relationship, is already a major factor distinguishing populations of different intelligence~\citep{carpenter1990one}.
	\item While current visual reasoning tests only require spatial and semantic understanding, \ac{rpm} needs joint spatial-temporal reasoning in the problem matrix and the answer set. The limit of \emph{short-term memory}, the ability to understand \emph{analogy}, and the grasp of \emph{structure} must be taken into consideration in order to solve an \ac{rpm} problem.
	\item Structures in \ac{rpm} make the compositions of rules much more complicated. Problems in \ac{rpm} usually include more sophisticated logic with recursions. Combinatorial rules composed at various levels also make the reasoning process extremely difficult.
\end{itemize}

The RAVEN dataset~\citep{zhang2019raven} was created to push the limit of current vision systems' reasoning and analogy-making ability, and to promote further research in this area. The dataset is designed to focus on reasoning and analogizing instead of only visual recognition. It is unique in the sense that it builds a semantic link between the visual reasoning and structural reasoning in \ac{rpm} by grounding each problem into a sentence derived from an attributed stochastic image grammar \acf{asig}: each instance is a sentence sampled from a predefined A-SIG, and a rendering engine transforms the sentence into its corresponding image. (See \cref{fig:raven}~\citep{zhang2019raven} for a graphical illustration of the generation process.) This semantic link between vision and structure representation opens new possibilities by breaking down the problem into image understanding and abstract-level structure reasoning. Zhang \etal~\citep{zhang2019raven} empirically demonstrated that models using a simple structural reasoning module to incorporate both vision-level understanding and abstract-level reasoning and analogizing notably improved their performance in \ac{rpm}, whereas a variety of prior approaches to relational learning performed only slightly better than a random guess. 

Analogy consists of more than mere spatiotemporal parsing and structural reasoning. For example, the \emph{contrast effect}~\citep{bower1961contrast} has been proven to be one of the key ingredients in relational and analogical reasoning for both human and machine learning~\citep{meyer1951effects,schrier1956effect,shapley1978effect,lawson1957brightness,amsel1962frustrative}. Originating from perceptual learning~\citep{gibson1955perceptual,gibson2014ecological}, it is well established in the field of psychology and education~\citep{catrambone1989overcoming,gentner2001structural,hammer2009development,gick1992contrasting,haryu2011object} that teaching new concepts by comparing noisy examples is quite effective. Smith and Gentner~\citep{smith2014role} summarized that comparing cases facilitates transfer learning and problem-solving, as well as the ability to learn relational categories. In his structure-mapping theory, Gentner~\citep{gentner1983structure} postulated that learners generate a structural alignment between two representations when they compare two cases. A later article~\citep{gentner1994structural} firmly supported this idea and showed that finding the individual difference is easier for humans when similar items are compared. A more recent study from Schwartz \etal~\citep{schwartz2011practicing} also showed that contrasting cases helps to foster an appreciation of deep understanding. To retrieve this missing treatment of contrast in machine learning, computer vision and, more broadly, in \ac{ai}, Zhang \etal~\citep{zhang2019learning} proposed methods of learning perceptual inference that explicitly introduce the notion of contrast in model training. Specifically, a contrast module and a contrast loss are incorporated into the algorithm at the model level and at the objective level, respectively. The permutation-invariant contrast module summarizes the common features from different objects and distinguishes each candidate by projecting it onto its residual on the common feature space. The final model, which comprises ideas from contrast effects and perceptual inference, achieved state-of-the-art performance on major \ac{rpm} datasets.

Parallel to work on \ac{rpm}, work on \emph{number sense}~\cite{dehaene2011number} bridges the induction of symbolic concepts and the competence of problem-solving; in fact, number sense could be regarded as a mathematical counterpart to the visual reasoning task of \ac{rpm}. A recent work approaches the analogy problem from this perspective of strong mathematical reasoning~\cite{zhang2020machine}. Zhang \etal~\cite{zhang2020machine} studied the machine number-sense problem and proposed a dataset of visual arithmetic problems for abstract and relational reasoning, where the machine is given two figures of numbers following hidden arithmetic computations and is tasked to work out a missing entry in the final answer. Solving machine number-sense problems is non-trivial: the system must both recognize a number and interpret the number with its contexts, shapes, and relationships (\eg, symmetry), together with its proper operations. Experiments show that the current neural-network-based models do not acquire mathematical reasoning abilities after learning, whereas classic search-based algorithms equipped with an additional perception module achieve a sharp performance gain with fewer search steps. This work also sheds some light on how machine reasoning could be improved: the fusing of classic search-based algorithms with modern neural networks in order to discover essential number concepts in future research would be an encouraging development.

\section{Acknowledgments}

This article presents representative work selected from a US and UK Multidisciplinary University Research Initiative (MURI) collaborative project on visual commonsense reasoning, focusing on human vision and computer vision. The team consists of interdisciplinary researchers in computer vision, psychology, cognitive science, machine learning, and statistics from both the US (in alphabetical order: Carnegie Mellon University (CMU), Massachusetts Institute of Technology (MIT), Stanford University, University of California at Los Angeles (UCLA), University of Illinois at Urbana-Champaign (UIUC), and Yale University) and the UK (in alphabetical order: University of Birmingham, University of Glasgow, University of Leeds, and University of Oxford)\footnote{See~\url{https://vcla.stat.ucla.edu/MURI_Visual_CommonSense/}}. The MURI team also holds an annual review meeting at various locations together with two related series of CVPR/CogSci workshops\footnote{Workshop on VisionMeetsCognition: Functionality, Physics, Intentionality, and Causality:~\url{https://www.visionmeetscognition.org/}}\footnote{Workshop on 3D Scene Understanding for Vision, Graphics, and Robotics:~\url{https://scene-understanding.com/}}.

We are grateful to the editor of the special issue and the two reviewers for their valuable comments that have helped improve the presentation of the paper. We thank the following colleagues for helpful discussions on various sections: Professor Chenfanfu Jiang at the University of Pennsylvania; Dr. Behzad Kamgar-Parsi at the Office of Naval Research (ONR) and Dr. Bob Madahar at the Defence Science and Technology Laboratory (DSTL); Luyao Yuan, Shuwen Qiu, Zilong Zheng, Xu Xie, Xiaofeng Gao, and Qingyi Zhao at UCLA; Dr. Mark Nitzberg, Dr. Mingtian Zhao, and Helen Fu at DMAI, Inc.; and Dr. Yibiao Zhao at ISEE, Inc.

The work reported herein is supported by MURI ONR (N00014-16-1-2007), DARPA XAI (N66001-17-2-4029), and ONR (N00014-19-1-2153).

\renewcommand*{\bibfont}{\fontsize{8.2}{8.2} \selectfont}
\bibliography{reference}

\end{document}